\documentclass{article}

\usepackage[preprint]{neurips_2026}
\usepackage{xspace}
\usepackage{cancel}

\usepackage{amsmath}
\usepackage{amsfonts}
\usepackage{nicefrac}

\usepackage{booktabs, multirow, array}
\usepackage{makecell}
\usepackage{adjustbox}
\usepackage{colortbl}

\usepackage{graphicx}
\usepackage{subcaption}
\usepackage{placeins}

\usepackage{hyperref}
\usepackage{url}

\usepackage{xcolor}
\definecolor{darkblue}{rgb}{0, 0, 0.5}
\definecolor{lightblue}{RGB}{173, 216, 230}
\definecolor{meanrow}{RGB}{232, 241, 251}

\usepackage{enumitem}

\usepackage{wrapfig} 
\usepackage{caption}

\usepackage{amssymb}

\newcommand{\cmark}{\textcolor{green!60!black}{$\checkmark$}}
\newcommand{\xmark}{\textcolor{red}{$\times$}}

\usepackage{algorithm}
\usepackage{algpseudocode}

\title{\underline{C}ontrastive C\underline{o}nceptor \underline{A}ctivation 
\underline{St}eering (COAST): Unlocking Vision-Language-Action Models through Hidden States}

\author{Miranda Muqing Miao\thanks{Equal contribution. Corresponding author: \texttt{miaom@seas.upenn.edu}.}$^{*}$\quad Subin Kim$^{*}$\quad Brandon Yang\quad Lyle Ungar \\ \\ University of Pennsylvania \\ }

\begin{document}
\maketitle

\begin{abstract}
Vision-Language-Action (VLA) models leverage powerful perceptual priors from web-scale Vision-Language Model (VLM) pre-training, yet they remain surprisingly brittle in practice, frequently failing at simple robotic tasks. To mitigate this, we propose \underline{C}ontrastive C\underline{o}nceptor \underline{A}ctivation \underline{St}eering (COAST).
COAST builds on the notion of a ``conceptor'', a linear operator that soft-projects data into the principal components of a target distribution. COAST uses conceptors to identify success-critical subspaces for a target robotic task from a few examples of success and failure rollouts. At inference time, it steers VLA latents into these identified success subspaces to improve task outcomes. Across three architecturally distinct neural policies (flow-matching VLA, autoregressive VLA, and Diffusion Policy), COAST improves absolute mean simulation and real-robot task success rate by over 20 and 40\% respectively. The activation subspace geometry reveals that failure modes share substantial structure across tasks while success representations remain largely task-specific. When tasks share similar failure modes, this structure enables previously fitted conceptors to improve performance on new tasks without refitting. Ultimately, our results suggest that current VLAs retain substantial task-relevant knowledge in their latent representations, and that the action expert's decoding bottleneck could be mitigated by steering its residual stream toward task-relevant subspaces. COAST provides a lightweight, training-free path to unlocking these latent capabilities by steering the model towards its own ``success'' distributions. 
\end{abstract}

\section{Introduction}

Today’s predominant paradigm for developing generalist robotic policies is to train Vision-Language-Action (VLA) models on multi-modal datasets at scale \citep{black2024pi0, intelligence2025pi05, bjorck2025gr00t}. VLAs have been shown to enable robots to interpret natural language commands in rich visual contexts and execute complex behaviors, exhibiting improved generalization~\citep{kim2024openvla, collaboration2024open}. Yet, it is challenging to pre-train on the exhaustive distribution of all possible deployment conditions, and in practice these policies frequently fail to execute down-stream tasks reliably~\citep{liu2026vls, kim2025contrastive}. Approaches such as post-training and fine-tuning~\citep{yuan2024policy, wagenmaker2025steering} and contrastive representation regularization~\citep{kim2025contrastive} have attempted to solve this problem. However, such remedies are costly and conceptually misaligned, as they attempt to relearn behaviors rather than trying to maximize the use of generalization ability that the model already learned from large-scale pre-training.

A natural point of comparison for VLAs is large language models (LLMs), where a growing body of work on \textit{activation steering} has shown that lightweight, closed-form interventions on intermediate representations can recover correct behavior without retraining. \citet{subramani2022extracting} and \citet{miao2026coral} show that latent steering vectors can be extracted from frozen model weights. \citet{turner2023activation} introduced activation addition using contrastive prompt pairs, \citet{panickssery2023steering} generalized this to contrastive activation addition across positive and negative examples. These methods steer along a single direction, which limits their reach when the relevant signal is distributed across multiple dimensions. \citet{zou2023representation} and \citet{postmus2024steering} broadened these single-direction methods into a subspace-level framework.


VLAs consist of a frozen vision-language backbone, which inherits rich perceptual and semantic representations from web-scale pretraining, and a separately trained action expert that decodes these representations into motor commands. We hypothesize that the action expert is the primary bottleneck. Its intermediate hidden states preserve much of the backbone's high-dimensional task understanding, but its final action outputs fail to fully exploit this information. If so, targeted interventions on the action expert's residual stream should be able to recover performance without modifying either the backbone or the expert's weights. Prior works have explored steering VLAs through their internal representations, but focus on the VLM backbone rather than the action expert. \citet{haon2025mechanistic} identify semantically meaningful neurons in the backbone's FFN layers through token projection, and \citet{swann2026sparse} and \citet{khan2025controlling} train sparse autoencoders on VLA hidden states to isolate useful features. However, these interventions steer individual backbone features and have not reported comprehensive improvements in overall task success rates. By contrast, COAST intervenes directly on the action expert's residual stream, where the decoding bottleneck resides.

Motivated by this, we introduce \textbf{COAST (\underline{C}ontrastive C\underline{o}nceptor \underline{A}ctivation \underline{St}eering)}, a plug-and-play framework that steers frozen VLA policies at inference time by applying a multiplicative gate to residual-stream activations.
Conceptors were originally developed for recurrent networks~\citep{jaeger2014controlling} and recently applied to LLM steering~\citep{postmus2024steering, postmus2025affine, triantafyllopoulos2026conceptorssemanticsteering}. We adapt this machinery to VLA action experts, using Boolean subspace algebra to construct contrastive conceptors from success and failure rollouts. The resulting operator isolates high-variance, outcome-critical directions while suppressing shared low-variance noise, and multiplicatively gates the residual stream during action generation.
Unlike prior VLA activation steering methods that require learned auxiliary models, gradient computation, or per-feature interventions, COAST operates through closed-form operators with efficient parameter selections. We validate COAST across flow-matching VLAs, autoregressive VLAs, and Diffusion Policy, and find that it consistently and significantly outperforms existing steering approaches across all three architecture families.

Our main contributions are:
\begin{itemize}[leftmargin=*]

\item \textbf{Lightweight Training-Free Policy Steering.} We apply conceptor-based subspace gating, previously used only in NLP, to VLA action experts for the first time. The resulting method, COAST, constructs a closed-form multiplicative gate from Boolean conceptor algebra over success and failure rollouts, improving mean simulation and real-robot task success rate by 20\% and 40\% respectively.

\item \textbf{Mechanistic Insight.} Upon analysis to discover the sources of these large gains, we find that the eigenvalue structure of VLA action experts is well-suited for conceptor-based steering. Outcome-relevant representations occupy a low-rank but not rank-one subspace, explaining why conceptors outperform single-direction additive methods. We further show that COAST reshapes latent-space activations toward the success subspace.

\item \textbf{Plug-And-Play Improvements.} We validate COAST across three architecturally distinct robot policy backbones (flow-matching VLAs, autoregressive VLAs, and Diffusion Policy), three simulation benchmarks of increasing complexity, cross-task conceptor transfer, and real-robot deployment, establishing that the method and the geometric structure it exploits are not specific to any single architecture, benchmark, task, or environment.

\end{itemize}
\section{Related Work}
\label{sec:related}

\paragraph{Inference-time steering of robot policies.}

Recent work adapts frozen VLA policies at deployment through reward-guided steering~\citep{liu2026vls}, latent substitution with auxiliary world models~\citep{das2025lae}, or VLM-based planning and verification~\citep{liu2026vls, song2026omniguide, shah2025liten, wu2025dowhatyousay}. Other approaches instead modify the policy itself by contrastively updating representations~\citep{kim2025contrastive}, training explicit steering interfaces~\citep{chen2026steerable}, or selectively fine-tuning model components~\citep{mitra2025robotic}. 
Unlike these approaches, COAST does not require external foundation models, runtime search, or gradient-based optimization, operating instead via a closed-form linear intervention. We compare COAST against internal baselines, while still showing that it surpasses gradient-update techniques like SFT in task performance (Sec.~\ref{sec:main_results}).
Please refer to App.~\ref{app:appendix_related_works} for a more in-depth analysis.

\paragraph{Reward-based policy improvement.}
A separate line of work improves VLA and diffusion policies through reinforcement learning rather than activation steering. \citet{wagenmaker2025steering} optimize a reward function in the latent noise space of a diffusion policy, while \citet{pfrommer2025rl} train flow-matching policies via reward-weighted regression and group relative policy optimization. These methods require an explicit reward signal and gradient-based optimization of policy parameters or auxiliary modules. In contrast, COAST leaves the policy weights frozen and requires no reward beyond binary success/failure labels used to partition rollouts. 


\paragraph{Conceptors.}
Conceptors are closed-form multiplicative gates on activation subspaces,
introduced by \citet{jaeger2014controlling} for recurrent networks. We introduce their technical details in Section~\ref{sec:conceptors_filter}.
\citet{postmus2024steering} applied them to LLM steering additively as
steering vectors, and \citet{postmus2025affine} extended the framework to optimal affine steering, showing that conceptor-based intervention  outperforms additive methods and that Boolean composition of
conceptors outperforms naive vector combination. In concurrent work with overlapping authors, \citet{triantafyllopoulos2026conceptorssemanticsteering} apply contrastive conceptors to steer LLM semantic outputs, confirming the effectiveness of Boolean conceptor algebra for NLP tasks.
We extend these insights to VLAs, applying
contrastive subspace construction via Boolean negation and multiplicative residual-stream gating to VLA action experts. To our knowledge, this is the first use of conceptors for embodied policy steering.
\section{Activation Steering with Contrastive Conceptor}
\label{sec:method}

\begin{figure*}[t!]
    \centering
    \includegraphics[width=\textwidth]{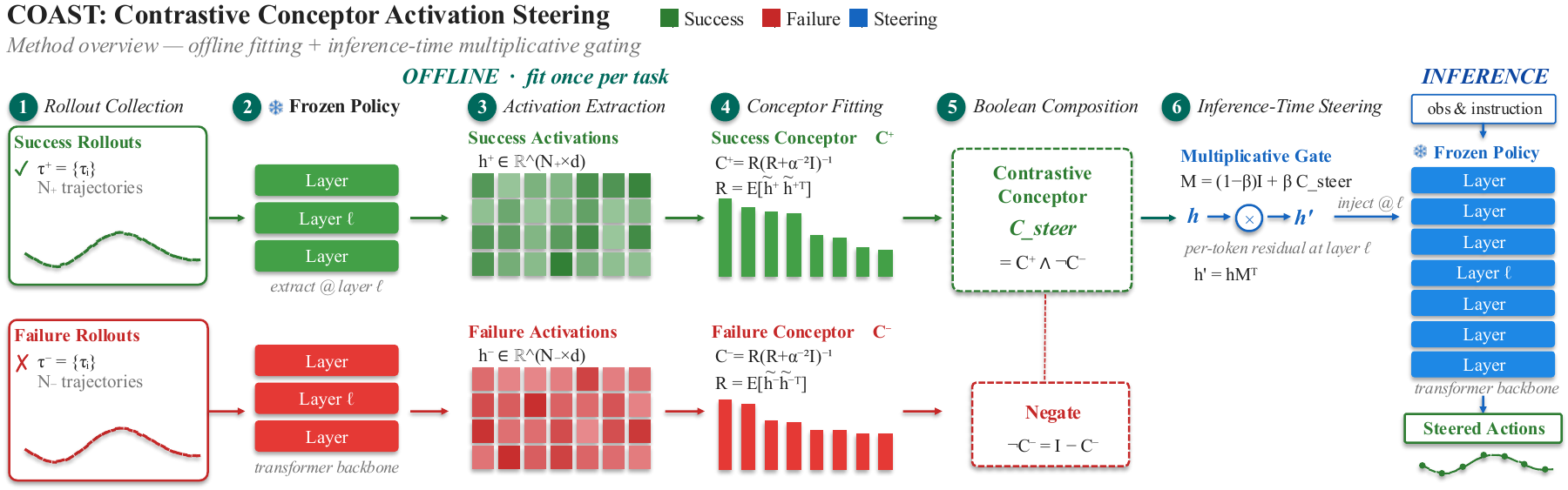}
 \caption{\textbf{Overview of COAST.}
    COAST steers a frozen robot policy at inference time by multiplicatively gating its residual stream with a contrastive conceptor fit from rollout activations.
    \textit{Offline,} success and failure rollouts yield layer-$\ell$ activations, from which closed-form conceptors $C^{+}$ and $C^{-}$ are fit and composed via Boolean subspace algebra into $C_{\text{steer}} = C^{+} \wedge \neg C^{-}$.
    \textit{Inference,} the residual $h$ at layer $\ell$ is gated by $M = (1-\beta)I + \beta\, C_{\text{steer}}$ to produce $h' = hM^\top$, which flows through the rest of the frozen model. No gradient computation or weight modification is required.}
    \label{fig:coast_chart}
    \vspace{-5pt}
\end{figure*}

COAST builds on existing conceptor algebra~\citep{jaeger2014controlling, postmus2024steering} and adapts it to VLA action experts. The key observation motivating this adaptation is that a VLA's hidden states during task execution contain both task-critical and task-irrelevant features. When comparing successes and failures, the irrelevant features remain largely unchanged and naturally exhibit low variance within collected samples, while task-relevant and outcome-critical states tend to manifest high variance. Conceptors are well-suited for this setting because they capture these highly variable and discriminative directions by suppressing shared low-variance noise, effectively steering the model toward the representations essential for task completion.

\subsection{Background: Contrastive Conceptors For Identifying Task-Relevant State Space}
\label{sec:conceptors_filter}
\textbf{Building Conceptors.}
A conceptor matrix $C$ is a positive semi-definite linear operator that captures the principal directions and variances of a set of neural activations~\citep{jaeger2014controlling}. Given an activation matrix $X \in \mathbb{R}^{N \times d}$ whose rows are individual hidden states, let $\tilde{X}$ denote the mean-centered matrix with rows $\tilde{x}_i = x_i - \bar{x}$, where $\bar{x} = \frac{1}{N}\sum_i x_i$. $C$ minimizes a regularized reconstruction objective
\begin{equation}
    \min_C \frac{1}{N}\|\tilde{X} - \tilde{X}C\|_F^2 \;+\; \alpha^{-2}\|C\|_F^2.
    \label{eq:conceptor_objective}
\end{equation}
The first term ensures $C$ preserves the variance structure of the activations, while the second, controlled by the aperture $\alpha > 0$, penalizes complexity. We follow the convention of \citet{jaeger2014controlling}, where larger $\alpha$ yields a less regularized conceptor that faithfully reconstructs the data, while smaller $\alpha$ favors a more compressed, generalized representation. The closed-form solution is $C = R(R + \alpha^{-2}I)^{-1}$, where $R = \tilde{X}^\top \tilde{X} / N$ is the covariance matrix. Crucially, this allows us to compute C directly via simple matrix multiplication and inversion, bypassing the need for gradient descent.

\textbf{Conceptors as filters.} Applying a conceptor $C$ to a new sample $x$ to produce $\tilde{x}=x^TC$ can be seen as ``filtering'' the sample. The eigenvalue structure of $C$ makes this filtering behavior transparent. If $R$ has eigenvalues $\{\lambda_i\}$, the corresponding eigenvalues of $C$ are
\begin{equation}
    \mu_i = \frac{\lambda_i}{\lambda_i + \alpha^{-2}}.
    \label{eq:conceptor_eigenvalues}
\end{equation}
Directions of high variance ($\lambda_i \gg \alpha^{-2}$) receive $\mu_i \approx 1$ and pass through nearly unchanged, while low-variance directions ($\lambda_i \ll \alpha^{-2}$) are suppressed toward zero. The conceptor thus acts as a soft, variance-adaptive filter on the activation space. This way, a conceptor scales each direction differently based on the discovered subspace structure.

\textbf{Contrastive Conceptor.}
A conceptor as described above captures the principal subspace of the activation distribution, but effective steering requires discriminating success from failure distributions. This can be captured by first computing separate conceptors $C_{\text{success}}$ and $C_{\text{failure}}$ for success and failure samples following Equation~\ref{eq:conceptor_objective}, and then appropriately \textit{composing} them.  

Conceptors compose conveniently under a soft Boolean algebra~\citep{jaeger2014controlling, postmus2024steering}. In particular, NOT is defined as $\neg C = I - C$, and AND is defined as
\begin{equation}
    A \wedge B \;=\; \left(A^{-1} + B^{-1} - I\right)^{-1}.
    \label{eq:and_operator}
\end{equation}

following the canonical conceptor algebra of \citet{jaeger2014controlling} and \citet{postmus2024steering}. Intuitively, AND computes the soft intersection of two subspaces: directions that both operands retain receive high weight, while directions retained by only one are attenuated. Thus, a ``contrastive conceptor'' that separates success and failure states may be composed as:

\begin{equation}
    C_{\text{steer}} = C_{\text{success}} \wedge \neg C_{\text{failure}}.
    \label{eq:contrastive_conceptor}
\end{equation}
The AND-NOT operation retains directions present in the success subspace and absent from the failure subspace. Shared directions, which carry outcome-irrelevant computation, are suppressed. What remains is the thin residual subspace that differentiates the two outcomes.

\subsection{Steering VLAs with Contrastive Conceptors}
\label{sec:conceptors_steering}

At inference time, we use the contrastive conceptor as a soft projection that reweights the action expert’s activations without overwriting them, giving direct control over the model’s latent state. This gently steers hidden states toward the success-associated subspace while suppressing directions linked to failure. Because the intervention acts only on a small, outcome-critical subspace, it is more efficient and less disruptive than manipulating the full representation.

\textbf{Activation extraction.}
We focus on activations rather than other residual hidden states because they more cleanly capture the network’s task-relevant dynamics as a structured, low-dimensional manifold~\citep{jaeger2014controlling}.
The VLA policies we evaluate share a common backbone-expert architecture in which a frozen vision-language model produces a conditioning representation and a trainable action expert generates actions through either flow-matching denoising~\citep{intelligence2025pi05, bjorck2025gr00t} or autoregressive token prediction~\citep{pertsch2025fast}. Diffusion Policy~\citep{chi2025diffusion} shares the flow-matching denoising architecture. We extract activations from the action expert's residual stream at a single layer $\ell$, mean-pooling across action tokens to obtain one vector $h \in \mathbb{R}^d$ per denoising or autoregressive step. We roll out $N$ samples, where success and failure are determined either by simulation or by real world outcomes. Stacking $N$ such vectors yields the matrix $X = [h_1; \ldots; h_N] \in \mathbb{R}^{N \times d}$ from which the conceptors in Section~\ref{sec:conceptors_filter} are constructed. Details of the extraction procedure for each model family are in Appendix~\ref{app:activation-extraction}.

\textbf{Multiplicative Gating.}
At inference time, we apply the contrastive conceptor as a multiplicative gate on the residual stream. The gating matrix is

\begin{equation}
    M = (1 - \beta)\,I + \beta\, C_{\text{steer}}.
    \label{eq:soft_projection}
\end{equation}
where $\beta \in [0,1]$ controls the steering strength. Each residual activation $h$ at layer $\ell$ is replaced by $h' = hM^\top$ before continuing through the remaining layers of the frozen model. The interpolation with the identity ensures that the gate acts as a soft projection. At $\beta = 0$ the forward pass is unmodified, at $\beta = 1$ the full conceptor is applied, and intermediate values blend the original and steered activations. The interpolation with the identity is important because even though $C_{\text{steer}}$ is already a soft projection (its eigenvalues are fractional rather than binary), applying it at full strength can still be too aggressive. The $\beta$ parameter provides an additional degree of control, allowing the gate to attenuate directions according to their discriminative value while preserving enough of the original activation to keep the policy's downstream computation stable.

\textbf{Steering strategies.} We evaluate three variants of conceptor steering. \textit{Global} steering fits a single contrastive conceptor $C_{\text{steer}} = C_{\text{success}} \wedge \neg C_{\text{failure}}$ from activations pooled across all denoising steps and applies the same gate at every step. This is the simplest strategy and assumes the discriminative subspace is stable throughout the denoising process. \textit{Per-step} steering instead fits a separate contrastive conceptor for each denoising step from activations collected at that step alone, allowing the gate to track shifts in the discriminative subspace as the action is progressively refined from noise to a clean action chunk. \textit{Positive-only} steering sets $C_{\text{steer}} = C_{\text{success}}$, omitting the failure subtraction entirely. This variant requires only successful demonstrations and serves as a test of whether the contrastive operation is necessary or whether projecting toward the success subspace alone suffices.

We also introduce a highly efficient hyperparameter selection heuristic by identifying the optimal layer through conceptor quota and narrowing the aperture range via success-failure overlap. This recovers 93\% of oracle performance while evaluating under 3-8\% of the full configuration grid (App.~\ref{sec:param-opt}). Additionally, we test the latency of COAST and find it to be minimal and in-line with existing steering methods (App.~\ref{app:overhead}).



\section{Experiments}
\label{sec:experiments}
 
In our experiments, we aim to evaluate the following questions. (1) \textbf{Performance.} Does COAST reliably improve task success across diverse VLA models and benchmarks?
(2) \textbf{Mechanism \& Interpretability.} How does conceptor steering geometrically reshape internal activations? (3) \textbf{Shared Failure Geometry.} Do tasks share failure-mode structure, and if so, can a conceptor fitted on one task improve a different task without refitting?
 
\subsection{Experimental Setup}
\label{sec:results}
 
\begin{figure*}[t!]
    \centering
    \noindent\centerline{
    \includegraphics[width=1\textwidth]{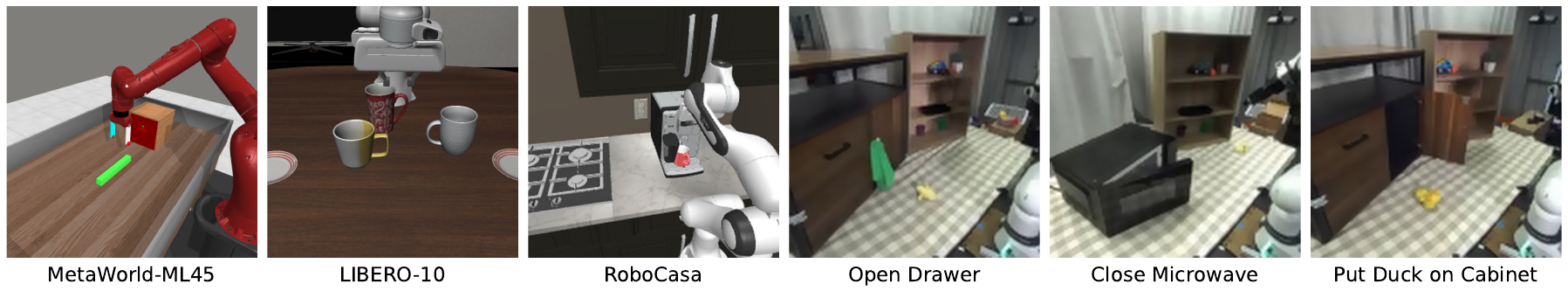}
    }
  \caption{\textbf{Overview of Environment Setup.}
     We evaluate on three simulation benchmarks of increasing difficulty, MetaWorld ML45~\citep{yu2020meta},
LIBERO-10~\citep{liu2023libero}, and a select RoboCasa~\citep{nasiriany2024robocasa}
subset, plus three real-robot tasks on the DROID platform.}
    \label{fig:env_setup}
\end{figure*}
 
\textbf{Environments.} We evaluate across three simulation benchmarks and three real-world tasks chosen to span a range of task complexity, from short-horizon single-object manipulation to long-horizon multi-step reasoning in visually diverse scenes. Further setups can be found in App.~\ref{app:rollout-protocol}.
\begin{itemize}[leftmargin=*]
    \item \textbf{MetaWorld ML45}~\citep{yu2020meta} comprises 45 tabletop robotic manipulation tasks spanning a broad spectrum of motor skills, each with randomized goal positions per episode.
 
    \item \textbf{LIBERO-10}~\citep{liu2023libero} consists of 10 long-horizon manipulation tasks in tabletop kitchen scenes, each requiring multi-step object interactions such as picking, placing, and opening across semantically diverse objects including containers, utensils, and appliances.
 
    \item \textbf{RoboCasa}~\citep{nasiriany2024robocasa} covers 7 atomic-seen tasks. Every episode re-samples the kitchen layout, object instances, placements, textures, and lighting, producing a combinatorially large state space that demands strong generalization across visually and physically diverse configurations unseen during training.
 
    \item \textbf{Real-world validation.} We use the DROID platform~\citep{khazatsky2024droid}, evaluated on three tasks: \texttt{Open Drawer}, \texttt{Close Microwave}, and \texttt{Put Duck in Cabinet} over 15 independent trials per task. 
    Success is defined strictly by task completion: drawer fully extended, microwave door latched, or duck inside the cabinet. Further details can be found in App.~\ref{app:real-robot-setup}.
    
\end{itemize}

\paragraph{Evaluated Policies.} We evaluate several backbone policies across our environments: 

\(\pi_{0.5}\)~\citep{intelligence2025pi05} (Metaworld, LIBERO, Robocasa), \(\pi_0\)-FAST~\citep{pertsch2025fast} (Metaworld, LIBERO), GR00T N1.5~\citep{bjorck2025gr00t} (Robocasa) and Diffusion Policy~\citep{chi2025diffusion} (Robocasa).
By showing that our method also works with DP, we demonstrate our method's scalability to the diffusion/flow-matching family of policies other than VLA models. All VLA models are also evaluated on real-world tasks. VLA models achieve very high success rates on MetaWorld and LIBERO when fully trained, so we use early fine-tuning checkpoints that produce the mix of successes and failures needed for contrastive fitting. For RoboCasa, the fully trained checkpoints for both GR00T N1.5 and Diffusion Policy exhibit sufficient failure rates, so no early stopping is required. We show that our method is robust across different checkpoints and parameters in App.~\ref{app:dp-robocasa-per-epoch}. 


\paragraph{Baselines.} 

We evaluate COAST against two activation steering baselines and one Supervised Fine-Tuning (SFT) baseline. We compare against Contrastive Activation Addition (CAA) \citep{panickssery2023steering} to represent single-direction additive steering, and Sparse Autoencoder (SAE) steering \citep{khan2025controlling} to represent feature-based steering. Furthermore, we evaluate SFT with LoRA~\citep{hu2022lora} using the exact same 15 trajectories utilized for COAST.
Implementation details for each baseline can be found in App. ~\ref{app:caa}, App. ~\ref{app:sae}, and App.~\ref{app:filtered-bc} respectively. 
Finally, aperture $\alpha$ and steering strength $\beta$ are optimized on gathered rollouts (Sec.~\ref{sec:param-opt}). 

\paragraph{Evaluation Protocol.} For each task, we use a strict train/test split: 15 training rollouts to extract conceptors and select hyperparameters, and 30 held-out test rollouts with entirely unseen environment configurations for final evaluation (see App.~\ref{app:rollout-protocol} for details).

\definecolor{lightblue}{RGB}{230, 240, 255}                                                                                                              
                                                                                                                                                           
  \begin{table*}[t!]
  \centering                                                                                                                                               
  \renewcommand{\arraystretch}{1.1}                                   

  \noindent\centerline{%
  \resizebox{\textwidth}{!}{%
  \begin{tabular}{l | cccc | ccc | cccc | ccc | ccc | ccc}            
  \toprule                                                                                                                                                 
  \multicolumn{21}{c}{\textbf{RoboCasa}} \\
  \midrule                                                                                                                                                 
  \multirow{3}{*}{\textbf{Task}}                                      
    & \multicolumn{7}{c|}{\textbf{$\pi_{0.5}$}}                                                                                                            
    & \multicolumn{7}{c|}{\textbf{GR00T N1.5}}                                                                                                             
    & \multicolumn{6}{c}{\textbf{Diffusion Policy}} \\                                                                                                     
  \cmidrule(lr){2-8} \cmidrule(lr){9-15} \cmidrule(lr){16-21}                                                                                              
    & \multicolumn{4}{c|}{\textit{Baselines}} & \multicolumn{3}{c|}{\textit{COAST}}                                                                        
    & \multicolumn{4}{c|}{\textit{Baselines}} & \multicolumn{3}{c|}{\textit{COAST}}                                                                        
    & \multicolumn{3}{c|}{\textit{Baselines}} & \multicolumn{3}{c}{\textit{COAST}} \\                                                                      
  \cmidrule(lr){2-5} \cmidrule(lr){6-8} \cmidrule(lr){9-12} \cmidrule(lr){13-15} \cmidrule(lr){16-18} \cmidrule(lr){19-21}                                 
    & Base & +SFT & +SAE & +CAA & +Glob. & +Per. & +Pos.                                                                                                   
    & Base & +SFT & +SAE & +CAA & +Glob. & +Per. & +Pos.                                                                                                   
    & Base & +SAE & +CAA & +Glob. & +Per. & +Pos. \\                                                                                                       
  \midrule                                                                                                                                                 
  Close Fridge & 0.20 & 0.03 & \textbf{0.47} & 0.27 & \textbf{0.47} & \underline{0.40} & \underline{0.40}                                                  
               & 0.67 & 0.43 & 0.73 & 0.80 & \underline{0.93} & \textbf{1.00} & 0.87                                                                       
               & 0.43 & 0.53 & 0.53 & \underline{0.56} & \textbf{0.60} & 0.50 \\                                                                           
  Coffee Mug   & 0.13 & 0.23 & \underline{0.33} & 0.13 & \underline{0.33} & \underline{0.33} & \textbf{0.53}                                               
               & 0.20 & 0.10 & 0.40 & 0.13 & \underline{0.27} & \textbf{0.33} & \textbf{0.33}                                                              
               & 0.13 & \underline{0.20} & 0.16 & \textbf{0.23} & 0.16 & 0.16 \\                                                                           
  Open Drawer  & 0.53 & 0.47 & \underline{0.60} & 0.53 & \textbf{0.67} & 0.53 & \textbf{0.67}                                                              
               & 0.53 & 0.57 & 0.67 & \underline{0.73} & \textbf{0.80} & 0.53 & 0.67                                                                       
               & 0.10 & 0.20 & 0.23 & \textbf{0.33} & \underline{0.30} & 0.23 \\                                                                           
  Stand Mixer  & 0.67 & 0.30 & 0.53 & 0.53 & \underline{0.73} & \textbf{0.80} & 0.53                                                                       
               & 0.60 & 0.77 & 0.73 & \underline{0.80} & \underline{0.80} & \underline{0.80} & \textbf{0.93}                                               
               & \underline{0.63} & \textbf{0.83} & \underline{0.63} & \textbf{0.83} & \textbf{0.83} & \textbf{0.83} \\                                    
  PP Cabinet   & 0.60 & 0.50 & \underline{0.67} & \underline{0.67} & \textbf{0.74} & \textbf{0.74} & \underline{0.67}                                      
               & \underline{0.73} & 0.63 & 0.73 & 0.53 & \textbf{0.80} & \textbf{0.80} & \underline{0.73}                                                  
               & 0.30 & \underline{0.53} & 0.40 & 0.46 & \textbf{0.56} & 0.46 \\                                                                           
  PP Stove     & 0.33 & 0.43 & 0.53 & 0.53 & 0.53 & \textbf{0.67} & \underline{0.60}                                                                       
               & 0.73 & 0.60 & \textbf{0.87} & \textbf{0.87} & \textbf{0.87} & \underline{0.75} & \textbf{0.87}                                            
               & \underline{0.10} & \underline{0.10} & \underline{0.10} & \textbf{0.13} & \textbf{0.13} & \underline{0.10} \\                              
  Kettle       & 0.33 & 0.20 & 0.27 & 0.27 & \underline{0.40} & \underline{0.40} & \textbf{0.53}                                                           
               & 0.67 & 0.37 & 0.73 & 0.47 & \textbf{0.80} & \underline{0.73} & \underline{0.73}                                                           
               & 0.56 & 0.56 & 0.50 & \underline{0.60} & \textbf{0.66} & 0.46 \\                                                                           
  \midrule                                                                                                                                                 
  Mean                                                                                                                                                     
    & 0.40 & 0.31 & 0.49 & 0.42 & \cellcolor{lightblue}\underline{0.55} & \cellcolor{lightblue}\underline{0.55} & \cellcolor{lightblue}\textbf{0.56}       
    & 0.59 & 0.50 & 0.69 & 0.62 & \cellcolor{lightblue}\textbf{0.75} & \cellcolor{lightblue}0.71 & \cellcolor{lightblue}\underline{0.73}                   
    & 0.32 & 0.42 & 0.36 & \cellcolor{lightblue}\underline{0.45} & \cellcolor{lightblue}\textbf{0.46} & \cellcolor{lightblue}0.39 \\                       
  $\Delta$                                                                                                                                                 
    & -- & $-0.09$ & $+0.09$ & $+0.02$ & \cellcolor{lightblue}+0.15 & \cellcolor{lightblue}+0.15 & \cellcolor{lightblue}+0.16                              
    & -- & -0.09 & +0.10 & +0.03 & \cellcolor{lightblue}+0.16 & \cellcolor{lightblue}+0.12 & \cellcolor{lightblue}+0.14                                    
    & -- & +0.10 & +0.04 & \cellcolor{lightblue}+0.13 & \cellcolor{lightblue}+0.14 & \cellcolor{lightblue}+0.07 \\                                         
  $p_t$                                                                                                                                                    
    & -- & $.195$ & $.172$ & $.642$ & \cellcolor{lightblue}$.002^{**}$ & \cellcolor{lightblue}$.009^{**}$ & \cellcolor{lightblue}$.044^{*}$                      
    & -- & $0.89$ & $.006^{**}$ & $.690$ & \cellcolor{lightblue}$.002^{**}$ & \cellcolor{lightblue}$.039^{*}$ & \cellcolor{lightblue}$.011^{*}$                    
    & -- & $.025^{*}$ & $.151$ & \cellcolor{lightblue}$.004^{**}$ & \cellcolor{lightblue}$.006^{**}$ & \cellcolor{lightblue}$.123$ \\                            
  $p_z$                                                                                                                                                    
    & -- & $.101$ & $.211$ & $.779$ & \cellcolor{lightblue}$.014^{*}$ & \cellcolor{lightblue}$.014^{*}$ & \cellcolor{lightblue}$.009^{**}$                       
    & -- & $0.92$ & $.113$ & $.672$ & \cellcolor{lightblue}$.006^{**}$ & \cellcolor{lightblue}$.046^{*}$ & \cellcolor{lightblue}$.014^{*}$                         
    & -- & $.134$ & $.512$ & \cellcolor{lightblue}$.058$ & \cellcolor{lightblue}$.035^{*}$ & \cellcolor{lightblue}$.289$ \\                                      
  \bottomrule                                                                                                                                              
  \end{tabular}%
  }}                                                                                                                                                       
                                                                                                                                                           
  \vspace{0.6em}                                                                                                                                           
  
  \noindent                                                           
  \begin{minipage}[t]{0.74\textwidth}                                                                                                                      
  \centering                                                                                                                                               
  \vspace{0pt}
                                                                                                                                                           
  \resizebox{\linewidth}{!}{%
  \begin{tabular}{l | cccc | ccc | cccc | ccc}                        
  \toprule                                                                                                                                                 
  \multicolumn{15}{c}{\textbf{LIBERO-10}} \\
  \midrule                                                                                                                                                 
  \multirow{3}{*}{\textbf{Task}}                                      
    & \multicolumn{7}{c|}{\textbf{$\pi_{0.5}$}}                                                                                                            
    & \multicolumn{7}{c}{\textbf{$\pi_{0}$-FAST$^\dagger$}} \\                                                                                             
  \cmidrule(lr){2-8} \cmidrule(lr){9-15}                                                                                                                   
    & \multicolumn{4}{c|}{\textit{Baselines}} & \multicolumn{3}{c|}{\textit{COAST}}                                                                        
    & \multicolumn{4}{c|}{\textit{Baselines}} & \multicolumn{3}{c}{\textit{COAST}} \\                                                                      
  \cmidrule(lr){2-5} \cmidrule(lr){6-8} \cmidrule(lr){9-12} \cmidrule(lr){13-15}                                                                           
    & Base & +SFT & +SAE & +CAA                                                                                                                            
    & +Glob. & +Per. & +Pos.                                                                                                                               
    & Base & +SFT & +SAE & +CAA                                                                                                                            
    & +Glob. & +Per. & +Pos. \\                                       
  \midrule                                                                                                                                                 
  Stove+Moka  & 0.53 & 0.63 & 0.67 & 0.67                                                                                                         
       & \textbf{0.93} & \underline{0.87} & 0.80                                                                                                           
       & 0.80 & 0.73 & 0.80 & 0.73                                                                                                                         
       & \textbf{1.00} & \underline{0.87} & 0.80 \\                                                                                                        
  Bowl+Drawer  & 0.40 & 0.20 & \underline{0.80} & 0.47                                                                                                             
       & 0.73 & \underline{0.80} & \textbf{0.87}                                                                                                           
       & 0.67 & 0.60 & \underline{0.87} & 0.73                                                                                                             
       & \textbf{0.93} & \textbf{0.93} & 0.80 \\                                                                                                           
  Mug+Micro  & 0.13 & 0.10 & 0.20 & 0.13                                                                                                         
       & \underline{0.40} & \textbf{0.60} & 0.33                                                                                                           
       & 0.47 & 0.47 & 0.67 & \textbf{0.80}                                                                                                                
       & \underline{0.73} & \textbf{0.80} & \underline{0.73} \\                                                                                            
  Two Mokas  & 0.20 & 0.07 & 0.13 & 0.07                                                                                                         
       & \underline{0.47} & \textbf{0.53} & 0.13                                                                                                           
       & 0.33 & 0.10 & 0.33 & 0.27                                                                                                                         
       & \textbf{0.53} & \textbf{0.53} & \underline{0.40} \\                                                                                               
  Soup+Cheese  & 0.80 & 0.43 & 0.73 & 0.80                                                                                                         
       & \underline{0.93} & \textbf{1.00} & \underline{0.93}                                                                                               
       & 0.67 & 0.73 & 0.80 & 0.80                                                                                                                         
       & \underline{0.87} & \underline{0.87} & \textbf{1.00} \\                                                                                            
  Soup+Tomato & 0.40 & 0.53 & \textbf{0.80} & 0.53                                                                                                                
       & \textbf{0.80} & \textbf{0.80} & \underline{0.60}                                                                                                  
       & 0.67 & 0.77 & \textbf{0.93} & 0.80                                                                                                                
       & \underline{0.87} & \textbf{0.93} & 0.73 \\                                                                                                        
  Cheese+Butter & 0.60 & 0.67 & 0.80 & 0.60                                                                                                         
       & \textbf{0.93} & \textbf{0.93} & \underline{0.87}                                                                                                  
       & \textbf{1.00} & \underline{0.90} & -- & --                                                                                                        
       & -- & -- & \textbf{1.00} \\                                                                                                                        
  Mugs+Plates  & 0.07 & 0.13 & \underline{0.33} & 0.07                                                                                                             
       & \textbf{0.60} & \textbf{0.60} & 0.20                                                                                                              
       & 0.60 & 0.73 & \textbf{0.87} & \underline{0.80}                                                                                                    
       & \textbf{0.87} & \textbf{0.87} & \underline{0.80} \\                                                                                               
  Mug+Choc  & 0.53 & 0.57 & \underline{0.73} & 0.60                                                                                                             
       & \textbf{0.87} & \textbf{0.87} & 0.67                                                                                                              
       & 0.73 & 0.77 & \underline{0.87} & 0.67                                                                                                             
       & \textbf{0.93} & \textbf{0.93} & \underline{0.87} \\                                                                                               
  Book+Caddy  & 0.67 & 0.70 & 0.87 & 0.80                                                                                                         
       & \underline{0.93} & \textbf{1.00} & \underline{0.93}                                                                                               
       & 0.60 & 0.43 & 0.73 & \underline{0.80}                                                                                                             
       & \textbf{0.87} & \textbf{0.87} & 0.73 \\                                                                                                           
  \midrule                                                                                                                                                 
  Mean                                                                
    & 0.43 & 0.40 & 0.61 & 0.47                                                                                                                            
    & \cellcolor{lightblue}\underline{0.76} & \cellcolor{lightblue}\textbf{0.80} & \cellcolor{lightblue}0.63                                               
    & 0.65 & 0.62 & 0.76 & 0.71                                                                                                                            
    & \cellcolor{lightblue}\textbf{0.84} & \cellcolor{lightblue}\textbf{0.84} & \cellcolor{lightblue}\underline{0.79} \\                                   
  $\Delta$                                                                                                                                                 
    & -- & $-0.03$ & $+0.17$ & $+0.04$                                                                                                                     
    & \cellcolor{lightblue}+0.33 & \cellcolor{lightblue}+0.37 & \cellcolor{lightblue}+0.20                                                                 
    & -- & $-0.03$ & +0.15 & +0.10                                                                                                                         
    & \cellcolor{lightblue}+0.23 & \cellcolor{lightblue}+0.23 & \cellcolor{lightblue}+0.13 \\
  $p_t$                                                                                                                                                    
    & -- & $.562$ & $.009^{**}$ & $.157$                              
    & \cellcolor{lightblue}$<.001^{***}$ & \cellcolor{lightblue}$<.001^{***}$ & \cellcolor{lightblue}$.001^{**}$                                           
    & -- & $.425$ & $.002^{**}$ & $.074$                                                                                                                   
    & \cellcolor{lightblue}$<.001^{***}$ & \cellcolor{lightblue}$<.001^{***}$ & \cellcolor{lightblue}$.004^{**}$ \\                                        
  $p_z$                                                                                                                                                    
    & -- & $.118$ & $.003^{**}$ & $.487$                              
    & \cellcolor{lightblue}$<.001^{***}$ & \cellcolor{lightblue}$<.001^{***}$ & \cellcolor{lightblue}$<.001^{***}$
    & -- & 0.672 & $.009^{**}$ & $.094$
    & \cellcolor{lightblue}$<.001^{***}$ & \cellcolor{lightblue}$<.001^{***}$ & \cellcolor{lightblue}$.009^{**}$ \\
  \bottomrule
  \end{tabular}%
  }

  \vspace{0.6em}

  \resizebox{\linewidth}{!}{%
  \begin{tabular}{l | cccc | ccc | cccc | ccc}
  \toprule
  \multicolumn{15}{c}{\textbf{MetaWorld ML45}} \\
  \midrule
  \multirow{3}{*}{\textbf{Task}}
    & \multicolumn{7}{c|}{\textbf{$\pi_{0.5}$}}
    & \multicolumn{7}{c}{\textbf{$\pi_{0}$-FAST}} \\
  \cmidrule(lr){2-8} \cmidrule(lr){9-15}
    & \multicolumn{4}{c|}{\textit{Baselines}} & \multicolumn{3}{c|}{\textit{COAST}}
    & \multicolumn{4}{c|}{\textit{Baselines}} & \multicolumn{3}{c}{\textit{COAST}} \\
  \cmidrule(lr){2-5} \cmidrule(lr){6-8} \cmidrule(lr){9-12} \cmidrule(lr){13-15}
    & Base & +SFT & +SAE & +CAA
    & +Glob. & +Per. & +Pos.
    & Base & +SFT & +SAE & +CAA
    & +Glob. & +Per. & +Pos. \\
  \midrule
  coffee-push      & 0.80 & 0.63 & \textbf{1.00} & \underline{0.87}
                   & \textbf{1.00} & \textbf{1.00} & \textbf{1.00}
                   & 0.88 & 0.87 & 0.88 & \textbf{1.00}
                   & \underline{0.94} & \underline{0.94} & \underline{0.94} \\                                                                             
  push             & \underline{0.93} & 0.80 & \underline{0.93} & \underline{0.93}                                                                         
                   & \textbf{1.00} & \textbf{1.00} & \underline{0.93}                                                                                      
                   & 0.75 & 0.83 & \underline{0.88} & 0.75                                                                                                 
                   & \textbf{0.94} & \underline{0.88} & 0.75 \\                                                                                            
  pick-place       & \underline{0.87} & \underline{0.87} & 0.80 & \underline{0.87}                                                                         
                   & \textbf{1.00} & \textbf{1.00} & 0.80                                                                                                  
                   & 0.69 & 0.67 & 0.62 & 0.75                                                                                                             
                   & \textbf{0.88} & \textbf{0.88} & \underline{0.81} \\                                                                                   
  plate-slide-back & 0.60 & 0.37 & 0.53 & 0.53                        
                   & \textbf{0.93} & \textbf{0.93} & \underline{0.73}                                                                                      
                   & 0.88 & \underline{0.93} & 0.56 & 0.75                                                                                                 
                   & 0.88 & 0.88 & \textbf{0.94} \\                                                                                                        
  faucet-close     & 0.80 & 0.67 & \textbf{1.00} & \underline{0.93}                                                                                        
                   & \textbf{1.00} & \underline{0.93} & \underline{0.93}                                                                                   
                   & 0.44 & 0.80 & \textbf{1.00} & 0.50                                                                                                    
                   & 0.69 & 0.69 & \underline{0.81} \\                                                                                                     
  pick-place-wall  & 0.20 & 0.23 & 0.33 & 0.40                                                                                                             
                   & \underline{0.87} & \textbf{1.00} & 0.47                                                                                               
                   & 0.69 & 0.50 & \textbf{0.88} & 0.69               
                   & \underline{0.81} & 0.75 & 0.75 \\                                                                                                     
  reach            & \underline{0.93} & 0.73 & \underline{0.93} & \textbf{1.00}                                                                            
                   & \underline{0.93} & \underline{0.93} & \textbf{1.00}                                                                                   
                   & \textbf{0.81} & \underline{0.77} & 0.69 & \textbf{0.81}                                                                               
                   & \textbf{0.81} & \textbf{0.81} & \textbf{0.81} \\                                                                                      
  coffee-pull      & \underline{0.93} & 0.67 & \textbf{1.00} & \textbf{1.00}                                                                               
                   & \underline{0.93} & \underline{0.93} & \textbf{1.00}                                                                                   
                   & \underline{0.62} & 0.60 & \underline{0.62} & 0.50                                                                                     
                   & \textbf{0.69} & \underline{0.62} & \textbf{0.69} \\                                                                                   
  disassemble      & 0.60 & 0.10 & 0.80 & 0.73                                                                                                             
                   & \textbf{0.93} & \textbf{0.93} & \underline{0.87}                                                                                      
                   & 0.56 & 0.70 & \textbf{0.81} & 0.56                                                                                                    
                   & 0.69 & 0.56 & \underline{0.75} \\                                                                                                     
  stick-push       & 0.20 & 0.20 & 0.40 & 0.60                                                                                                             
                   & \textbf{0.80} & \underline{0.73} & 0.40                                                                                               
                   & 0.81 & \underline{0.90} & \textbf{1.00} & 0.81                                                                                        
                   & 0.88 & 0.88 & 0.88 \\                                                                                                                 
  \midrule                                                                                                                                                 
  Mean                                                                                                                                                     
    & 0.69 & 0.53 & 0.77 & 0.79                                                                                                                            
    & \cellcolor{lightblue}\textbf{0.94} & \cellcolor{lightblue}\textbf{0.94} & \cellcolor{lightblue}\underline{0.81}                                      
    & 0.71 & 0.76 & 0.79 & 0.71                                                                                                                            
    & \cellcolor{lightblue}\textbf{0.82} & \cellcolor{lightblue}0.79 & \cellcolor{lightblue}\underline{0.81} \\                                            
  $\Delta$                                                                                                                                                 
    & -- & $-0.16$ & $+0.08$ & +0.10                                  
    & \cellcolor{lightblue}+0.25 & \cellcolor{lightblue}+0.25 & \cellcolor{lightblue}+0.13                                                                 
    & -- & +0.05 & +0.08 & +0.00                                                                                                                           
    & \cellcolor{lightblue}+0.11 & \cellcolor{lightblue}+0.08 & \cellcolor{lightblue}+0.10 \\                                                              
  $p_t$                                                                                                                                                    
    & -- & $.011^{*}$ & $.064$ & $.039^{*}$                           
    & \cellcolor{lightblue}$.008^{**}$ & \cellcolor{lightblue}$.012^{*}$ & \cellcolor{lightblue}$.006^{**}$                                                
    & -- & $.358$ & $.315$ & $.968$                                                                                                                        
    & \cellcolor{lightblue}$.003^{**}$ & \cellcolor{lightblue}$.023^{*}$ & \cellcolor{lightblue}$.018^{*}$ \\                                              
  $p_z$                                                                                                                                                    
    & -- & $<.001^{***}$ & $.029^{*}$ & $.004^{**}$                                                                                                        
    & \cellcolor{lightblue}$<.001^{***}$ & \cellcolor{lightblue}$<.001^{***}$ & \cellcolor{lightblue}$<.001^{***}$                                         
    & -- & 0.0118 & $.092$ & $.300$                                                                                                                        
    & \cellcolor{lightblue}$<.001^{***}$ & \cellcolor{lightblue}$<.001^{***}$ & \cellcolor{lightblue}$<.001^{***}$ \\                                      
  \bottomrule                                                                                                                                              
  \end{tabular}%
  }                                                                                                                                                        
                                                                                                                                                           
  \end{minipage}%
  \hfill
  \begin{minipage}[t]{0.26\textwidth}
  \centering                                                                                                                                               
  \vspace{0pt}
  \includegraphics[width=0.96\linewidth]{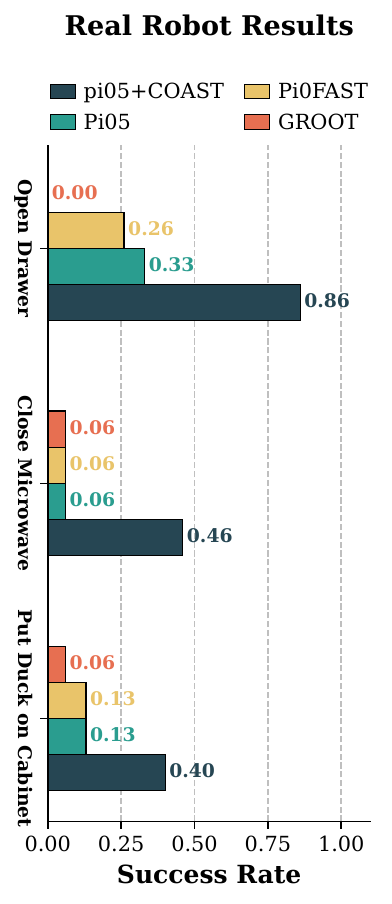}                                                                              
  \end{minipage}                                                                                                                                           
  
  \caption{\textbf{Contrastive conceptors deliver the largest and most consistent gains across every                                                       
  model--benchmark pair.} We evaluate three conceptor steering strategies (+Glob., +Per., +Pos.) defined in
  Section~\ref{sec:conceptors_steering} and compare against +CAA (linear contrastive activation addition),                                                 
  a fine-tuning baseline (+SFT), and a +SAE baseline (TopK SAE feature selection;                                                                          
  per-task best $\alpha$ over the grid $\{0.25, 0.5, 1.0, 2.0\}$) \citep{khan2025controlling}.                                                             
  LIBERO-10 task names describe the manipulated objects (e.g., Stove+Moka = turn on stove and place moka pot; full names in App.~\ref{app:rollout-protocol}).
  $\Delta$: mean gain over unsteered baseline.                                                                                                             
  $p_t$: paired $t$-test across tasks; $p_z$: pooled two-proportion $z$-test.                                                                              
  $^{*}p<0.05$, $^{**}p<0.01$, $^{***}p<0.001$. \textbf{Bold}: best method per row; \underline{underline}: second-best.                                    
  $^\dagger$The Cheese+Butter task has zero failures under the $\pi_0$-FAST checkpoint, so no contrastive conceptor can be fitted; means and $\Delta$ for that column                                                   
  are over 9 contrastive-eligible tasks.                                                                                                                   
  \textbf{(Right)} Real-robot success rates on three manipulation tasks. We compare $\pi_{0.5}$+COAST,                                                     
  $\pi_{0.5}$, $\pi_0$-FAST, and GR00T on 3 three real-world robot tasks. }                                                                                
  \label{tab:all_results}                                                                                                                                  
  \vspace{-12pt}                                                                                                                                           
  \end{table*}
 
\subsection{Does COAST reliably improve task success across diverse VLA models and benchmarks?}
\label{sec:main_results}

Across all model-benchmark pairs, COAST produces statistically significant improvements over the unsteered baseline, with contrastive strategies (global and per-step) consistently outperforming both additive (CAA) and feature-based (SAE) steering as well as supervised fine-tuning (Table~\ref{tab:all_results}). The gains are largest where the baseline is weakest and where success and failure subspaces overlap most heavily, consistent with the mechanistic analysis in Section~\ref{sec:mechanism}. We summarize the key patterns per benchmark below and return to cross-benchmark observations at the end of this subsection.
 
\paragraph{MetaWorld ML45.}
Conceptor steering improves $\pi_{0.5}$ mean success rate from 0.57 to 0.82 under per-step contrastive steering ($+0.25$, $p < .001$), with individual tasks showing dramatic success rate recoveries: \texttt{pick-place-wall} from 0.20 to 1.00 and \texttt{stick-push} from 0.20 to 0.73 (Table~\ref{tab:all_results}; per-task detail in Table~\ref{tab:pi05_metaworld_full}). $\pi_0$-FAST shows a consistent but smaller improvement, rising from 0.57 to 0.69 under global steering ($+0.13$, $p < .001$).
 
\paragraph{LIBERO-10.}
LIBERO produces the largest absolute gains of any benchmark. Per-step steering lifts $\pi_{0.5}$ from 0.43 to 0.80 ($+0.37$, $p < .001$), with the strongest recoveries on \texttt{LR5} ($0.07{\to}0.60$) and \texttt{KS3} ($0.53{\to}0.93$). $\pi_0$-FAST improves from 0.62 to 0.84 ($+0.23$, $p < .001$).
Both CAA and SFT falter here, showing only minimal gains ($+0.04$ and $-0.03$ for $\pi_{0.5}$, respectively); We attribute this to long-horizon dynamics, where small early deviations compound across steps and explicitly suppressing failure modes prevents this accumulation.

\paragraph{RoboCasa.}
RoboCasa requires generalization across randomized placements, textures, and cluttered scenes—the hardest setting. Steering nonetheless yields significant gains across models. $\pi_{0.5}$ reaches 0.56 ($+0.17$, $p_z = .009$) and GR00T N1.5 achieves 0.75 ($+0.16$, $p_z = .006$), showing sharp recoveries on \texttt{Stand Mixer}. While Diffusion Policy also improves ($0.32{\to}0.46$, $+0.14$), its absolute gains are slightly lower than those of the VLA models; this suggests that while steerable success geometry exists in continuous diffusion representations, 
the inherited caches from VLM-backbones
of VLAs might offer more robust decodability for activation steering. That all strategies remain significant confirms this geometry persists even under high visual variability.
 
\paragraph{Real World.} To evaluate whether our method transfers beyond simulation, we test on three real-robot manipulation tasks: \texttt{Open Drawer}, \texttt{Close Microwave}, and \texttt{Put Duck on Cabinet}.
We highlight that these tasks each have very different dynamics, and also were done on $\pi_{0.5}$-DROID policy that was not fine-tuned on any demos.
Across all three tasks, \textsc{$\pi_{0.5}$+COAST} achieves the highest success rate with a 40\% average improvement, with the largest gain on the challenging \texttt{Open Drawer} task.
We also note that with just a single successful rollout, we were able to increase the success rate to 46\% on the \texttt{Close Microwave} task. Visual examples of rollouts are in Fig.~\ref{fig:real_robot_rollouts}.

\paragraph{Comparison against baselines.} 

COAST outperforms CAA because CAA applies a strictly additive shift along a single direction, while conceptors act as subspace-aware filters that selectively scale task-relevant directions and suppress irrelevant ones. COAST also outperforms SAE steering: SAE methods select features in a sparse basis but collapse them back into a single additive direction, inheriting CAA's lack of subspace-aware scaling and adding a reconstruction bottleneck on top. SFT underperforms the base policy on five environments because fine-tuning on a strictly limited budget of just 15 rollouts per task often induces overfitting, degrading the model's pre-trained robustness.

 \paragraph{Cross-benchmark patterns.}
Three patterns hold across all benchmarks and architectures. First, contrastive steering (global or per-step) consistently outperforms positive-only steering, confirming that the failure-subtraction operation contributes meaningfully beyond simply projecting toward success. Second, per-step steering matches or exceeds global steering on long-horizon tasks (LIBERO, RoboCasa) but offers little additional benefit on short-horizon tasks (MetaWorld), suggesting that the discriminative subspace shifts during denoising primarily in settings where multi-step reasoning is required. Third, gains scale inversely with baseline success rate. Tasks where the unsteered policy struggles most show the largest absolute improvements, while near-ceiling tasks show more moderate improvements. 
 

\subsection{How does conceptor steering geometrically reshape internal activations?}
\label{sec:mechanism}

\begin{wrapfigure}{r}{0.50\textwidth}
    \vspace{-\baselineskip}
    \centering
    \includegraphics[width=0.48\textwidth]{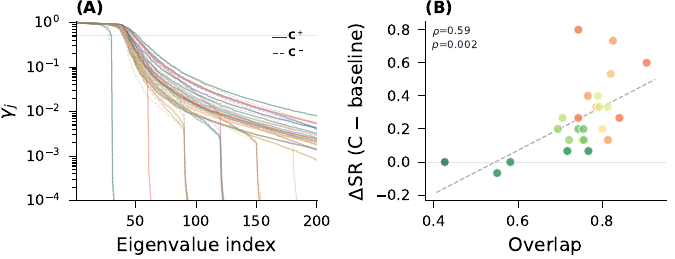}
    \captionsetup{font=small, width=0.46\textwidth}
    \caption{\textbf{The contrastive subspace is low-rank and predictive.}
    \textbf{(A)} Eigenvalue spectra of success ($\mathbf{C}^{+}$, solid) and failure
    ($\mathbf{C}^{-}$, dashed) conceptors on MetaWorld ML45 decay rapidly.
    \textbf{(B)} Per-task steering gain $\Delta\mathrm{SR}$ against overlap
    $\mathrm{sim}(\mathbf{C}^{+}, \mathbf{C}^{-})$. Spearman $\rho = 0.59$, $p = 0.002$.}
    \label{fig:mech_interp_master}
    \vspace{-2pt}
\end{wrapfigure}

\begin{figure*}[h]
    \centering
    \includegraphics[width=\textwidth]{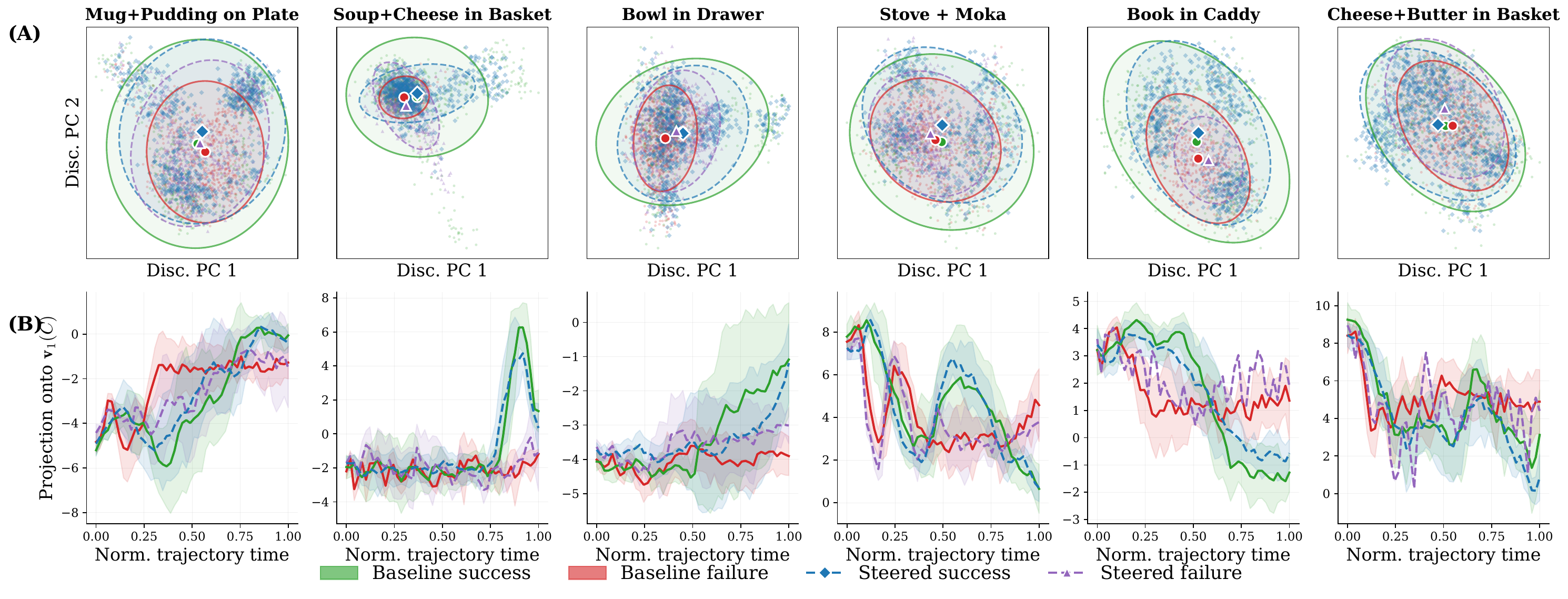}
    \caption{\textbf{Conceptor steering pulls activations toward the success region.}
    \textbf{(A)} Per-step activations projected onto the top two eigenvectors of $\mathbf{C}_{\mathrm{steer}}$, the subspace maximally separating baseline success from failure. Solid ellipses: baseline $2\sigma$ regions; dashed: steered; large markers: centroids.
    \textbf{(B)} Projection onto $\mathbf{v}_1(\mathbf{C}_{\mathrm{steer}})$ over normalized trajectory time (means $\pm 1\sigma$).}
    \label{fig:steering_geometry}
\end{figure*}

The gains reported in Section~\ref{sec:main_results} raise a natural question. What property of the residual stream lets a single closed-form gate produce double-digit success-rate improvements? We identify three geometric facts about the action expert's hidden states (Fig.~\ref{fig:mech_interp_master} and Fig.~\ref{fig:steering_geometry}) to shed light on the effect.
 
\paragraph{Outcome-relevant computation is low-rank, but not rank-one.}

Although the residual stream of the $\pi_{0.5}$ model operates in a 1024-dimensional space, the eigenvalue spectra of both the success ($\mathbf{C}^{+}$) and failure ($\mathbf{C}^{-}$) conceptors decay sharply across all steered MetaWorld ML45 tasks (Figure~\ref{fig:mech_interp_master}A). By applying the Boolean contrastive operation ($\mathbf{C}^{+} \wedge \lnot \mathbf{C}^{-}$), the conceptor cancels out dimensions shared by both outcomes, reducing the effective steering subspace to roughly one percent of the total hidden dimension. 
This low-rank structure enables a closed-form operator to isolate success-relevant directions without disrupting the policy's overall computation. However, this discriminative subspace is strictly multi-dimensional, rather than rank-one. Additive steering along a single mean-difference vector recovers only partial gains (illustrated by the CAA baseline in Table~\ref{tab:all_results}), confirming that rank-one methods cannot fully capture this compact, multi-dimensional outcome geometry.

\paragraph{Subspace overlap predicts steering efficacy.}
The magnitude of steering gain depends heavily on how much the success and failure subspaces overlap. Plotting per-task improvement ($\Delta\mathrm{SR}$) against the success-failure overlap $\mathrm{sim}(\mathbf{C}^{+}, \mathbf{C}^{-}) = \mathrm{tr}(\mathbf{C}^{+}\mathbf{C}^{-})/\sqrt{\mathrm{tr}((\mathbf{C}^{+})^2)\,\mathrm{tr}((\mathbf{C}^{-})^2)}$ yields a strong positive correlation ($\rho = 0.59$, $p = 0.002$; Figure~\ref{fig:mech_interp_master}B). When success and failure subspaces are nearly disjoint, the policy has already separated the two outcomes geometrically and there is little contrastive signal to amplify. When they overlap heavily, the thin residual directions distinguishing them are exactly what the Boolean AND-NOT operation isolates. Because overlap requires only a single matrix product and no rollouts, it also serves as a cheap diagnostic for predicting where steering will be most effective (App.~\ref{sec:param-opt}).

 \paragraph{Steering reshapes the activation geometry toward success.}
Figure~\ref{fig:steering_geometry} visualizes the intervention's geometric effect. Panel A projects per-step activations onto the top two eigenvectors of each task's $\mathbf{C}_{\mathrm{steer}}$, the directions along which success and failure are maximally separated. Green and red points denote individual steered-success and steered-failure activations respectively. The key observation is not simply that steered activations fall within the baseline-success ellipse, but that their \textit{centroids shift} toward the baseline-success centroid and away from the baseline-failure centroid. In Panel B, this shift is visible over trajectory time. Steered successes consistently track the baseline-success mean throughout the rollout, while steered failures either sustain a partial shift or revert to the baseline-failure trajectory entirely. The fact that the physical outcome (success or failure) closely tracks which distribution the activations align with confirms that these geometric regions are not merely descriptive but linked to task execution.

\begin{table*}[t!]
\centering
\resizebox{\textwidth}{!}{%
\begin{tabular}{l | c | cc | l | c | cc}
\toprule
\multicolumn{4}{c|}{\textbf{(a) $\pi_{0.5}$ LIBERO}} & \multicolumn{4}{c}{\textbf{(b) $\pi_{0.5}$ RoboCasa}} \\
\cmidrule(lr){1-4} \cmidrule(lr){5-8}
\textbf{Target} & \textbf{Self} & \textbf{Top-1} & \textbf{Top-2} & \textbf{Target} & \textbf{Self} & \textbf{Top-1} & \textbf{Top-2} \\
\midrule
Stove+Moka     & +0.40 & \textbf{Two Mokas (PS) +0.60}      & Soup+Tomato (G) +0.33           & CloseFridge & +0.27 & PP\_Stove (PS) +0.07          & OpenDrawer (PS) $-$0.20       \\
Bowl+Drawer    & +0.40 & Soup+Tomato (PS) +0.20           & Mug+Pudding (PS) +0.14             & CoffeeSetup & +0.20 & Kettle (G) +0.00              & OpenDrawer (PS) $-$0.20       \\
Mug+Micro      & +0.47 & Two Mokas (PS) +0.20               & Mug+Pudding (G) $-$0.13            & OpenDrawer  & +0.14 & \textbf{CoffeeSetup (PS) +0.60} & CoffeeSetup (PS) +0.47      \\
Two Mokas       & +0.40 & Mugs+Plates (PS) +0.40           & Mug+Pudding (PS) $-$0.06           & StandMixer  & +0.13 & \textbf{CoffeeSetup (G) +0.54}         & \textbf{CoffeeSetup (G) +0.53}         \\
Soup+Cheese   & +0.20 & \textbf{Two Mokas (G) +0.67}                & Soup+Tomato (G) +0.47           & PP\_Cabinet & +0.13 & \textbf{CoffeeSetup (PS) +0.54}        & CoffeeSetup (G) +0.47         \\
Soup+Tomato   & +0.40 & Bowl+Drawer (PS) +0.27            & Bowl+Drawer (PS) +0.27           & PP\_Stove   & +0.34 & CoffeeSetup (G) +0.40         & CoffeeSetup (PS) +0.34        \\
Cheese+Butter & +0.33 & \textbf{Mug+Micro (G) +0.80}      & \textbf{Soup+Tomato (PS) +0.60} & Kettle      & +0.07 & CoffeeSetup (PS) +0.14        & CloseFridge (PS) +0.07        \\
Mugs+Plates   & \textbf{+0.60} & \textbf{Mug+Micro (G) +0.60} & Soup+Tomato (PS) +0.40          &             &       &                               &                               \\
Mug+Pudding      & +0.34 & \textbf{Mug+Micro (G) +0.67}               & Soup+Tomato (PS) +0.40          &             &       &                               &                               \\
Book+Caddy     & +0.33 & \textbf{Soup+Tomato (PS) +0.53}  & Soup+Tomato (G) +0.47           &             &       &                               &                               \\
\bottomrule
\end{tabular}%
}
\vspace{0.6em}
\caption{Transfer experiment across benchmarks. For each target task, we report the best self-steered result (Self) and the top-2 transfer sources. Each transfer cell shows \emph{source (strategy) SR delta}, indicating the change in success rate compared to the target task's unsteered baseline. G = global strategy, PS = per-step strategy. Values with SR delta $\geq +0.50$ are shown in \textbf{bold}. LIBERO task names follow Table~\ref{tab:all_results}.}
\label{tab:transfer_master}
\vspace{-10pt}
\end{table*}

\begin{figure*}[t!]
\centering
\includegraphics[width=1.0\textwidth]{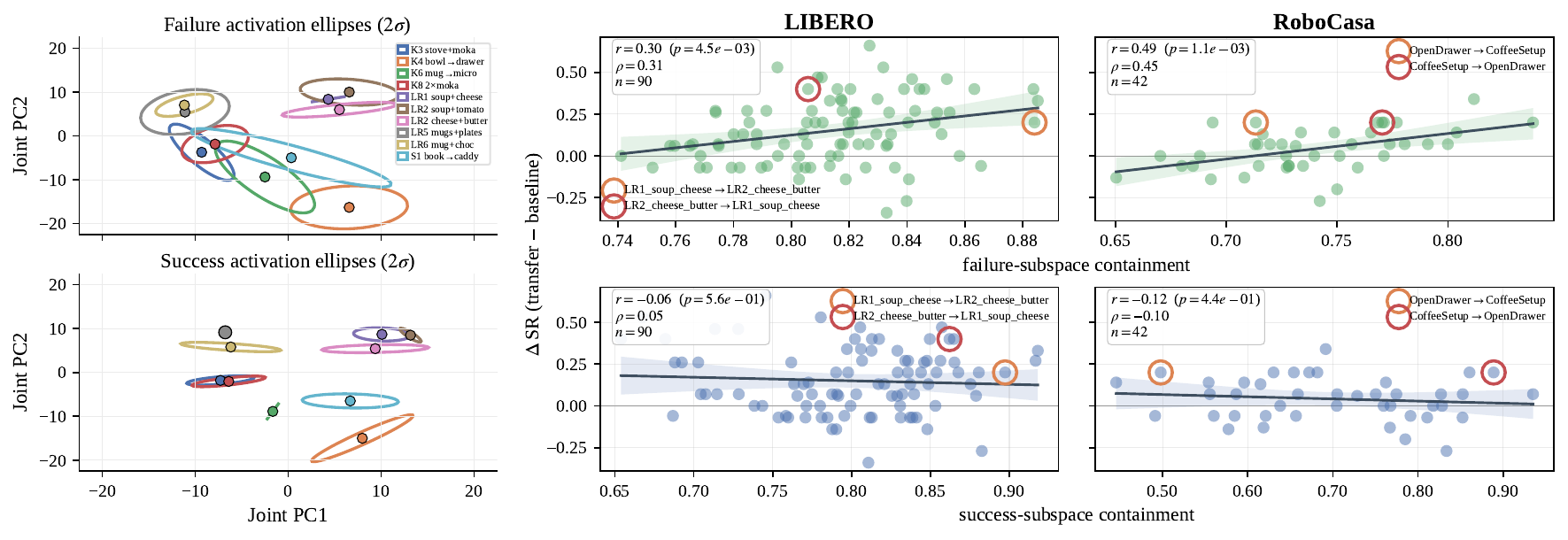}
\caption{\textbf{Tasks share failure geometry but not success geometry, enabling cross-task steering.}
\emph{Left:} Joint PCA of layer-11 activations on $\pi_{0.5}$ LIBERO, $2\sigma$
ellipses per task. Failure activations (top) spread into task-specific regions whose overlap varies;
success activations (bottom) cluster uniformly across tasks.
\emph{Right:} Each dot is one source$\to$target pair on LIBERO or RoboCasa.
Failure-subspace containment
$\mathrm{tr}(C^f_\mathrm{src} C^f_\mathrm{tgt})/\mathrm{tr}((C^f_\mathrm{src})^2)$
correlates with transfer success-rate gain (top; $r{=}0.30$ and $0.49$, $p{<}0.005$);
success containment does not (bottom; $|r|{<}0.13$, $p{>}0.4$).}
\label{fig:sim_vs_sr}
\vspace{-5pt}
\end{figure*}

\subsection{Do tasks share failure geometry, and does it enable cross-task steering?}
\label{sec:transfer}

The self-steering results above prompt a natural follow-up question about the structure of the representations COAST exploits. Examining the activation geometry across tasks reveals an asymmetry. Success activations cluster tightly and uniformly across tasks, while failure activations spread into distinct, task-specific regions (Figure~\ref{fig:sim_vs_sr}, left). In other words, many tasks share overlapping failure modes even though their success representations remain largely disjoint.

This asymmetry has a practical consequence. Because the contrastive conceptor from Section~\ref{sec:method} acts subtractively along failure directions, a conceptor fitted on one source task should also help a different target task whenever their failure subspaces overlap. We test this by fitting conceptors on a source task and applying them to a different target within the same benchmark (Table~\ref{tab:transfer_master}). Transferred conceptors retain most of the gain of the self-fit configuration, and in several cases a transferred conceptor matches or exceeds it.

We quantify this effect by measuring failure-subspace containment $\mathrm{tr}(C^f_\mathrm{src} C^f_\mathrm{tgt})/\mathrm{tr}((C^f_\mathrm{src})^2)$ between source and target conceptors at $L{=}11$. Across off-diagonal source/target pairs, containment correlates with the transfer success-rate gain over baseline (Pearson $r{=}0.30$, $p{=}4.5\mathrm{e}{-3}$ on LIBERO; $r{=}0.49$, $p{=}1.1\mathrm{e}{-3}$ on RoboCasa; Figure~\ref{fig:sim_vs_sr}, top right). The analogous success-subspace containment shows no such relationship (both $|r|{<}0.13$, $p{>}0.4$; bottom right), ruling out the simpler explanation that tasks just share geometry in general. This confirms that shared failure structure, not shared success structure, is the driver of cross-task steering gains.

\section{Limitations, Broad Impact, and Conclusion}

\paragraph{Limitations.}
Contrastive conceptors require both success and failure rollouts, which means uniformly successful tasks might fall back to positive-only steering, which is safe but could be less effective than its contrastive counterpart (App.~\ref{app:posonly-ablation}). Furthermore, tasks with nearly disjoint success and failure subspaces show smaller gains, as the overlap-efficacy relationship predicts. Lastly, whether the geometric hyperparameter selection generalizes beyond the architectures tested here remains open.

\vspace{-10pt}

\paragraph{Broader Impact.}
COAST lowers the cost of adapting frozen robot policies to new tasks, replacing finetuning with a small number of rollouts and a closed-form computation. This accessibility is a double-edged sword. Because steering acts on internal activations rather than outputs, a steered policy is difficult to distinguish from an unmodified one based on action traces alone, and the same operator could be constructed to steer toward failure-aligned behavior. We recommend that deployments log applied conceptors as part of the policy specification and pair them with output-level monitoring.
\vspace{-10pt}

\paragraph{Conclusion.} 
Frozen action experts in VLA models already encode the information needed to distinguish successful from failed behavior. COAST recovers it effectively through a simple closed-form operator, improving mean simulation and real robot success rate by over 20\% and 40\% respectively. We find that the geometric structure that enables steering, low-rank subspaces with contrastively separable success and failure directions, is not specific to any single architecture. Instead, it persists across flow-matching diffusion, autoregressive policies, and Diffusion Policy, suggesting it is a general property of how neural networks encode task-relevant information. We also find that failures tend to share geometric structure across tasks while success representations remain task-specific, and that this shared failure geometry enables a conceptor fitted on one task to improve performance on related tasks without refitting. More broadly, VLA execution failure could be mitigated without retraining if the policy's internal representations already encode the information needed for success, and a closed-form geometric operator is sufficient to surface it.


\bibliographystyle{neurips_2026}
\bibliography{neurips_2026}
\newpage
\appendix
\newpage
\appendix
\section{Appendix}
 
This appendix provides supplementary results, technical details, and reproducibility information. 
App.~\ref{app:appendix_related_works} provides an extended related work section that categorizes prior adaptation baselines by their deployment burdens, justifying our focus on training-free, self-contained steering methods. App.~\ref{app:results_with_params} reports per-task results with optimal hyperparameters for all model-benchmark pairs. App.~\ref{app:linear_random_ablation} presents linear and random-direction ablations isolating the contribution of the contrastive subspace and the multiplicative form. App.~\ref{app:posonly-ablation} tests positive-only conceptor steering on high-baseline tasks to verify it does not degrade performance. App.~\ref{app:failure_only_ablation} tests failure-only steering with $\neg C^f$ on tasks with 0\% baseline success. App.~\ref{app:dp-robocasa-per-epoch} evaluates steering across multiple Diffusion Policy training epochs, network layers, and hyperparameters to demonstrate that performance gains are consistent and not an artifact of checkpoint or layer selection. App.~\ref{app:activation-extraction} details the activation extraction procedure for each model family. App.~\ref{app:rollout-protocol} describes the rollout collection protocol per benchmark. App.~\ref{app:conceptor-computation} covers conceptor computation, steering application, runtime overhead, numerical precision, and memory requirements. App.~\ref{app:hyperparameter-sweep} specifies the full hyperparameter sweep grid and the geometric selection procedure. App.~\ref{app:checkpoint-training} provides checkpoint and training details for all models. App.~\ref{app:real-robot-setup} describes the real-world evaluation hardware setup. App.~\ref{app:caa} and App.~\ref{app:sae} contain the implementations details for the contrastive activation addition and the sparse autoencoder baselines respectively. App. ~\ref{app:activation_datasets} contains the activation dataset composition, showing the success vs. failure counts in the $15$ training rollouts.  App.~\ref{app:overhead} includes analyses on COAST's steering overhead compared with other steering baselines. 

\subsection{Baseline Analysis}
\label{app:appendix_related_works}

\begin{table}[h]
\centering
\caption{Comparison of COAST against existing VLA adaptation paradigms. COAST is the only method that achieves holistic, multi-dimensional steering without requiring gradient updates, external models, or computationally expensive test-time search.}
\label{tab:method_comparison}
\renewcommand{\arraystretch}{1.3}
\resizebox{\textwidth}{!}{
\begin{tabular}{lcccc}
\toprule
\textbf{Method}
& \textbf{\shortstack{Training-\\Free}}
& \textbf{\shortstack{Self-\\Contained}}
& \textbf{\shortstack{Single-\\Pass}} \\
\midrule
Fine-Tuning / SFT \cite{yuan2024policy} 
& \xmark & \cmark & \cmark \\

VLM Search / VLS \cite{liu2026vls} 
& \cmark & \xmark & \xmark \\

Test-Time Search / TACO \cite{yang2025steering} 
& \cmark & \xmark & \xmark \\

SAE Steering \cite{swann2026sparse} 
& \xmark & \cmark & \cmark \\

Activation Addition \cite{turner2023activation} 
& \cmark & \cmark & \cmark \\

\midrule
\textbf{COAST (Ours)} 
& \cmark & \cmark & \cmark \\
\bottomrule
\end{tabular}}
\end{table}

To contextualize our contribution, Table \ref{tab:method_comparison} compares COAST against recent methods for inference-time adaptation and steering of robot policies. Because the computational requirements and operational paradigms of these methods vary significantly, we categorize them along four distinct axes that define their deployment burden. Below are the definitions of each comparison criterion:

\paragraph{Training-Free.} 
A method is considered training-free if it does not require gradient updates to the policy weights, auxiliary model weights, or any explicitly learned parameter modules. Approaches such as supervised fine-tuning (SFT) \citep{yuan2024policy, mitra2025robotic} and explicitly trained steering interfaces \citep{chen2026steerable, das2025lae} require offline retraining or the optimization of auxiliary modules, thus incurring a high computational cost before deployment. Contrastive Representation Regularization \citep{kim2025contrastive} similarly modifies the core policy weights. Conversely, COAST relies on closed-form Boolean subspace algebra computed directly from rollouts \citep{postmus2024steering}, bypassing gradient-based optimization entirely. 

\paragraph{Self-Contained.}
A method is self-contained if it does not rely on querying external foundation models (e.g., Vision-Language Models) or explicit external reward models at inference time. Several recent inference-time adaptation frameworks rely on VLM-based planning, scoring, or verification to guide the robot's actions. For example, VLS \citep{liu2026vls} relies on external VLM-based rewards for steering, while \citet{shah2025liten} and \citet{wu2025dowhatyousay} use VLMs for runtime reasoning and action alignment verification. These external dependencies introduce significant latency and point-of-failure risks. COAST is purely internal, operating exclusively on the native residual stream of the frozen VLA.
We exclude methods like VLS~\citep{liu2026vls} and TACO~\citep{yang2025steering} because their reliance on external VLMs or multi-sample search makes their operational cost structurally incomparable to COAST's single-pass, negligible-overhead execution (App.~\ref{app:appendix_related_works}). 

\paragraph{Single-Pass.}
A single-pass method alters the model's behavior without requiring multi-sample search, Monte Carlo sampling, or iterative refinement loops during action generation. Methods that rely on test-time search or trajectory sampling, such as TACO \citep{yang2025steering} or VLM-guided rollout evaluation, multiply the inference latency of the policy by the number of samples required. COAST is applied as a fixed multiplicative gate to the hidden states during the standard forward pass, yielding effectively zero latency overhead. 

\paragraph{Subspace Control.}
This criterion distinguishes methods that intervene on multi-dimensional, outcome-relevant geometric structures from those that intervene on isolated directions. Standard Activation Addition (CAA) \citep{turner2023activation, panickssery2023steering} shifts activations along a single rank-one vector. Similarly, Sparse Autoencoder (SAE) steering \citep{swann2026sparse, buurmeijer2026observing, khan2025controlling} isolates and scales individual, disentangled features. While effective for localized semantic concepts \citep{haon2025mechanistic}, these methods offer only partial, constrained control. Drawing on conceptor theory \citep{jaeger2014controlling, postmus2025affine}, COAST leverages a continuous, high-dimensional subspace \citep{zou2023representation} that explicitly separates successful and failed behaviors. This structure enables holistic, multi-dimensional steering rather than feature-by-feature manipulation.

By satisfying all four criteria, COAST uniquely addresses the representation bottleneck in VLA policies without the latency of external VLM queries \citep{liu2026vls}, the computational cost of test-time search \citep{yang2025steering}, or the partial control of single-feature interventions \citep{khan2025controlling}. We view this framing as essential for standardizing how inference-time adaptation methods are evaluated in embodied AI, as matched-cost comparisons (e.g., COAST vs. Activation Addition) yield significantly more rigorous insights into the geometry of policy latent spaces than comparing single-pass internal methods against multi-sample external searches.

\subsection{Results by Tasks With Optimal Parameters}
\label{app:results_with_params}
Tables \ref{tab:pi05_libero_full}--\ref{tab:groot_robocasa_full} report the per-task success rate on held-out test rollouts for the steering configuration that achieved the highest success rate on the fitting rollouts, under the full hyperparameter sweep described in Appendix~\ref{app:hyperparameter-sweep}. For each entry, we also list the layer $\ell$, aperture $\alpha$, and steering strength $\beta$ of the selected configuration. These tables back the aggregated numbers in Table~\ref{tab:all_results} and expose two patterns that recur across model--benchmark pairs. First, the oracle-optimal layer concentrates in a narrow band: $\ell{=}11$ (of 18) for the $\pi_{0.5}$ action expert and $\ell{=}10$ (of 16) for the GR00T DiT, with occasional shifts to $\ell{=}5$ on LIBERO tasks where the discriminative subspace forms earlier. This aligns with the quota peak used in Stage 1 of our geometric selection procedure (Sec.~\ref{sec:param-opt}). Second, small steering strengths dominate: $\beta \in \{0.1, 0.3\}$ attains the best fitting-set success rate in over 90\% of cells, with $\beta{=}0.5$ preferred only on a handful of tasks where the baseline is already high and the conceptor acts as a light nudge. Together these tables justify the pruned grid used by geometric selection and show that the same small configuration neighborhood recovers near-oracle performance across tasks.

\begin{table*}[h]
\centering
\caption{Conceptor-based activation steering results of $\pi_0.5$ on LIBERO-10 with layer sweep in $\in \{5, 11\}$. Each condition is evaluated over 30 held-out test rollouts.}
\label{tab:pi05_libero_full}
\resizebox{\textwidth}{!}{%
\begin{tabular}{l c c c}
\toprule
\textbf{Task} & \textbf{Global} & \textbf{Per-Step} & \textbf{Pos.-Only} \\
\midrule
\texttt{KS3: Turn On The Stove And Put The Moka Pot On It} & \textbf{0.93} \scriptsize{(L5, $\alpha$\!=0.5, $\beta$\!=0.1)} & \textbf{0.87} \scriptsize{(L5, $\alpha$\!=10.0, $\beta$\!=0.3)} & \textbf{0.80} \scriptsize{(L11, $\alpha$\!=1.0, $\beta$\!=0.1)} \\
\texttt{KS4: Put The Black Bowl In The Bottom Drawer Of...} & \textbf{0.73} \scriptsize{(L5, $\alpha$\!=1.0, $\beta$\!=0.1)} & \textbf{0.80} \scriptsize{(L5, $\alpha$\!=0.1, $\beta$\!=0.1)} & \textbf{0.87} \scriptsize{(L5, $\alpha$\!=1.0, $\beta$\!=0.1)} \\
\texttt{KS6: Put The Yellow And White Mug In The Microw...} & \textbf{0.40} \scriptsize{(L11, $\alpha$\!=0.1, $\beta$\!=0.3)} & \textbf{0.60} \scriptsize{(L11, $\alpha$\!=0.5, $\beta$\!=0.3)} & \textbf{0.33} \scriptsize{(L5, $\alpha$\!=1.0, $\beta$\!=0.1)} \\
\texttt{KS8: Put Both Moka Pots On The Stove} & \textbf{0.47} \scriptsize{(L5, $\alpha$\!=1.0, $\beta$\!=0.1)} & \textbf{0.53} \scriptsize{(L11, $\alpha$\!=2.0, $\beta$\!=0.3)} & 0.13 \scriptsize{(L5, $\alpha$\!=1.0, $\beta$\!=0.1)} \\
\texttt{LR1: Put Both The Alphabet Soup And The Cream C...} & \textbf{0.93} \scriptsize{(L11, $\alpha$\!=0.5, $\beta$\!=0.3)} & \textbf{1.00} \scriptsize{(L11, $\alpha$\!=2.0, $\beta$\!=0.1)} & \textbf{0.93} \scriptsize{(L5, $\alpha$\!=1.0, $\beta$\!=0.1)} \\
\texttt{LR2: Put Both The Alphabet Soup And The Tomato...} & \textbf{0.80} \scriptsize{(L11, $\alpha$\!=0.5, $\beta$\!=0.3)} & \textbf{0.80} \scriptsize{(L5, $\alpha$\!=2.0, $\beta$\!=0.1)} & \textbf{0.60} \scriptsize{(L5, $\alpha$\!=1.0, $\beta$\!=0.3)} \\
\texttt{LR2: Put Both The Cream Cheese Box And The Butt...} & \textbf{0.93} \scriptsize{(L5, $\alpha$\!=0.5, $\beta$\!=0.1)} & \textbf{0.93} \scriptsize{(L5, $\alpha$\!=0.1, $\beta$\!=0.3)} & \textbf{0.87} \scriptsize{(L5, $\alpha$\!=0.5, $\beta$\!=0.1)} \\
\texttt{LR5: Put The White Mug On The Left Plate And Pu...} & \textbf{0.60} \scriptsize{(L11, $\alpha$\!=0.5, $\beta$\!=0.5)} & \textbf{0.60} \scriptsize{(L11, $\alpha$\!=2.0, $\beta$\!=0.5)} & \textbf{0.20} \scriptsize{(L11, $\alpha$\!=0.5, $\beta$\!=0.1)} \\
\texttt{LR6: Put The White Mug On The Plate And Put The...} & \textbf{0.87} \scriptsize{(L5, $\alpha$\!=0.5, $\beta$\!=0.1)} & \textbf{0.87} \scriptsize{(L5, $\alpha$\!=0.1, $\beta$\!=0.1)} & \textbf{0.67} \scriptsize{(L5, $\alpha$\!=0.5, $\beta$\!=0.3)} \\
\texttt{ST1: Pick Up The Book And Place It In The Back...} & \textbf{0.93} \scriptsize{(L5, $\alpha$\!=0.5, $\beta$\!=0.1)} & \textbf{1.00} \scriptsize{(L5, $\alpha$\!=0.5, $\beta$\!=0.1)} & \textbf{0.93} \scriptsize{(L5, $\alpha$\!=0.5, $\beta$\!=0.1)} \\
\midrule
\textbf{Mean} & \textbf{0.76} & \textbf{0.80} & \textbf{0.63} \\
\bottomrule
\end{tabular}%
}
\end{table*}
\begin{table*}[h]                      
  \centering
  \caption{Conceptor-based activation steering results of $\pi_0$-FAST on LIBERO-10. LR2 (cream cheese \& 
  butter) has $n_{\text{fail}}=0$ under this checkpoint, so contrastive conceptors (Global, Per-Step)     
  cannot be constructed; means are computed over the 9 contrastive-eligible tasks.Each condition is evaluated over 30 held-out test rollouts.}                       
  \label{tab:pi0fast_libero_full}                                                                         
  \resizebox{\textwidth}{!}{%
  \begin{tabular}{l c c c}                                                                                
  \toprule
  \textbf{Task} & \textbf{Global} & \textbf{Per-Step} & \textbf{Pos.-Only} \\                             
  \midrule                                                                                                
  \texttt{KS3: Turn On The Stove And Put The Moka Pot On It} & \textbf{1.00} \scriptsize{($\alpha$\!=0.1,
  $\beta$\!=0.02)} & \textbf{0.87} \scriptsize{($\alpha$\!=0.1, $\beta$\!=0.3)} & 0.80                    
  \scriptsize{($\alpha$\!=0.1, $\beta$\!=0.05)} \\                
  \texttt{KS4: Put The Black Bowl In The Bottom Drawer Of...} & \textbf{0.93} \scriptsize{($\alpha$\!=2.0,
   $\beta$\!=0.1)} & \textbf{0.93} \scriptsize{($\alpha$\!=0.1, $\beta$\!=0.2)} & \textbf{0.80}           
  \scriptsize{($\alpha$\!=0.5, $\beta$\!=0.1)} \\
  \texttt{KS6: Put The Yellow And White Mug In The Microw...} & \textbf{0.73}                             
  \scriptsize{($\alpha$\!=10.0, $\beta$\!=0.3)} & \textbf{0.80} \scriptsize{($\alpha$\!=0.5,              
  $\beta$\!=0.1)} & \textbf{0.73} \scriptsize{($\alpha$\!=2.0, $\beta$\!=0.1)} \\
  \texttt{KS8: Put Both Moka Pots On The Stove} & \textbf{0.53} \scriptsize{($\alpha$\!=5.0,              
  $\beta$\!=0.15)} & \textbf{0.53} \scriptsize{($\alpha$\!=3.0, $\beta$\!=0.2)} & \textbf{0.40}           
  \scriptsize{($\alpha$\!=3.0, $\beta$\!=0.01)} \\
  \texttt{LR1: Put Both The Alphabet Soup And The Cream C...} & \textbf{0.87}                             
  \scriptsize{($\alpha$\!=10.0, $\beta$\!=0.2)} & \textbf{0.87} \scriptsize{($\alpha$\!=0.1,              
  $\beta$\!=0.2)} & \textbf{1.00} \scriptsize{($\alpha$\!=0.1, $\beta$\!=0.1)} \\
  \texttt{LR2: Put Both The Alphabet Soup And The Tomato...} & \textbf{0.87} \scriptsize{($\alpha$\!=2.0, 
  $\beta$\!=0.2)} & \textbf{0.93} \scriptsize{($\alpha$\!=0.1, $\beta$\!=0.2)} & \textbf{0.73}            
  \scriptsize{($\alpha$\!=0.2, $\beta$\!=0.05)} \\
  \texttt{LR2: Put Both The Cream Cheese Box And The Butt...} & -- \scriptsize{(no $C_{\text{fail}}$)} &  
  -- \scriptsize{(no $C_{\text{fail}}$)} & 1.00 \scriptsize{($\alpha$\!=0.5, $\beta$\!=0.1)} \\           
  \texttt{LR5: Put The White Mug On The Left Plate And Pu...} & \textbf{0.87} \scriptsize{($\alpha$\!=1.0,
   $\beta$\!=0.2)} & \textbf{0.87} \scriptsize{($\alpha$\!=0.1, $\beta$\!=0.3)} & \textbf{0.80}           
  \scriptsize{($\alpha$\!=0.1, $\beta$\!=0.05)} \\                
  \texttt{LR6: Put The White Mug On The Plate And Put The...} & \textbf{0.93} \scriptsize{($\alpha$\!=1.0,
   $\beta$\!=0.1)} & \textbf{0.93} \scriptsize{($\alpha$\!=0.3, $\beta$\!=0.05)} & \textbf{0.87}          
  \scriptsize{($\alpha$\!=0.1, $\beta$\!=0.1)} \\
  \texttt{ST1: Pick Up The Book And Place It In The Back...} & \textbf{0.87} \scriptsize{($\alpha$\!=10.0,
   $\beta$\!=0.3)} & \textbf{0.87} \scriptsize{($\alpha$\!=0.07, $\beta$\!=0.1)} & \textbf{0.73}          
  \scriptsize{($\alpha$\!=0.07, $\beta$\!=0.12)} \\
  \midrule                                                                                                
  \textbf{Mean} & \textbf{0.84} & \textbf{0.84} & \textbf{0.79} \\
  \bottomrule                                                                                             
  \end{tabular}%
  }                                                                                                       
  \end{table*}   
\begin{table*}[t!]
\centering
\caption{Conceptor-based activation steering results of $\pi_{0.5}$ on RoboCasa with layer sweep $\in \{5, 11\}$. Each condition is evaluated over 30 held-out test rollouts.}
\label{tab:pi05_robocasa_full}
\resizebox{\textwidth}{!}{%
\begin{tabular}{l c c c}
\toprule
\textbf{Task} & \textbf{Global} & \textbf{Per-Step} & \textbf{Pos.-Only} \\
\midrule
\texttt{Close Fridge} & \textbf{0.47} \scriptsize{(L11, $\alpha$\!=0.5, $\beta$\!=0.3)} & \textbf{0.40} \scriptsize{($t$\!=0, L11, $\alpha$\!=0.1, $\beta$\!=0.5)} & \textbf{0.40} \scriptsize{(L5, $\alpha$\!=2.0, $\beta$\!=0.1)} \\
\texttt{Coffee Setup Mug} & \textbf{0.33} \scriptsize{(L11, $\alpha$\!=0.1, $\beta$\!=0.1)} & \textbf{0.33} \scriptsize{($t$\!=0, L5, $\alpha$\!=0.5, $\beta$\!=0.1)} & \textbf{0.53} \scriptsize{(L5, $\alpha$\!=0.5, $\beta$\!=0.5)} \\
\texttt{Open Drawer} & \textbf{0.67} \scriptsize{(L11, $\alpha$\!=0.5, $\beta$\!=0.1)} & 0.53 \scriptsize{($t$\!=0, L5, $\alpha$\!=0.5, $\beta$\!=0.1)} & \textbf{0.67} \scriptsize{(L5, $\alpha$\!=1.0, $\beta$\!=0.3)} \\
\texttt{Open Stand Mixer Head} & \textbf{0.73} \scriptsize{(L11, $\alpha$\!=0.5, $\beta$\!=0.1)} & \textbf{0.80} \scriptsize{($t$\!=9, L11, $\alpha$\!=1.0, $\beta$\!=0.1)} & 0.53 \scriptsize{(L5, $\alpha$\!=0.1, $\beta$\!=0.3)} \\
\texttt{Pick Place Counter To Cabinet} & \textbf{0.73} \scriptsize{(L5, $\alpha$\!=0.1, $\beta$\!=0.1)} & \textbf{0.73} \scriptsize{($t$\!=0, L5, $\alpha$\!=0.5, $\beta$\!=0.1)} & \textbf{0.67} \scriptsize{(L5, $\alpha$\!=10.0, $\beta$\!=0.3)} \\
\texttt{Pick Place Counter To Stove} & \textbf{0.53} \scriptsize{(L11, $\alpha$\!=0.1, $\beta$\!=0.1)} & \textbf{0.67} \scriptsize{($t$\!=0, L11, $\alpha$\!=1.0, $\beta$\!=0.1)} & \textbf{0.60} \scriptsize{(L5, $\alpha$\!=0.5, $\beta$\!=0.1)} \\
\texttt{Turn On Electric Kettle} & \textbf{0.40} \scriptsize{(L11, $\alpha$\!=0.5, $\beta$\!=0.1)} & \textbf{0.40} \scriptsize{($t$\!=0, L11, $\alpha$\!=2.0, $\beta$\!=0.1)} & \textbf{0.53} \scriptsize{(L5, $\alpha$\!=0.5, $\beta$\!=0.3)} \\
\midrule
\textbf{Mean} & \textbf{0.55} & \textbf{0.55} & \textbf{0.56} \\
\bottomrule
\end{tabular}%
}
\end{table*}
\begin{table*}[t!]                                                                                    
  \centering
  \caption{\textbf{Conceptor steering results of $\pi_0.5$ on MetaWorld ML45 (all 26 steered tasks).} For each task 
  and steering strategy, we report the best success rate over a hyperparameter sweep of $\alpha \in    
  \{0.1, 0.5, 1.0\}$ and $\beta \in \{0.1, 0.3, 0.5\}$ at layer 11. \textbf{Global}: contrastive
  conceptor ($C_{\text{succ}} \wedge \neg C_{\text{fail}}$) applied uniformly across denoise steps.    
  \textbf{Per-Step}: contrastive conceptor applied independently at each denoise step.
  \textbf{Pos.-Only}: success conceptor without failure subtraction. Best hyperparameters shown in
  parentheses. Each condition is evaluated over 30 held-out test rollouts.}
  \label{tab:pi05_metaworld_full}
  \resizebox{\textwidth}{!}{%
  \begin{tabular}{l c c c c}
  \toprule                                                                                             
  \textbf{Task} & \textbf{Baseline} & \textbf{Global} & \textbf{Per-Step} & \textbf{Pos.-Only} \\
  \midrule                                                                                             
  \texttt{assembly} & 0.27 & \textbf{1.00} \scriptsize{($\alpha$\!=\!0.5, $\beta$\!=\!0.1)} &
  \textbf{0.80} \scriptsize{($\alpha$\!=\!0.5, $\beta$\!=\!0.1)} & 0.27 \scriptsize{($\alpha$\!=\!0.1, 
  $\beta$\!=\!0.5)} \\                               
  \texttt{basketball} & 0.40 & \textbf{0.47} \scriptsize{($\alpha$\!=\!0.1, $\beta$\!=\!0.1)} &        
  \textbf{0.60} \scriptsize{($\alpha$\!=\!0.1, $\beta$\!=\!0.1)} & \textbf{0.60}                       
  \scriptsize{($\alpha$\!=\!0.5, $\beta$\!=\!0.1)} \\
  \texttt{coffee-pull} & 0.93 & 0.93 \scriptsize{($\alpha$\!=\!0.1, $\beta$\!=\!0.1)} & 0.93           
  \scriptsize{($\alpha$\!=\!1.0, $\beta$\!=\!0.1)} & \textbf{1.00} \scriptsize{($\alpha$\!=\!0.1,      
  $\beta$\!=\!0.3)} \\                         
  \texttt{coffee-push} & 0.80 & \textbf{1.00} \scriptsize{($\alpha$\!=\!1.0, $\beta$\!=\!0.1)} &       
  \textbf{1.00} \scriptsize{($\alpha$\!=\!0.5, $\beta$\!=\!0.1)} & \textbf{1.00}                       
  \scriptsize{($\alpha$\!=\!0.1, $\beta$\!=\!0.1)} \\
  \texttt{disassemble} & 0.60 & \textbf{0.93} \scriptsize{($\alpha$\!=\!0.1, $\beta$\!=\!0.1)} &       
  \textbf{0.93} \scriptsize{($\alpha$\!=\!0.1, $\beta$\!=\!0.1)} & \textbf{0.87}                       
  \scriptsize{($\alpha$\!=\!0.5, $\beta$\!=\!0.5)} \\
  \texttt{door-open} & 0.93 & \textbf{1.00} \scriptsize{($\alpha$\!=\!0.1, $\beta$\!=\!0.1)} & \textbf{1.00}             
  \scriptsize{($\alpha$\!=\!0.1, $\beta$\!=\!0.1)} & \textbf{1.00} \scriptsize{($\alpha$\!=\!0.1,               
  $\beta$\!=\!0.1)} \\                         
  \texttt{faucet-close} & 0.80 & \textbf{1.00} \scriptsize{($\alpha$\!=\!1.0, $\beta$\!=\!0.1)} &      
  \textbf{0.93} \scriptsize{($\alpha$\!=\!0.1, $\beta$\!=\!0.1)} & \textbf{0.93}                       
  \scriptsize{($\alpha$\!=\!0.1, $\beta$\!=\!0.1)} \\
  \texttt{hammer} & 0.33 & \textbf{0.87} \scriptsize{($\alpha$\!=\!0.5, $\beta$\!=\!0.1)} &            
  \textbf{0.87} \scriptsize{($\alpha$\!=\!0.1, $\beta$\!=\!0.1)} & \textbf{0.40}                       
  \scriptsize{($\alpha$\!=\!0.1, $\beta$\!=\!0.1)} \\
  \texttt{handle-pull-side} & 0.20 & \textbf{0.47} \scriptsize{($\alpha$\!=\!0.1, $\beta$\!=\!0.1)} &  
  \textbf{0.33} \scriptsize{($\alpha$\!=\!0.5, $\beta$\!=\!0.1)} & \textbf{0.53}                       
  \scriptsize{($\alpha$\!=\!0.5, $\beta$\!=\!0.1)} \\
  \texttt{handle-pull} & 0.47 & 0.47 \scriptsize{($\alpha$\!=\!0.1, $\beta$\!=\!0.1)} & \textbf{0.60}  
  \scriptsize{($\alpha$\!=\!0.1, $\beta$\!=\!0.1)} & \textbf{0.53} \scriptsize{($\alpha$\!=\!0.1,      
  $\beta$\!=\!0.1)} \\                         
  \texttt{lever-pull} & 0.27 & \textbf{0.60} \scriptsize{($\alpha$\!=\!0.1, $\beta$\!=\!0.1)} &        
  \textbf{0.67} \scriptsize{($\alpha$\!=\!0.1, $\beta$\!=\!0.3)} & \textbf{0.73}                       
  \scriptsize{($\alpha$\!=\!0.5, $\beta$\!=\!0.3)} \\
  \texttt{peg-insert-side} & 0.33 & \textbf{0.47} \scriptsize{($\alpha$\!=\!0.1, $\beta$\!=\!0.3)} &   
  \textbf{0.67} \scriptsize{($\alpha$\!=\!0.1, $\beta$\!=\!0.3)} & \textbf{0.53}                       
  \scriptsize{($\alpha$\!=\!0.1, $\beta$\!=\!0.3)} \\
  \texttt{pick-out-of-hole} & 0.20 & \textbf{0.27} \scriptsize{($\alpha$\!=\!0.1, $\beta$\!=\!0.3)} &  
  \textbf{0.47} \scriptsize{($\alpha$\!=\!0.1, $\beta$\!=\!0.1)} & \textbf{0.47}                       
  \scriptsize{($\alpha$\!=\!1.0, $\beta$\!=\!0.1)} \\
  \texttt{pick-place} & 0.87 & \textbf{1.00} \scriptsize{($\alpha$\!=\!0.5, $\beta$\!=\!0.1)} &        
  \textbf{1.00} \scriptsize{($\alpha$\!=\!0.1, $\beta$\!=\!0.1)} & 0.80 \scriptsize{($\alpha$\!=\!0.1, 
  $\beta$\!=\!0.3)} \\                         
  \texttt{pick-place-wall} & 0.20 & \textbf{0.87} \scriptsize{($\alpha$\!=\!0.5, $\beta$\!=\!0.1)} &   
  \textbf{1.00} \scriptsize{($\alpha$\!=\!0.1, $\beta$\!=\!0.1)} & \textbf{0.47}                       
  \scriptsize{($\alpha$\!=\!1.0, $\beta$\!=\!0.3)} \\
  \texttt{plate-slide-back-side} & 0.60 & \textbf{0.93} \scriptsize{($\alpha$\!=\!1.0,                 
  $\beta$\!=\!0.1)} & \textbf{1.00} \scriptsize{($\alpha$\!=\!0.1, $\beta$\!=\!0.3)} & 0.60            
  \scriptsize{($\alpha$\!=\!1.0, $\beta$\!=\!0.1)} \\
  \texttt{plate-slide-back} & 0.60 & \textbf{0.93} \scriptsize{($\alpha$\!=\!0.1, $\beta$\!=\!0.1)} &  
  \textbf{0.93} \scriptsize{($\alpha$\!=\!1.0, $\beta$\!=\!0.1)} & \textbf{0.73}                       
  \scriptsize{($\alpha$\!=\!0.1, $\beta$\!=\!0.1)} \\
  \texttt{push-back} & 0.67 & \textbf{0.93} \scriptsize{($\alpha$\!=\!0.1, $\beta$\!=\!0.1)} &         
  \textbf{0.87} \scriptsize{($\alpha$\!=\!1.0, $\beta$\!=\!0.3)} & \textbf{0.87}                       
  \scriptsize{($\alpha$\!=\!0.1, $\beta$\!=\!0.3)} \\
  \texttt{push} & 0.93 & \textbf{1.00} \scriptsize{($\alpha$\!=\!0.1, $\beta$\!=\!0.1)} & \textbf{1.00}
   \scriptsize{($\alpha$\!=\!1.0, $\beta$\!=\!0.1)} & 0.93 \scriptsize{($\alpha$\!=\!1.0,              
  $\beta$\!=\!0.1)} \\                         
  \texttt{reach} & 0.93 & 0.93 \scriptsize{($\alpha$\!=\!0.1, $\beta$\!=\!0.1)} & 0.93                 
  \scriptsize{($\alpha$\!=\!0.1, $\beta$\!=\!0.1)} & \textbf{1.00} \scriptsize{($\alpha$\!=\!0.1,      
  $\beta$\!=\!0.1)} \\                         
  \texttt{shelf-place} & 0.73 & \textbf{0.93} \scriptsize{($\alpha$\!=\!0.5, $\beta$\!=\!0.1)} &       
  \textbf{0.93} \scriptsize{($\alpha$\!=\!0.5, $\beta$\!=\!0.1)} & \textbf{0.93}                       
  \scriptsize{($\alpha$\!=\!0.1, $\beta$\!=\!0.3)} \\
  \texttt{soccer} & 0.27 & \textbf{0.40} \scriptsize{($\alpha$\!=\!0.1, $\beta$\!=\!0.3)} &            
  \textbf{0.40} \scriptsize{($\alpha$\!=\!1.0, $\beta$\!=\!0.5)} & \textbf{0.40}                       
  \scriptsize{($\alpha$\!=\!0.1, $\beta$\!=\!0.5)} \\
  \texttt{stick-pull} & 0.73 & \textbf{0.87} \scriptsize{($\alpha$\!=\!1.0, $\beta$\!=\!0.1)} &        
  \textbf{0.87} \scriptsize{($\alpha$\!=\!0.1, $\beta$\!=\!0.1)} & \textbf{0.87}                       
  \scriptsize{($\alpha$\!=\!0.1, $\beta$\!=\!0.1)} \\
  \texttt{stick-push} & 0.20 & \textbf{0.80} \scriptsize{($\alpha$\!=\!0.1, $\beta$\!=\!0.1)} &        
  \textbf{0.73} \scriptsize{($\alpha$\!=\!1.0, $\beta$\!=\!0.1)} & \textbf{0.40}                       
  \scriptsize{($\alpha$\!=\!0.1, $\beta$\!=\!0.3)} \\
  \texttt{sweep-into} & 0.73 & \textbf{0.87} \scriptsize{($\alpha$\!=\!0.5, $\beta$\!=\!0.1)} &        
  \textbf{0.87} \scriptsize{($\alpha$\!=\!0.1, $\beta$\!=\!0.1)} & \textbf{0.93}                       
  \scriptsize{($\alpha$\!=\!1.0, $\beta$\!=\!0.1)} \\
  \texttt{sweep} & 0.93 & \textbf{1.00} \scriptsize{($\alpha$\!=\!1.0, $\beta$\!=\!0.1)} &             
  \textbf{1.00} \scriptsize{($\alpha$\!=\!0.5, $\beta$\!=\!0.1)} & \textbf{1.00}                       
  \scriptsize{($\alpha$\!=\!0.1, $\beta$\!=\!0.1)} \\
  \midrule                                                                                             
  \textbf{Steered tasks (26)} & 0.57 & 0.81 & 0.82 & 0.72 \\
  \quad $\Delta$ vs.\ Baseline & --- & +0.23 & +0.25 & +0.15 \\
  \bottomrule                                                                                          
  \end{tabular}%
  }                                                                                                    
  \end{table*}     
 \begin{table*}[t!]                                                                                                                         
    \centering                                                    
    \caption{\textbf{Conceptor steering results for $\pi_0$-FAST on MetaWorld ML45 (all 27 steered tasks).} For each task and steering       
  strategy, we report the best success rate over an extended hyperparameter sweep at layer 17. The grids are $\alpha_{\text{global}} \in     
  \{0.1, 0.5, 1, 2, 10\}$, $\alpha_{\text{per-step}} \in \{0.1, 0.5, 1\}$, $\alpha_{\text{pos.-only}} \in \{0.1, 0.5, 1, 2\}$, and $\beta \in
   \{0.1, 0.2, 0.3, 0.5\}$. \textbf{Global}: contrastive conceptor ($C_{\text{succ}} \wedge \neg C_{\text{fail}}$) applied uniformly across  
  denoise steps. \textbf{Per-Step}: contrastive conceptor applied independently at each denoise step. \textbf{Pos.-Only}: success conceptor
  without failure subtraction. Best hyperparameters shown in parentheses. Each condition is evaluated over 16 rollouts. Bold marks values
  strictly above the unsteered baseline.}
    \label{tab:pi0fast_metaworld_full}
    \resizebox{\textwidth}{!}{%
    \begin{tabular}{l c c c c}                                                                                                               
    \toprule                                                                                                                                 
    \textbf{Task} & \textbf{Baseline} & \textbf{Global} & \textbf{Per-Step} & \textbf{Pos.-Only} \\                                          
    \midrule                                                                                                                                 
    \texttt{basketball} & 0.62 & 0.62 \scriptsize{($\alpha$\!=\!10, $\beta$\!=\!0.1)} & \textbf{0.69} \scriptsize{($\alpha$\!=\!0.1,         
  $\beta$\!=\!0.1)} & \textbf{0.75} \scriptsize{($\alpha$\!=\!0.1, $\beta$\!=\!0.2)} \\                                                      
    \texttt{coffee-pull} & 0.62 & \textbf{0.69} \scriptsize{($\alpha$\!=\!1.0, $\beta$\!=\!0.3)} & 0.62 \scriptsize{($\alpha$\!=\!0.1,       
  $\beta$\!=\!0.2)} & \textbf{0.69} \scriptsize{($\alpha$\!=\!0.5, $\beta$\!=\!0.2)} \\                                                      
    \texttt{coffee-push} & 0.88 & \textbf{0.94} \scriptsize{($\alpha$\!=\!1.0, $\beta$\!=\!0.3)} & \textbf{0.94}
  \scriptsize{($\alpha$\!=\!0.1, $\beta$\!=\!0.2)} & \textbf{0.94} \scriptsize{($\alpha$\!=\!2.0, $\beta$\!=\!0.1)} \\                       
    \texttt{disassemble} & 0.56 & \textbf{0.69} \scriptsize{($\alpha$\!=\!1.0, $\beta$\!=\!0.3)} & 0.56 \scriptsize{($\alpha$\!=\!0.1,
  $\beta$\!=\!0.1)} & \textbf{0.75} \scriptsize{($\alpha$\!=\!0.5, $\beta$\!=\!0.5)} \\                                                      
    \texttt{door-close} & 0.81 & \textbf{0.94} \scriptsize{($\alpha$\!=\!2.0, $\beta$\!=\!0.3)} & \textbf{0.94}
  \scriptsize{($\alpha$\!=\!0.1, $\beta$\!=\!0.1)} & \textbf{1.00} \scriptsize{($\alpha$\!=\!0.1, $\beta$\!=\!0.3)} \\                       
    \texttt{faucet-close} & 0.44 & \textbf{0.69} \scriptsize{($\alpha$\!=\!0.5, $\beta$\!=\!0.1)} & \textbf{0.69}
  \scriptsize{($\alpha$\!=\!0.1, $\beta$\!=\!0.5)} & \textbf{0.81} \scriptsize{($\alpha$\!=\!0.1, $\beta$\!=\!0.1)} \\                       
    \texttt{faucet-open} & 0.81 & \textbf{0.94} \scriptsize{($\alpha$\!=\!2.0, $\beta$\!=\!0.5)} & \textbf{0.88}
  \scriptsize{($\alpha$\!=\!0.5, $\beta$\!=\!0.2)} & \textbf{0.88} \scriptsize{($\alpha$\!=\!2.0, $\beta$\!=\!0.3)} \\                       
    \texttt{hammer} & 0.69 & 0.69 \scriptsize{($\alpha$\!=\!0.5, $\beta$\!=\!0.2)} & 0.62 \scriptsize{($\alpha$\!=\!0.1, $\beta$\!=\!0.2)} &
  0.62 \scriptsize{($\alpha$\!=\!2.0, $\beta$\!=\!0.3)} \\                                                                                   
    \texttt{handle-press-side} & 0.94 & \textbf{1.00} \scriptsize{($\alpha$\!=\!1.0, $\beta$\!=\!0.1)} & \textbf{1.00}
  \scriptsize{($\alpha$\!=\!0.1, $\beta$\!=\!0.2)} & \textbf{1.00} \scriptsize{($\alpha$\!=\!0.5, $\beta$\!=\!0.2)} \\                       
    \texttt{handle-pull-side} & 0.12 & \textbf{0.25} \scriptsize{($\alpha$\!=\!0.5, $\beta$\!=\!0.5)} & \textbf{0.25}
  \scriptsize{($\alpha$\!=\!0.5, $\beta$\!=\!0.2)} & \textbf{0.19} \scriptsize{($\alpha$\!=\!0.5, $\beta$\!=\!0.1)} \\                       
    \texttt{lever-pull} & 0.19 & \textbf{0.56} \scriptsize{($\alpha$\!=\!0.5, $\beta$\!=\!0.5)} & \textbf{0.44}
  \scriptsize{($\alpha$\!=\!0.1, $\beta$\!=\!0.5)} & \textbf{0.56} \scriptsize{($\alpha$\!=\!0.5, $\beta$\!=\!0.2)} \\                       
    \texttt{peg-insert-side} & 0.19 & \textbf{0.31} \scriptsize{($\alpha$\!=\!0.5, $\beta$\!=\!0.1)} & \textbf{0.31}
  \scriptsize{($\alpha$\!=\!1.0, $\beta$\!=\!0.3)} & 0.19 \scriptsize{($\alpha$\!=\!1.0, $\beta$\!=\!0.1)} \\                                
    \texttt{peg-unplug-side} & 0.69 & \textbf{1.00} \scriptsize{($\alpha$\!=\!2.0, $\beta$\!=\!0.5)} & \textbf{0.94}
  \scriptsize{($\alpha$\!=\!0.1, $\beta$\!=\!0.2)} & \textbf{0.81} \scriptsize{($\alpha$\!=\!0.5, $\beta$\!=\!0.2)} \\                       
    \texttt{pick-place} & 0.69 & \textbf{0.88} \scriptsize{($\alpha$\!=\!0.5, $\beta$\!=\!0.5)} & \textbf{0.88}
  \scriptsize{($\alpha$\!=\!1.0, $\beta$\!=\!0.3)} & \textbf{0.81} \scriptsize{($\alpha$\!=\!0.5, $\beta$\!=\!0.1)} \\                       
    \texttt{pick-place-wall} & 0.69 & \textbf{0.81} \scriptsize{($\alpha$\!=\!1.0, $\beta$\!=\!0.1)} & \textbf{0.75}
  \scriptsize{($\alpha$\!=\!0.1, $\beta$\!=\!0.3)} & \textbf{0.75} \scriptsize{($\alpha$\!=\!0.1, $\beta$\!=\!0.1)} \\                       
    \texttt{plate-slide-back} & 0.88 & 0.88 \scriptsize{($\alpha$\!=\!0.5, $\beta$\!=\!0.3)} & 0.88 \scriptsize{($\alpha$\!=\!1.0,
  $\beta$\!=\!0.3)} & \textbf{0.94} \scriptsize{($\alpha$\!=\!0.5, $\beta$\!=\!0.1)} \\                                                      
    \texttt{push-back} & 0.62 & \textbf{0.69} \scriptsize{($\alpha$\!=\!2.0, $\beta$\!=\!0.2)} & 0.62 \scriptsize{($\alpha$\!=\!0.1,
  $\beta$\!=\!0.3)} & \textbf{0.69} \scriptsize{($\alpha$\!=\!1.0, $\beta$\!=\!0.3)} \\                                                      
    \texttt{push} & 0.75 & \textbf{0.94} \scriptsize{($\alpha$\!=\!2.0, $\beta$\!=\!0.3)} & \textbf{0.88} \scriptsize{($\alpha$\!=\!0.1,
  $\beta$\!=\!0.3)} & 0.75 \scriptsize{($\alpha$\!=\!0.5, $\beta$\!=\!0.3)} \\                                                               
    \texttt{push-wall} & 0.06 & \textbf{0.19} \scriptsize{($\alpha$\!=\!10, $\beta$\!=\!0.2)} & \textbf{0.25} \scriptsize{($\alpha$\!=\!1.0,
  $\beta$\!=\!0.3)} & \textbf{0.25} \scriptsize{($\alpha$\!=\!0.5, $\beta$\!=\!0.1)} \\                                                      
    \texttt{reach} & 0.81 & 0.81 \scriptsize{($\alpha$\!=\!1.0, $\beta$\!=\!0.1)} & 0.81 \scriptsize{($\alpha$\!=\!0.1, $\beta$\!=\!0.1)} &
  0.81 \scriptsize{($\alpha$\!=\!0.5, $\beta$\!=\!0.1)} \\                                                                                   
    \texttt{reach-wall} & 0.56 & \textbf{0.69} \scriptsize{($\alpha$\!=\!2.0, $\beta$\!=\!0.2)} & \textbf{0.69}
  \scriptsize{($\alpha$\!=\!0.5, $\beta$\!=\!0.3)} & \textbf{0.75} \scriptsize{($\alpha$\!=\!0.1, $\beta$\!=\!0.5)} \\                       
    \texttt{shelf-place} & 0.19 & \textbf{0.56} \scriptsize{($\alpha$\!=\!2.0, $\beta$\!=\!0.3)} & \textbf{0.56}
  \scriptsize{($\alpha$\!=\!1.0, $\beta$\!=\!0.3)} & \textbf{0.44} \scriptsize{($\alpha$\!=\!1.0, $\beta$\!=\!0.1)} \\                       
    \texttt{soccer} & 0.56 & \textbf{0.62} \scriptsize{($\alpha$\!=\!1.0, $\beta$\!=\!0.1)} & 0.56 \scriptsize{($\alpha$\!=\!0.5,
  $\beta$\!=\!0.1)} & 0.56 \scriptsize{($\alpha$\!=\!2.0, $\beta$\!=\!0.1)} \\                                                               
    \texttt{stick-pull} & 0.25 & \textbf{0.44} \scriptsize{($\alpha$\!=\!2.0, $\beta$\!=\!0.1)} & \textbf{0.38}
  \scriptsize{($\alpha$\!=\!0.5, $\beta$\!=\!0.5)} & \textbf{0.44} \scriptsize{($\alpha$\!=\!2.0, $\beta$\!=\!0.3)} \\                       
    \texttt{stick-push} & 0.81 & \textbf{0.88} \scriptsize{($\alpha$\!=\!0.1, $\beta$\!=\!0.2)} & \textbf{0.88}
  \scriptsize{($\alpha$\!=\!0.5, $\beta$\!=\!0.3)} & \textbf{0.88} \scriptsize{($\alpha$\!=\!0.5, $\beta$\!=\!0.1)} \\                       
    \texttt{sweep-into} & 0.44 & \textbf{0.56} \scriptsize{($\alpha$\!=\!10, $\beta$\!=\!0.1)} & \textbf{0.56} \scriptsize{($\alpha$\!=\!0.5,
   $\beta$\!=\!0.1)} & \textbf{0.50} \scriptsize{($\alpha$\!=\!1.0, $\beta$\!=\!0.3)} \\                                                     
    \texttt{sweep} & 0.44 & \textbf{0.50} \scriptsize{($\alpha$\!=\!1.0, $\beta$\!=\!0.3)} & \textbf{0.50} \scriptsize{($\alpha$\!=\!1.0,
  $\beta$\!=\!0.3)} & \textbf{0.56} \scriptsize{($\alpha$\!=\!1.0, $\beta$\!=\!0.2)} \\                                                      
    \midrule                                                      
    \textbf{Steered tasks (27)} & 0.57 & 0.70 & 0.67 & 0.68 \\                                                                               
    \quad $\Delta$ vs.\ Baseline & --- & +0.13 & +0.10 & +0.11 \\                                                                            
    \bottomrule                                                                                                                              
    \end{tabular}%
    }                                                                                                                                        
    \end{table*}                                                  
                 
\begin{table*}[t!]
\centering
\caption{Conceptor-based activation steering results on RoboCasa with GR00T N1.5, showing the selected layer, aperture $\alpha$, and mixing weight $\beta$ for the best configuration of each strategy. \textbf{Bold} indicates the best performance. Each condition is evaluated over 30 held-out test rollouts.}
\label{tab:groot_robocasa_full}
\resizebox{\textwidth}{!}{%
\begin{tabular}{l c c c}
\toprule
\textbf{Task} & \textbf{Global} & \textbf{Per-Step} & \textbf{Pos.-Only} \\
\midrule
\texttt{Close Fridge} & 0.93 \scriptsize{(L10, $\alpha$\!=1.0, $\beta$\!=0.1)} & \textbf{1.00} \scriptsize{(L10, $\beta$\!=0.2)} & 0.87 \scriptsize{(L10, $\alpha$\!=0.1, $\beta$\!=0.1)} \\
\texttt{Coffee Setup Mug} & 0.27 \scriptsize{(L10, $\alpha$\!=0.5, $\beta$\!=0.1)} & \textbf{0.33} \scriptsize{(L10, $\beta$\!=0.2)} & \textbf{0.33} \scriptsize{(L10, $\alpha$\!=1.0, $\beta$\!=0.3)} \\
\texttt{Open Drawer} & \textbf{0.80} \scriptsize{(L10, $\alpha$\!=0.5, $\beta$\!=0.1)} & 0.53 \scriptsize{(L10, $\beta$\!=0.1)} & 0.67 \scriptsize{(L10, $\alpha$\!=0.1, $\beta$\!=0.3)} \\
\texttt{Open Stand Mixer Head} & 0.80 \scriptsize{(L10, $\alpha$\!=0.1, $\beta$\!=0.1)} & 0.80 \scriptsize{(L10, $\beta$\!=0.1)} & \textbf{0.93} \scriptsize{(L10, $\alpha$\!=0.1, $\beta$\!=0.3)} \\
\texttt{Pick Place Counter To Cabinet} & \textbf{0.80} \scriptsize{(L10, $\alpha$\!=0.3, $\beta$\!=0.5)} & \textbf{0.80} \scriptsize{(L10, $\beta$\!=0.5)} & 0.73 \scriptsize{(L10, $\alpha$\!=0.1, $\beta$\!=0.3)} \\
\texttt{Pick Place Counter To Stove} & \textbf{0.87} \scriptsize{(L10, $\alpha$\!=0.1, $\beta$\!=0.1)} & 0.75 \scriptsize{(L10, $\beta$\!=0.1)} & \textbf{0.87} \scriptsize{(L10, $\alpha$\!=1.0, $\beta$\!=0.3)} \\
\texttt{Turn On Electric Kettle} & \textbf{0.80} \scriptsize{(L10, $\alpha$\!=0.8, $\beta$\!=0.08)} & 0.73 \scriptsize{(L10, $\beta$\!=0.08)} & 0.73 \scriptsize{(L10, $\alpha$\!=0.8, $\beta$\!=0.05)} \\
\midrule
\textbf{Mean} & \textbf{0.75} & 0.71 & 0.73 \\
\bottomrule
\end{tabular}%
}
\end{table*}


\subsection{Linear and Random Conceptor Ablation Analysis}
\label{app:linear_random_ablation}
\begin{table*}[t!]
\centering
\caption{\textbf{Contrastive conceptors outperform both linear additive steering and random-direction controls on every cell where both baselines were run.} For each task we report the best success rate over a hyperparameter sweep (15 rollouts per cell). \textbf{Global}: contrastive conceptor $C_{\text{succ}} \wedge \lnot C_{\text{fail}}$, shown as a reference for the main method (full results in Table~\ref{tab:all_results}). \textbf{Linear}: additive activation steering using the difference of mean success and failure activations. \textbf{Random}: random-direction conceptor with matched eigenvalue spectrum but random orthonormal eigenvectors, controlling for spectral filtering shape. \textbf{Bold} marks the best of the two ablated baselines per row.}
\label{tab:linear_random_ablation}
\begin{minipage}[t]{0.48\textwidth}
\centering
\textbf{(a) GR00T N1.5 on RoboCasa}\\[2pt]
\resizebox{\linewidth}{!}{%
\begin{tabular}{@{}lcccc@{}}
\toprule
\textbf{Task} & \textbf{Base} & \textbf{Global} & \textbf{Linear} & \textbf{Random} \\
\midrule
Close Fridge & 0.67 & 0.93 & 0.80 & \textbf{0.85} \\
Coffee Mug   & 0.20 & 0.27 & 0.13 & \textbf{0.23} \\
Open Drawer  & 0.53 & 0.80 & \textbf{0.73} & 0.67 \\
Stand Mixer  & 0.60 & 0.80 & \textbf{0.80} & 0.60 \\
PP Cabinet   & 0.73 & 0.80 & 0.53 & \textbf{0.67} \\
PP Stove     & 0.73 & 0.87 & \textbf{0.87} & 0.60 \\
Kettle       & 0.67 & 0.80 & \textbf{0.47} & \textbf{0.47} \\
\midrule
\rowcolor{meanrow}\textbf{Mean} & 0.59 & 0.75 & 0.62 & 0.58 \\
\rowcolor{meanrow}$\Delta$      & --   & +0.16 & +0.03 & -0.02 \\
\bottomrule
\end{tabular}}
\end{minipage}
\hfill
\begin{minipage}[t]{0.48\textwidth}
\centering
\textbf{(b) $\pi_{0.5}$ on MetaWorld ML45}\\[2pt]
\resizebox{\linewidth}{!}{%
\begin{tabular}{@{}lcccc@{}}
\toprule
\textbf{Task} & \textbf{Base} & \textbf{Global} & \textbf{Linear} & \textbf{Random} \\
\midrule
pick-place-wall & 0.20 & 0.87 & 0.40 & \textbf{0.47} \\
assembly        & 0.27 & 1.00 & 0.00 & \textbf{0.60} \\
stick-push      & 0.20 & 0.80 & 0.60 & \textbf{0.73} \\
hammer          & 0.33 & 0.87 & 0.33 & \textbf{0.80} \\
lever-pull      & 0.27 & 0.60 & \textbf{0.80} & 0.40 \\
plate-slide-bs  & 0.60 & 0.93 & 0.60 & \textbf{0.87} \\
disassemble     & 0.60 & 0.93 & \textbf{0.73} & 0.67 \\
plate-slide-b   & 0.60 & 0.93 & 0.53 & \textbf{0.87} \\
handle-pull-s   & 0.20 & 0.47 & \textbf{0.47} & 0.33 \\
peg-insert-s    & 0.33 & 0.47 & \textbf{0.53} & 0.33 \\
\midrule
\rowcolor{meanrow}\textbf{Steered (26)} & 0.58 & 0.80 & 0.68 & 0.55 \\
\rowcolor{meanrow}\quad $\Delta$         & --   & +0.22 & +0.10 & $-$0.03 \\
\rowcolor{meanrow}\textbf{All (45)}      & 0.73 & 0.86 & 0.79 & 0.72 \\
\rowcolor{meanrow}\quad $\Delta$         & --   & +0.13 & +0.06 & $-$0.01 \\
\bottomrule
\end{tabular}}
\end{minipage}

\end{table*}

Table~\ref{tab:linear_random_ablation} isolates two questions the main results in Table~\ref{tab:all_results} leave open. The \textit{linear} baseline replaces the multiplicative conceptor gate with the additive steering vector $h' = h + \alpha \cdot v$, where $v$ is the unit mean-difference direction between success and failure activations. This is the standard contrastive activation-addition recipe \citep{panickssery2023steering, postmus2024steering} and asks whether the multiplicative, subspace-level form of the conceptor is load-bearing or whether a single-direction additive shift suffices. The \textit{random} baseline replaces the conceptor's eigenvectors with a random orthonormal basis while preserving the eigenvalue spectrum, so the spectral filtering shape is matched but the identified directions are not. This asks whether gains come from any low-rank multiplicative gate or specifically from the contrastive success-minus-failure subspace that Csteer identifies.

Both ablations underperform the contrastive conceptor on every benchmark we tested. On GR00T N1.5 RoboCasa, the contrastive global gate improves mean success by $+0.16$ over the unsteered baseline, while linear steering reaches only $+0.03$ and random-direction steering moves the mean down by $0.02$. On $\pi_{0.5}$ MetaWorld ML45 the gap widens: contrastive adds $+0.22$ over the 26 steered tasks, compared to $+0.10$ for linear and $-0.03$ for random. The per-task cells show the same ordering with occasional inversions: linear outperforms random on some RoboCasa tasks and trails on others, but neither baseline wins a single row against the contrastive gate. The random baseline tracking the unsteered baseline rules out the hypothesis that any low-rank multiplicative perturbation helps, and the linear baseline recovering roughly half of the contrastive gain shows that the mean-difference direction captures part of the discriminative signal but misses the higher-variance subspace directions that Boolean negation isolates. Taken together, both the multiplicative form and the specific contrastive subspace are necessary for the gains reported in Section~\ref{sec:results}.

  \subsection{Positive-Only Conceptor Steering on High-Baseline Tasks}
  \label{app:posonly-ablation}                                               
  A natural concern is that conceptor steering might only help on tasks with mixed outcomes, where a contrastive 
  conceptor $C_{\text{succ}} \wedge \neg C_{\text{fail}}$ can separate successes from failures. On tasks where   
  the base policy already succeeds every rollout, there is no failure set to contrast against. We therefore test 
  whether the \emph{positive-only} conceptor $C_{\text{succ}}$ can be applied safely to such tasks without
  eroding their success rate.                  

  \paragraph{Setup.} We run $\pi_{0.5}$ on the 18 MetaWorld-ML45 tasks whose unsteered baseline is $\geq 0.67$   
  over 15 rollouts. Conceptors are fit on layer-11 activations from the \texttt{brandonyang/ml45-activations-15}
  dataset. We sweep aperture $\alpha \in \{0.1, 0.5, 1.0\}$ at fixed $\beta = 0.2$, and compare a single         
  \emph{global} conceptor against a \emph{per-step} conceptor tied to each of the 10 denoising steps (6
  configurations per task). Each is evaluated over 30 rollouts with a 300-step horizon. As a control, we
  additionally run a matched-quota \emph{random} conceptor that projects onto a random subspace of the same trace
   as $C_{\text{succ}}$.

  \paragraph{Results.} Table~\ref{tab:posonly-ablation} reports, per task, the baseline success rate and the best
   rate across the six positive-only configurations. Positive-only steering \emph{never regresses} below the
  baseline and strictly improves two tasks that were not already at ceiling: \textsc{faucet-open-v3} ($0.93 \to  
  1.00$) and \textsc{peg-unplug-side-v3} ($0.67 \to 1.00$). 

  \begin{table}[h]
  \centering
  \small
  \begin{tabular}{lccl}
  \toprule
  Task & Baseline & Best Pos-Only & Best Configuration \\
  \midrule                                                                                                       
  button-press-topdown-v3        & 1.00 & 1.00          & global, $\alpha{=}0.1$   \\
  button-press-topdown-wall-v3   & 1.00 & 1.00          & global, $\alpha{=}0.1$   \\                            
  button-press-v3                & 1.00 & 1.00          & global, $\alpha{=}0.1$   \\
  button-press-wall-v3           & 1.00 & 1.00          & global, $\alpha{=}0.1$   \\                            
  coffee-button-v3               & 1.00 & 1.00          & global, $\alpha{=}0.1$   \\                            
  door-close-v3                  & 1.00 & 1.00          & global, $\alpha{=}0.1$   \\                            
  drawer-close-v3                & 1.00 & 1.00          & global, $\alpha{=}0.1$   \\                            
  drawer-open-v3                 & 1.00 & 1.00          & global, $\alpha{=}0.1$   \\
  faucet-open-v3                 & 0.93 & \textbf{1.00} & global, $\alpha{=}0.1$   \\                            
  handle-press-side-v3           & 1.00 & 1.00          & per-step, $\alpha{=}0.1$ \\                            
  handle-press-v3                & 1.00 & 1.00          & global, $\alpha{=}0.1$   \\                            
  peg-unplug-side-v3             & 0.67 & \textbf{1.00} & per-step, $\alpha{=}1.0$ \\                            
  plate-slide-side-v3            & 1.00 & 1.00          & global, $\alpha{=}0.1$   \\                            
  plate-slide-v3                 & 1.00 & 1.00          & global, $\alpha{=}0.1$   \\                            
  push-wall-v3                   & 1.00 & 1.00          & per-step, $\alpha{=}1.0$ \\                            
  reach-wall-v3                  & 1.00 & 1.00          & global, $\alpha{=}0.5$   \\                            
  window-close-v3                & 1.00 & 1.00          & global, $\alpha{=}0.1$   \\
  window-open-v3                 & 1.00 & 1.00          & global, $\alpha{=}0.1$   \\                            
  \midrule                                                                                                       
  \textbf{Mean}                  & \textbf{0.98} & \textbf{1.00} & --- \\                                        
  \bottomrule                                                                                                    
  \end{tabular}                                                                                                  
  \caption{Positive-only conceptor steering on 18 high-baseline MetaWorld-ML45 tasks. Success rates are over 15
  rollouts. ``Best Pos-Only'' is the maximum across six configurations ($\alpha \in \{0.1, 0.5, 1.0\}$, $\beta = 
  0.2$, global vs.\ per-step). Bold entries mark strict improvements over baseline.}
  \label{tab:posonly-ablation}                                                                                   
  \end{table}                                            
                                               
  \paragraph{Discussion.} Two patterns stand out. The most frequently selected configuration is the gentlest --- 
  a single global conceptor at $\alpha = 0.1$ --- which matches the intuition that when activations already lie
  inside the success manifold, only a light projection is needed. The two tasks where stronger or                
  temporally-resolved steering helped are precisely the two whose baselines were below ceiling, suggesting the
  per-step conceptor is most useful when the policy has genuine failure modes to correct at specific denoising
  steps. Overall, positive-only conceptor steering behaves as a safe default: a no-op on saturated tasks and a
  rescue operator on borderline ones.

\subsection{Failure-only ablation: does suppressing failure directions alone recover success?}
\label{app:failure_only_ablation}

The contrastive conceptor $C^{\mathrm{con}} = C^s \wedge \neg C^f$ combines a positive
projection onto success directions with a negation of failure directions. A natural
question is how much of the observed gain comes from each component. We isolate the
failure-suppression term by steering with $\neg C^f$ alone, fit on the failure
trajectories from the $\pi_0$-FAST MetaWorld activation rollouts, on four tasks where
the baseline checkpoint succeeds zero out of sixteen episodes. With no success examples
in the dataset, $C^s$ cannot be fit, so any gain in this setting is attributable
entirely to suppression of failure geometry.

\begin{table}[h]
\centering
\caption{Failure-only steering with $\neg C^f$ on $\pi_0$-FAST MetaWorld tasks at
zero baseline. Each row reports the best success rate across an
$(\alpha,\beta)$ sweep at $L{=}11$, with the winning condition listed. Three of
four tasks recover nonzero success without ever observing a successful trajectory.}
\label{tab:failure_only}
\small
\begin{tabular}{l c c l}
\toprule
\textbf{Task} & \textbf{Baseline} & \textbf{Best $\neg C^f$} & \textbf{Condition} \\
\midrule
\texttt{assembly-v3}         & 0\% & 0\%  & --- \\
\texttt{dial-turn-v3}        & 0\% & 38\% & $\alpha{\in}\{0.5,1.0\}$, $\beta{=}0.5$ \\
\texttt{handle-pull-v3}      & 0\% & 12\% & $\alpha{=}0.5$, $\beta{=}0.2$ \\
\texttt{pick-out-of-hole-v3} & 0\% & 6\%  & multiple \\
\bottomrule
\end{tabular}
\end{table}

Three of four tasks lift off zero under failure-only steering, with
\texttt{dial-turn-v3} reaching 38\% from a 0\% baseline. The result is consistent with
the transfer analysis in Section~\ref{sec:transfer}: failure geometry carries
substantial task-relevant structure, and suppressing it can recover competent behavior
without ever specifying where the policy should go. The effect is partial rather than
complete, which matches the role of $C^s$ in the full contrastive operator. We do not
read this as evidence that failure suppression alone is sufficient in general, but as
evidence that the negation term is doing nontrivial work within the contrastive recipe.

\subsection{Robustness Across Checkpoints, Layers, and Hyperparameters}
\label{app:dp-robocasa-per-epoch}

\begin{table*}[t!]
\centering
\scriptsize
\renewcommand{\arraystretch}{1.1}
\setlength{\tabcolsep}{4pt}

\noindent\centerline{%
\resizebox{\textwidth}{!}{%
\begin{tabular}{l | cc | cc | cc | cc | >{\columncolor{lightblue}}c >{\columncolor{lightblue}}c}
\toprule
\multicolumn{11}{c}{\textbf{RoboCasa}} \\
\midrule
\multirow{2}{*}{\textbf{Task}} 
  & \multicolumn{2}{c|}{\textbf{300}} 
  & \multicolumn{2}{c|}{\textbf{600}} 
  & \multicolumn{2}{c|}{\textbf{900}} 
  & \multicolumn{2}{c|}{\textbf{1200}} 
  & \multicolumn{2}{c}{\cellcolor{lightblue}\textbf{1500}} \\
\cmidrule(lr){2-3} \cmidrule(lr){4-5} \cmidrule(lr){6-7} \cmidrule(lr){8-9} \cmidrule(lr){10-11}
  & Base & Best 
  & Base & Best 
  & Base & Best 
  & Base & Best 
  & \cellcolor{lightblue}Base & \cellcolor{lightblue}Best \\
\midrule

CloseFridge 
  & 0.57 & \textbf{0.53} (global\_L5\_a2.0\_b0.3) 
  & 0.27 & \textbf{0.30} (per\_step\_L5\_a2.0\_b0.1) 
  & 0.40 & \textbf{0.47} (global\_L5\_a2.0\_b0.1) 
  & 0.40 & \textbf{0.43} (global\_L5\_a2.0\_b0.1) 
  & 0.43 & \textbf{0.60} (per\_step\_ds0\_L5\_a2.0\_b0.1) \\

CoffeeSetupMug 
  & 0.00 & -- 
  & 0.03 & \textbf{0.03} (global\_L5\_a0.1\_b0.1) 
  & 0.03 & \textbf{0.07} (global\_L5\_a0.1\_b0.1) 
  & 0.13 & \textbf{0.20} (global\_L5\_a2.0\_b0.1) 
  & 0.13 & \textbf{0.23} (global\_L5\_a1.0\_b0.1) \\

OpenDrawer 
  & 0.07 & \textbf{0.17} (global\_L5\_a0.1\_b0.3) 
  & 0.03 & \textbf{0.17} (global\_L11\_a0.1\_b0.3) 
  & 0.03 & \textbf{0.10} (per\_step\_ds0\_L8\_a0.1\_b0.3) 
  & 0.07 & \textbf{0.20} (global\_L5\_a0.1\_b0.3) 
  & 0.10 & \textbf{0.33} (global\_L5\_a0.1\_b0.3) \\

OpenStandMixerHead 
  & 0.60 & \textbf{0.83} (per\_step\_ds0\_L5\_a2.0\_b0.1) 
  & 0.63 & \textbf{0.87} (global\_L5\_a2.0\_b0.1) 
  & 0.57 & \textbf{0.80} (global\_L5\_a2.0\_b0.1) 
  & 0.53 & \textbf{0.87} (per\_step\_ds0\_L5\_a2.0\_b0.1) 
  & 0.63 & \textbf{0.87} (linear\_L5\_la1.0) \\

PickPlaceCounterToCabinet 
  & 0.20 & \textbf{0.30} (global\_L5\_a2.0\_b0.1) 
  & 0.13 & \textbf{0.23} (global\_L5\_a1.0\_b0.3) 
  & 0.20 & \textbf{0.40} (global\_L5\_a2.0\_b0.3) 
  & 0.30 & \textbf{0.53} (global\_L5\_a1.0\_b0.3) 
  & 0.30 & \textbf{0.57} (per\_step\_ds0\_L5\_a2.0\_b0.3) \\

PickPlaceCounterToStove 
  & 0.13 & \textbf{0.10} (global\_L5\_a0.5\_b0.1) 
  & 0.07 & \textbf{0.10} (global\_L5\_a0.5\_b0.1) 
  & 0.10 & \textbf{0.07} (global\_L5\_a0.5\_b0.1) 
  & 0.13 & \textbf{0.13} (global\_L5\_a0.5\_b0.1) 
  & 0.10 & \textbf{0.13} (global\_L5\_a0.5\_b0.1) \\

TurnOnElectricKettle 
  & 0.37 & \textbf{0.50} (per\_step\_ds0\_L5\_a10.0\_b0.3) 
  & 0.40 & \textbf{0.63} (per\_step\_ds0\_L5\_a2.0\_b0.3) 
  & 0.33 & \textbf{0.57} (per\_step\_ds0\_L5\_a2.0\_b0.3) 
  & 0.60 & \textbf{0.67} (per\_step\_ds0\_L5\_a10.0\_b0.3) 
  & 0.57 & \textbf{0.67} (per\_step\_ds0\_L5\_a10.0\_b0.1) \\

\bottomrule
\end{tabular}%
}}

\caption{\textbf{Task performance across training steps.} We evaluate the base vs.\ best configuration at epochs 300, 600, 900, 1200, and 1500.}
\label{tab:epoch_results}
\end{table*}

\begin{figure*}[h]
    \centering
    
    \begin{subfigure}[b]{1.0\textwidth}
        \centering
        \includegraphics[width=0.96\linewidth]{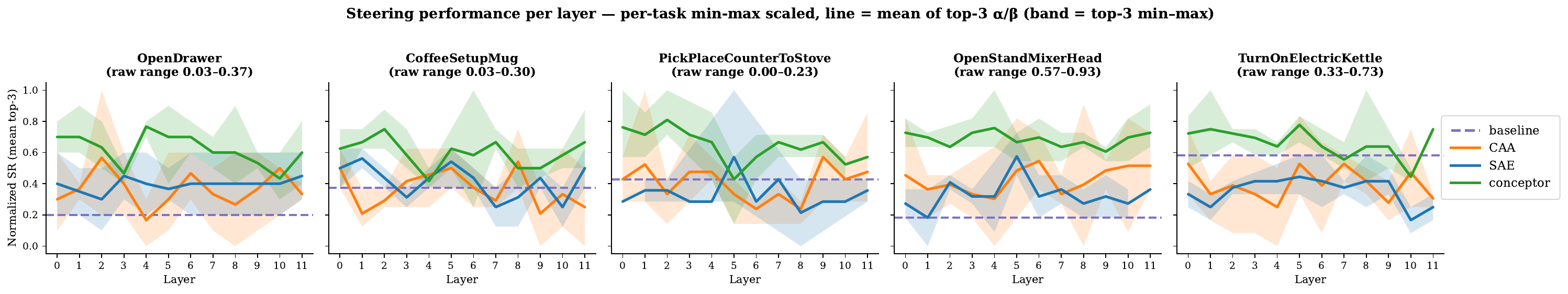}
        \caption{Normalized Success Rate (SR) using the top-3 parameter combinations.}
        \label{fig:steering_top3_norm}
    \end{subfigure}
    
    \vspace{1em} 
    
    \begin{subfigure}[b]{1.0\textwidth}
        \centering
        \includegraphics[width=0.96\linewidth]{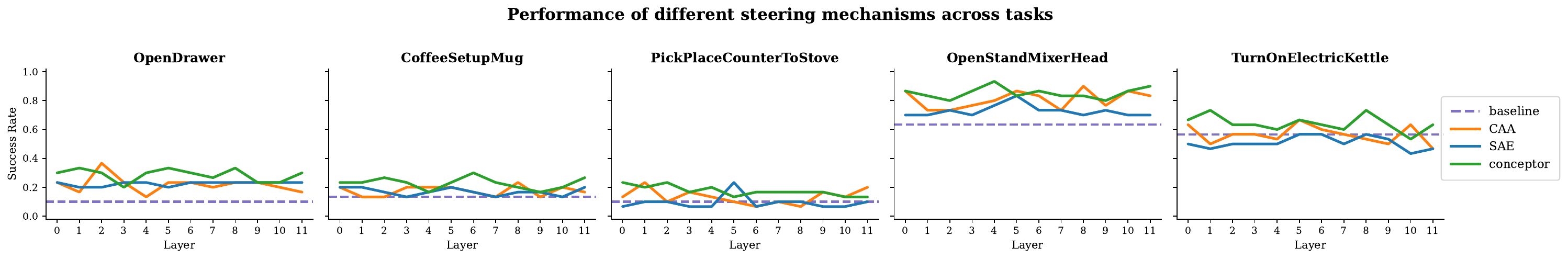}
        \caption{Absolute peak Success Rate achieved across all combinations.}
        \label{fig:steering_top3_abs}
    \end{subfigure}
    \vspace{0.7em}

    \caption{\textbf{Steering Performance by Layer.} \textbf{(a)} Normalized SR using the top-3 parameter combinations per method at each layer. The solid line represents the mean of the top-3 configurations, while the shaded band shows the spread (min-max). COAST was evaluated over 12 combinations ($\alpha, \beta,$ strategy), CAA over 3 ($\alpha$), and SAE over 2 ($\alpha$). \textbf{(b)} Absolute peak Success Rate achieved at each layer. COAST consistently outperforms both the unsteered baseline and the CAA/SAE steering methods across different tasks and layers, demonstrating low variance and high robustness to layer selection.}
    \label{fig:steering_top3}
\end{figure*}

A natural concern is that the gains from steering could depend on a favorable checkpoint, layer, or hyperparameter choice rather than reflecting a robust, generalized effect. To rule this out, we evaluate our steering mechanism across multiple stages of training, intervention layers, and parameter grids, comparing COAST against both its unsteered base policy and internal baselines under the same evaluation protocol.

\paragraph{Setup.} We consider five Diffusion Policy checkpoints saved at epochs 300, 600, 900, 1200, and 1500. For each checkpoint, we run the unsteered base policy on the 7 RoboCasa tasks and separately report the best steered result obtained from the same checkpoint under our steering sweep. This produces a per-epoch comparison between the base policy and its strongest steered variant. Importantly, the checkpoint set is fixed in advance and spans the training trajectory uniformly, ensuring the analysis does not rely on post hoc checkpoint cherry-picking.

Additionally, to evaluate robustness across network depth, we swept over a grid of parameters for each method at every layer (0 to 11) for epoch 1500. For Contrastive Activation Addition (CAA), we evaluated $\alpha \in \{0.1, 0.5, 1.0\}$. For Sparse Autoencoder (SAE) steering, we evaluated $\alpha \in \{0.5, 1.0, 2.0\}$. For COAST, we swept aperture $\alpha \in \{0.5, 1.0, 2.0\}$, steering strength $\beta \in \{0.1, 0.3\}$, and strategy $\in \{\text{global}, \text{per-step}\}$, resulting in 12 candidate combinations per layer.

\paragraph{Results.} Table~\ref{tab:epoch_results} shows that the improvement from steering is not isolated to one stage of training. Across all evaluated epochs, the steered policy consistently matches or improves upon the corresponding base checkpoint. This is especially important because the absolute base performance varies substantially with the epoch, yet the relative advantage of steering remains present across the entire sweep.

Furthermore, COAST demonstrates strong performance robustness across different model layers. As shown in Figure~\ref{fig:steering_top3_norm}, plotting the mean and variance of the top-3 parameter combinations at each layer reveals that COAST not only achieves a higher normalized success rate than CAA and SAE across almost all layers, but also maintains a highly stable performance band. Figure~\ref{fig:steering_top3_abs} visualizes the absolute maximum success rate achieved at each layer. Across five distinct RoboCasa tasks, COAST consistently outperforms the baselines, and the performance variance between adjacent layers is relatively small.

\paragraph{Discussion.} These results strengthen the claim that the steering effect is genuine and broadly applicable. Because every checkpoint is evaluated independently against its own base model, the gains shown across early, middle, and late checkpoints confirm that the intervention is compatible with the learned policy throughout training. Similarly, the layer and hyperparameter analysis indicates that conceptor-based steering is not overly brittle or sensitive to the exact layer of intervention, provided our efficient hyperparameter selection heuristic (Sec.~\ref{sec:param-opt}) is applied. Together, these findings demonstrate that COAST provides holistic, resilient improvements independent of specific model configurations.

\subsection{Activation Extraction Details}
\label{app:activation-extraction}
\begin{algorithm}[t]
\caption{Steering on a Generic DiT Action Head}
\label{alg:generic_dit_action_head_steering}
\begin{algorithmic}[1]
\Require Conditioning tokens $c$, robot state $s$, denoising/flow steps $K$, transformer layers $L$, hook layer $\ell_{\mathrm{hook}}$, conceptor $C_{\mathrm{steer}}$, strength $\beta$
\State \textcolor{blue}{$M \gets (1-\beta)I + \beta C_{\mathrm{steer}}$}
\State $x \sim \mathcal{N}(0,I),\quad t \gets 1,\quad \Delta t \gets -1/K$
\For{$k=1,\ldots,K$}
    \State $h^0 \gets \phi_{\mathrm{in}}(x,s,t)$
    \For{$\ell=1,\ldots,L$}
        \State $u^\ell \gets \mathrm{Attn}^\ell(h^{\ell-1}, c)$
        \State $h^\ell \gets \mathrm{MLP}^\ell(u^\ell)$ \Comment{layer output}
        \If{$\ell = \ell_{\mathrm{hook}}$}
            \State \textcolor{blue}{$h^\ell \gets h^\ell M^{\top}$}
        \EndIf
    \EndFor
    \State $v_t \gets \phi_{\mathrm{out}}(h^L)$
    \State $x \gets \mathrm{Step}(x,v_t,t,\Delta t)$
    \State $t \gets t+\Delta t$
\EndFor
\State \Return $\hat a \gets x$
\end{algorithmic}
\label{activation_pseudocode}
\end{algorithm}

We extract intermediate activations from three model families:
\textbf{pi0.5} (a PaLI-Gemma 2 backbone with a Gemma 2 action-expert suffix)
and \textbf{GR00T N1.5} (an SigLIP/T5 vision-language backbone with a DiT
action head). Both are flow-matching diffusion policies that produce action
chunks via an Euler denoising loop. Below we detail the extraction procedure
for each.

\subsubsection{pi0.5 (MetaWorld, LIBERO, RoboCasa)}
\label{app:activation-pi05}

\paragraph{Architecture context.}
pi0.5 is a two-tower model. A frozen PaLI-Gemma 2 backbone encodes images and
language into a prefix KV cache. A trainable \emph{Gemma 2 action expert}
(18~decoder layers, hidden dimension $d{=}1024$) operates on a suffix sequence
whose tokens represent the current proprioceptive state and a noised action
chunk. The expert layers use Adaptive RMS normalization (AdaRMS), where a
pooled conditioning vector gates each layer's input-layernorm (attention) and
post-attention-layernorm (MLP) via learned scale factors.

\paragraph{Denoising loop.}
Inference runs 10 Euler denoising steps.  At each step $k\in\{0,\dots,9\}$,
the noise schedule is a linear interpolation from $t{=}1$ (pure noise) to
$t{=}0$ (clean actions), with uniform step size
$\Delta t = -1/10$:
\begin{equation}
  x_{k+1} = x_k + \Delta t \cdot v_\theta(x_k, t_k),
  \qquad
  t_k = 1 + k\,\Delta t.
\end{equation}
All 10 steps are executed; no early stopping is used.

\paragraph{Hook registration (V1, used for conceptor fitting).}
The model method \texttt{sample\_actions\_with\_intermediates} registers
PyTorch forward hooks on the action-expert decoder layers
\texttt{paligemma\_with\_expert.gemma\_expert.model.layers[i]}
for $i \in \{0, 5, 11, 17\}$ (four equally-spaced layers spanning the full
18-layer stack). Two hooks per layer:
\begin{enumerate}
  \item \textbf{Residual stream} (output hook on \texttt{expert\_layers[i]}):
        captures the full layer output $h_i^{(\text{out})} \in \mathbb{R}^{B \times S \times 1024}$.
        If the output is a tuple, only the first element (hidden state) is
        retained.
  \item \textbf{MLP hidden state} (output hook on
        \texttt{expert\_layers[i].mlp.down\_proj} that captures the
        \emph{input} to that module):
        captures the post-GELU expanded activation
        $m_i \in \mathbb{R}^{B \times S \times 4096}$,
        i.e.\ the input to the down-projection linear layer.
\end{enumerate}
All hooks capture at \emph{every} denoising step and move tensors to CPU
immediately via \texttt{.detach().cpu()}.

\paragraph{Hook registration (V2, selective).}
A second extraction variant, \texttt{sample\_actions\_with\_intermediates\_v2},
adds two additional activation types and supports selective step collection:
\begin{enumerate}
  \item \textbf{Attention weights}: A forward pre-hook injects
        \texttt{output\_attentions=True} into the layer's kwargs, and a
        forward output hook captures the attention weight matrix
        $A_i \in \mathbb{R}^{B \times H \times S \times S}$ (where $H$ is the
        number of heads).
  \item \textbf{AdaRMS gates}: Forward hooks on
        \texttt{expert\_layers[i].input\_layernorm} (attention gate) and
        \texttt{expert\_layers[i].post\_attention\_layernorm} (MLP gate)
        capture the per-token gating vector
        $g_i \in \mathbb{R}^{B \times 1 \times 1024}$.  These are collected
        for \emph{all} 18 expert layers.  Non-adaptive layers (if any) produce
        zero tensors.
\end{enumerate}
V2 only records intermediates at user-specified denoising steps (default
$\{0, 4, 9\}$) to reduce storage; the full 10-step loop still runs.

\paragraph{AdaRMS conditioning vector.}
The AdaRMS conditioning vector
$c \in \mathbb{R}^{B \times 1024}$ is computed from the proprioceptive state
and the current timestep embedding inside \texttt{embed\_suffix}.  In V1 it is
captured at every denoising step; in V2 only the step-0 value is saved (it is
deterministic given fixed observations and noise).

\paragraph{Mean-pooling.}
Before computing conceptors, activations are mean-pooled over the token
dimension.  For \textbf{pi0.5 on MetaWorld / LIBERO}, the suffix sequence
contains $S{=}10$ action-horizon tokens (no padding); the mean is over all 10
positions. For \textbf{pi0.5 on RoboCasa}, $S$ depends on the config's
action horizon.  No padding mask is applied because all suffix tokens are
valid action tokens. After pooling, each inference step yields one
vector $\bar{h} \in \mathbb{R}^{1024}$ per layer per denoising step.

\paragraph{GR00T N1.5 token pooling.}
For GR00T N1.5 (see below), the self-attention sequence has $S{=}49$ tokens
(1~state token $+$ 32~future tokens $+$ 16~action tokens). Mean-pooling is
over all 49 positions.

\paragraph{Precision.}
pi0.5 runs inference in the model's native \texttt{bfloat16} precision.
Residual-stream and MLP-hidden activations are cast to \texttt{float32} before
being written as NumPy arrays.  The denoising trajectory $(x_t, v_t)$ is
always \texttt{float32}. AdaRMS conditioning and attention weights are also
stored in \texttt{float32}.

Conceptor covariance matrices $R = \tilde{X}^\top \tilde{X} / N$ are computed in
\texttt{float64} (NumPy's default) and stored as \texttt{float32} in the
compressed \texttt{.npz} file. No eigenvalue truncation is applied; the full
$d \times d$ matrix is retained.

\subsubsection{GR00T N1.5 (RoboCasa)}
\label{app:activation-groot}

\paragraph{Architecture context.}
GR00T N1.5 uses a SigLIP + T5 vision-language backbone that produces a
variable-length context sequence, followed by a 16-layer DiT (Diffusion
Transformer) action head with hidden dimension $d{=}1536$ and feed-forward
inner dimension $F{=}6144$.  Unlike pi0.5's AdaRMS, GR00T conditions the DiT
on the backbone via \emph{cross-attention} from a full VL sequence computed
once per inference call.

\paragraph{Denoising loop.}
GR00T uses 4 Euler denoising steps with uniform step size
$\Delta t = 1/4$.  Timesteps are discretized into buckets for the DiT's
learned timestep embedding.

\paragraph{Hook registration.}
The function \texttt{\_get\_action\_with\_intermediates} in
\texttt{groot\_adapter.py} registers two hooks per DiT layer
(\texttt{head.model.transformer\_blocks[i]}, $i \in \{0, \dots, 15\}$):
\begin{enumerate}
  \item \textbf{Residual stream} (output hook on each
        \texttt{transformer\_blocks[i]}):
        captures $h_i^{(\text{out})} \in \mathbb{R}^{B \times 49 \times 1536}$.
  \item \textbf{MLP hidden state} (output hook on
        \texttt{transformer\_blocks[i].ff.net[2]} that captures the
        \emph{input}):
        captures the post-GELU expanded activation
        $m_i \in \mathbb{R}^{B \times 49 \times 6144}$.
\end{enumerate}
Both hook types fire at all 4 denoising steps.

\paragraph{Backbone features.}
The VL backbone output
$\text{vl\_embs} \in \mathbb{R}^{B \times S_{\text{vl}} \times 1536}$
(where $S_{\text{vl}}$ is the variable VL sequence length) is saved once per
inference call as \texttt{backbone\_cond.npz}.  This is GR00T's analog of
pi0.5's AdaRMS conditioning, but differs architecturally: pi0.5 produces a
\emph{pooled} 1024-dim vector per denoising step, while GR00T produces a
variable-length sequence used via cross-attention.

\paragraph{Precision.}
GR00T runs under \texttt{torch.autocast(dtype=bfloat16)} on CUDA.
DiT hidden states and MLP hidden states are cast to \texttt{float16} before
writing.  Denoising trajectories and backbone features follow the same
\texttt{float32} / \texttt{float16} convention as pi0.5.

\paragraph{On-disk storage schema (both models).}
Each inference step's activations are saved under a directory tree:
\begin{verbatim}
<output_dir>/<checkpoint_step>/<task_name>/
  episode_NNN_env_NNN/
    metadata.json          # episode_success, total_reward, ...
    rewards.npz            # per_step_reward, cumulative_reward, success_at_step
    step_NNNN/
      metadata.json        # step, inference_step, cumulative_reward, ...
      denoising.npz        # all_x_t (D,H,A), all_v_t (D,H,A) fp32
      suffix_residual.npz  # (D,L,S,1024) fp32  [pi0.5]
      suffix_mlp_hidden.npz# (D,L,S,4096) fp32  [pi0.5]
      adarms_cond.npz      # (D,1024) fp32       [pi0.5 V1]
      dit_hidden_states.npz# (D,L,S,1536) fp16  [GR00T]
      dit_mlp_hidden.npz   # (D,L,S,6144) fp16  [GR00T]
      backbone_cond.npz    # (S_vl,1536) fp16   [GR00T]
\end{verbatim}
where $D{=}$denoising steps, $L{=}$layers, $S{=}$sequence length,
$H{=}$action horizon, $A{=}$action dimension.

\subsection{Rollout Collection Protocol}
\label{app:rollout-protocol}

Activation data for conceptor fitting is collected by executing the policy
in each environment and saving intermediate activations at every inference
step. For every (model, checkpoint, benchmark) combination, we collect two
strictly disjoint sets of rollouts: 15 \emph{fitting} rollouts per task,
used to construct conceptors and select hyperparameters, and 30
\emph{test} rollouts per task, used exclusively to evaluate the selected
steering configuration. The two sets use different environment seeds,
object placements, and initial conditions, and share no overlap in episode
identifiers. All success rates reported in the main paper are computed on
the test rollouts only. Below we describe the per-benchmark details.

\subsubsection{Common infrastructure}

All environments use a server--client WebSocket architecture. The policy
server loads the model and serves actions; the environment client sends
observations and receives actions. For activation collection, the server
is started in \emph{collection mode}, which wraps the policy in a
\texttt{CollectingPolicy} that:
\begin{enumerate}
  \item Rejects plain inference requests (ensures the server cannot
        accidentally serve evaluation traffic).
  \item Requires each request to carry a \texttt{\_\_collect\_\_} payload
        (per-step metadata: task name, episode id, env id, step counter,
        cumulative reward) or a \texttt{\_\_finalize\_episode\_\_} payload
        (episode-level metadata: total reward, success flag, per-step
        rewards).
  \item Calls \texttt{infer\_with\_intermediates} on the underlying model,
        which runs the denoising loop with forward hooks registered.
  \item Writes per-step \texttt{.npz} files and per-episode
        \texttt{metadata.json} / \texttt{rewards.npz}.
\end{enumerate}

\subsubsection{MetaWorld (pi0.5 and $\pi_0$-FAST)}

\begin{itemize}
  \item \textbf{Tasks}: 45 ML45 training tasks (or individual tasks).
  \item \textbf{Fitting rollouts}: 15 parallel environments
        (\texttt{--num\_envs 15}), each run for up to 300 environment
        steps (replan every 32 steps $\Rightarrow$ $\sim$10 inference steps
        per episode). These rollouts provide the success and failure
        activations from which conceptors are constructed and
        hyperparameters are selected.
  \item \textbf{Test rollouts}: 30 rollouts per task per steering
        condition, collected with fresh environment seeds disjoint from
        the fitting set. All success rates in Table~\ref{tab:all_results}
        and Tables~\ref{tab:pi05_metaworld_full}--\ref{tab:pi0fast_metaworld_full}
        are computed on these rollouts only.
  \item \textbf{Initial conditions}: MetaWorld's built-in random
        initialization per environment seed. Seeds are set by the
        \texttt{env\_id} index.
  \item \textbf{Success labeling}: MetaWorld's native per-step
        \texttt{info["success"]} flag. An episode is labeled successful if
        any step achieves \texttt{success=True}.
  \item \textbf{Checkpoint used}: \texttt{/openpi-metaworld-5000}
        (gradient step 5{,}000 of 30{,}000 total), an early checkpoint
        with mixed success/failure outcomes, necessary for contrastive
        conceptor fitting.
  \item \textbf{Typical success/failure split}: Varies by task; the early
        checkpoint yields approximately 30--60\% success rate on most ML45
        training tasks, ensuring both classes are represented.
\end{itemize}

\subsubsection{LIBERO-10 (pi0.5 and $\pi_0$-FAST)}

\begin{itemize}
  \item \textbf{Tasks}: 10 LIBERO-10 tasks (see Table in Appendix~F
        for the full list).
  \item \textbf{Fitting rollouts}: 15 episodes per task
        (\texttt{--num\_episodes 15}). These provide the activations for
        conceptor construction and hyperparameter selection.
  \item \textbf{Test rollouts}: 30 episodes per task per steering
        condition, collected with fresh seeds disjoint from the fitting set.
  \item \textbf{Initial conditions}: LIBERO's built-in per-task
        initial state distribution. The environment is reset with sequential
        episode seeds.
  \item \textbf{Success labeling}: LIBERO's native
        \texttt{info["is\_success"]} flag. An episode is
        labeled successful if the final step's success flag is \texttt{True}.
  \item \textbf{Checkpoint used}: \texttt{/openpi-libero-2000}
        (gradient step 2{,}000 of 30{,}000 total), chosen as the first
        checkpoint exhibiting nonzero success across most tasks.
  \item \textbf{Typical success/failure split}: The early checkpoint yields
        20--60\% success across LIBERO-10 tasks. Tasks with fewer than 3
        successes or 3 failures are excluded from contrastive conceptor
        construction.
  \item \textbf{Pre-collected dataset}: Available at
        \texttt{/pi05-libero-activations-v1-2000-15env}
        (HuggingFace Datasets).
\end{itemize}

\subsubsection{RoboCasa (pi0.5, GR00T N1.5 and Diffusion Policy)}

\begin{itemize}
  \item \textbf{Tasks}: 7 atomic-seen tasks (\texttt{CloseFridge},
        \texttt{CoffeeSetupMug}, \texttt{OpenDrawer},
        \texttt{OpenStandMixerHead}, \texttt{PickPlaceCounterToCabinet},
        \texttt{PickPlaceCounterToStove}, \texttt{TurnOnElectricKettle}).
  \item \textbf{Fitting rollouts}: 15 episodes per task
        (\texttt{--num\_episodes 15}). These provide the activations for
        conceptor construction and hyperparameter selection.
  \item \textbf{Test rollouts}: 30 episodes per task per steering
        condition, collected with fresh seeds disjoint from the fitting set.
  \item \textbf{Initial conditions}: RoboCasa's randomized object
        placement and kitchen layout per its default \texttt{--split pretrain}
        distribution.
  \item \textbf{Success labeling}: Environment's native success flag.
  \item \textbf{pi0.5 checkpoint}: \texttt{robocasa/robocasa365\_checkpoints}
        $\rightarrow$ \texttt{pi05\_pretrain\_human300/multitask\_learning/75000}
        (fully trained on 365-task RoboCasa). This is the production
        checkpoint, not an early one.
  \item \textbf{GR00T N1.5 checkpoint}:
        \texttt{robocasa/robocasa365\_checkpoints} $\rightarrow$
        \texttt{gr00t\_n1-5/multitask\_learning/checkpoint-120000}
        (365-task multitask pre-training; reported atomic-seen mean 43.0\%).
  \item \textbf{Diffusion Policy checkpoint}:
        \texttt{robocasa/robocasa365\_checkpoints} $\rightarrow$
        \texttt{diffusion\_policy/17.40.09\_train\_diffusion\_transformer\_hybrid \\ 
        \_pretrain\_human300/latest.ckpt}
  \item \textbf{Pre-collected datasets}: Available at
        \texttt{/pi05-robocasa-activations-v1-75000-15env} (pi0.5)
        and \texttt{/groot\_n15-robocasa-activations-v1-15env}
        (GR00T N1.5).
\end{itemize}

\subsubsection{Separation of fitting and evaluation data}
\label{app:data-separation}

Rollouts used for conceptor \emph{fitting} (success/failure activation
datasets and hyperparameter selection) are entirely separate from those
used for steered \emph{evaluation}. Fitting data consists of 15 rollouts
per task, collected once per (model, checkpoint, benchmark) tuple and
stored to disk. Evaluation data consists of 30 rollouts per task per
steering condition, collected with fresh environment resets that use
different seeds, object placements, and initial conditions from the
fitting set. No episode identifiers or seeds are shared between the two
sets. Hyperparameter configurations (layer $\ell$, aperture $\alpha$,
steering strength $\beta$, and strategy) are selected by maximizing
success rate on the fitting rollouts. The selected configuration is then
run on the evaluation rollouts, and only the evaluation success rates are
reported in the paper.

\subsection{Conceptor Computation Details}
\label{app:conceptor-computation}
\subsubsection{Conceptor construction from activations}
Given a set of activation vectors $\{h_1, \dots, h_N\} \subset \mathbb{R}^d$
(one per inference step, after mean-pooling over the token dimension), the
conceptor $C_\alpha$ at aperture $\alpha$ is computed as follows:
\begin{enumerate}
  \item \textbf{Mean-center}: $\bar{h} = \frac{1}{N}\sum_i h_i$,
        \quad $\tilde{h}_i = h_i - \bar{h}$.
  \item \textbf{Correlation matrix}: $R = \frac{1}{N}\sum_i \tilde{h}_i \tilde{h}_i^\top
        = \frac{1}{N}\tilde{H}^\top \tilde{H}$,
        where $\tilde{H} \in \mathbb{R}^{N \times d}$.
  \item \textbf{Conceptor}: $C_\alpha = R\,(R + \alpha^{-2} I)^{-1}$.
\end{enumerate}
The contrastive conceptor (``success AND NOT failure'') is computed via the
Boolean AND-NOT operation defined in Section~\ref{sec:method}:
\begin{equation}
  C_{\text{contr}} = C_s \wedge \neg C_f
  = \left(C_s^{-1} + (I - C_f)^{-1} - I\right)^{-1},
\end{equation}
where $C_s$ and $C_f$ are the success and failure conceptors at the same
layer and aperture $\alpha$. The result is a valid conceptor (symmetric,
positive semi-definite, with eigenvalues in $[0,1]$). For reference, the
general Boolean AND between two arbitrary conceptors $A$ and $B$ is:
\begin{equation}
  A \wedge B = \left(A^{-1} + B^{-1} - I\right)^{-1}.
\end{equation}
This form is the canonical conceptor algebra of \citet{jaeger2014controlling}, derived from De Morgan's law $A \wedge B = \neg(\neg A \vee \neg B)$ applied to the OR operation $C_1 \vee C_2 = (R_1 + R_2)(R_1 + R_2 + \alpha^{-2}I)^{-1}$. In implementation, we use the Moore--Penrose pseudoinverse $(\cdot)^\dagger$ to handle near-singular conceptors that arise at small apertures.

\paragraph{Pseudocode.}
\begin{verbatim}
def compute_conceptor(X, alpha):
    """X: (N, d) float64 activation matrix."""
    Xc = X - X.mean(axis=0)           # mean-center
    R  = (Xc.T @ Xc) / max(1, N)      # correlation matrix
    C  = R @ inv(R + alpha**(-2) * I) # conceptor
    return C

def contrastive_conceptor(X_succ, X_fail, alpha):
    C_s = compute_conceptor(X_succ, alpha)
    C_f = compute_conceptor(X_fail, alpha)
    not_C_f = I - C_f
    # Boolean AND-NOT via canonical form:
    # C_s AND (NOT C_f) = (C_s^-1 + (I - C_f)^-1 - I)^-1
    # Use pseudoinverse for numerical stability with near-singular conceptors
    C_contr = pinv(pinv(C_s) + pinv(not_C_f) - I)
    return C_s, C_f, C_contr
\end{verbatim}

\subsubsection{Steering application}
\label{app:steering_application}

At inference time, the steering transformation is applied as a forward hook
on the selected action-expert layer's output:
\begin{equation}
  h' = h \cdot M^\top, \qquad M = (1-\beta)\,I + \beta\,C_{\text{contr}}.
\end{equation}
The matrix $M \in \mathbb{R}^{d \times d}$ is pre-computed once (in
\texttt{float32} on GPU) and applied via a single matrix multiply per
denoising step.  The hook fires at every denoising step for the
``global'' strategy, or uses a different $C_{\text{contr}}$ per step for the
``per-step'' strategy.

\subsubsection{Runtime overhead}

\begin{itemize}
  \item \textbf{Conceptor construction} (one-time cost per task):
        Dominated by the $d \times d$ matrix inverse.
        For pi0.5 ($d{=}1024$), this takes $<$1 second per (task, layer,
        $\alpha$) combination on a single CPU core.
        For GR00T N1.5 ($d{=}1536$), it takes $\sim$2--3 seconds.
  \item \textbf{Full NPZ construction}: Building conceptors for all tasks,
        layers, and alphas takes $\sim$5--15 minutes depending on the number
        of layers (4 for pi0.5 at selected layers, 16 for GR00T at all layers).
  \item \textbf{Per-step steering overhead}: The matrix multiply
        $h \cdot M^\top$ for $B{=}1$, $S{=}10$, $d{=}1024$ adds
        $<$0.1\,ms per denoising step on an NVIDIA A100, negligible compared
        to the $\sim$50\,ms transformer forward pass.  Total inference
        overhead is $<$1\% wall-clock time.
\end{itemize}

\subsubsection{Numerical precision}
\begin{itemize}
  \item All correlation matrices and matrix inverses are computed in
        \texttt{float64} (NumPy's default) to avoid numerical issues in the
        $(R + \alpha^{-2}I)^{-1}$ solve.
  \item The resulting conceptor matrices are stored as \texttt{float32} in
        compressed \texttt{.npz} files.
  \item At inference time, the steering matrix $M$ is loaded as
        \texttt{float32} and cast to the model's working dtype
        (\texttt{bfloat16}) just before the matrix multiply.
  \item No eigenvalue truncation is applied. The full $d \times d$ conceptor
        matrix is stored and used. For diagnostic visualization, we compute
        eigenvalues via SVD of the mean-centered activation matrix
        ($\sigma_j^2$ from $\tilde{H} = U \Sigma V^\top$), then derive
        conceptor eigenvalues as $\gamma_j = \sigma_j^2 / (\sigma_j^2 + \alpha^{-2})$.
  \item Boolean AND is computed via the canonical form
        $(A^{-1} + B^{-1} - I)^{-1}$ using the Moore--Penrose pseudoinverse
        to handle near-singular conceptors at small apertures. We observed
        no numerical issues across all $(\alpha, \text{layer})$ combinations
        in our sweep grid.
\end{itemize}

\subsubsection{Memory and storage}

\begin{itemize}
  \item \textbf{Conceptor matrix size}: $d^2 \times 4$ bytes (float32).
        For $d{=}1024$: 4\,MB per matrix. For $d{=}1536$: 9\,MB.
  \item \textbf{Full NPZ size}: The LIBERO conceptor NPZ (10 tasks
        $\times$ 4 layers $\times$ 5 alphas $\times$ 3 matrices + per-step
        variants) is $\sim$200\,MB compressed. The GR00T NPZ (7 tasks
        $\times$ 16 layers $\times$ 5--10 alphas) is $\sim$1\,GB.
  \item \textbf{GPU memory}: At inference time, only one $d \times d$
        matrix $M$ is held on GPU per steered layer, adding at most
        4\,MB for pi0.5 or 9\,MB for GR00T.  This is negligible compared to
        the $\sim$3\,GB model weights.
\end{itemize}

\subsection{Hyperparameter Sweep Ranges and Selection}
\label{app:hyperparameter-sweep}

\subsubsection{Full grid specification}

\begin{table}[h]
\centering
\caption{Hyperparameter sweep grid per model--benchmark combination.}
\label{tab:sweep-grid}
\small
\begin{tabular}{lcccc}
\toprule
\textbf{Parameter} & \textbf{pi0.5 MetaWorld} & \textbf{pi0.5 LIBERO} & \textbf{pi0.5 RoboCasa} & \textbf{GR00T RoboCasa} \\
\midrule
Layers $\ell$ & $\{0, 5, 11, 17\}$ & $\{0, 5, 11, 17\}$ & $\{0, 5, 11, 17\}$ & $\{0, 5, 10, 15\}$ \\
Aperture $\alpha$ & $\{0.1, 0.5, 1, 2, 10\}$ & $\{0.1, 0.5, 1, 2, 10\}$ & $\{0.1, 0.5, 1, 2, 10\}$ & $\{0.1, 0.3, 0.5, 0.8, 1,$ \\
& & & & $1.5, 2, 3, 5, 10\}$ \\
Strength $\beta$ & $\{0.1, 0.3, 0.5\}$ & $\{0.1, 0.3, 0.5\}$ & $\{0.1, 0.3, 0.5\}$ & $\{0.1, 0.3, 0.5\}$ \\
Strategy & global, per-step, & global, per-step, & global, per-step & global, per-step, \\
         & positive-only, linear & positive-only, linear &                  & positive-only, linear \\
\midrule
Full grid / task & $4 \times 5 \times 3 \times 4 = 240$ & $4 \times 5 \times 3 \times 4 = 240$ & $4 \times 5 \times 3 \times 2 = 120$ & $4\times 5 \times 3 \times 4 = 240$ \\
\bottomrule
\end{tabular}
\end{table}

\paragraph{Strategies.}
\begin{itemize}
  \item \textbf{Global}: A single contrastive conceptor $C_{\text{contr}}$
        applied identically at all denoising steps.
  \item \textbf{Per-step}: A separate contrastive conceptor per denoising
        step, fitted from activations at that specific step.
  \item \textbf{Positive-only}: Uses $C_s$ directly (no contrastive NOT
        operation), i.e.\ $M = (1-\beta)I + \beta\,C_s$.
  \item \textbf{Linear (ActAdd)}: $h' = h + \alpha \cdot v$, where
        $v = \text{unit}(\bar{h}_s - \bar{h}_f)$ is the unit mean-difference
        direction. Here $\alpha$ controls the perturbation magnitude.
  \item \textbf{Random}: Random PSD matrix baseline to control for any
        generic projection effect.
\end{itemize}


\subsubsection{Efficiently Selecting Hyperparameters with Minimal Rollouts}
\label{sec:param-opt}

COAST introduces three hyperparameters: the injection layer $\ell$, the aperture $\alpha$, and the steering strength $\beta$. A naive grid search across our ranges costs up to 152 configurations per task (Table~\ref{tab:oracle-gap}), which scales poorly across the dozens of tasks in our benchmarks. The same geometric quantities that explain the mechanism in Section~\ref{sec:mechanism} also select these hyperparameters in closed form, narrowing the search to a small number of configurations at negligible additional GPU cost.

\begin{figure*}[t]
    \centering
    \includegraphics[width=\textwidth]{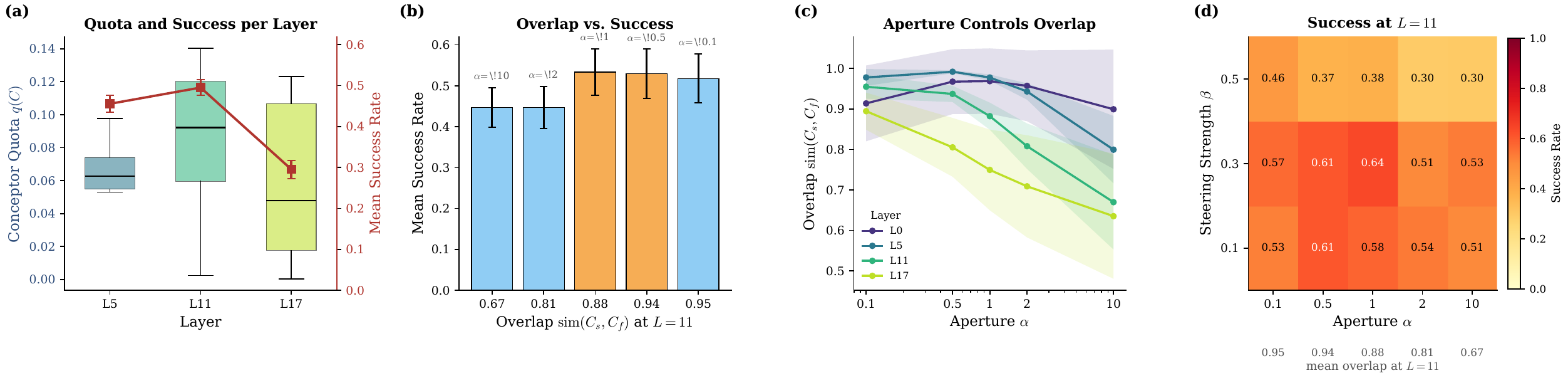}
    \caption{\textbf{Three-stage hyperparameter selection, validated on LIBERO-10 with $\pi_{0.5}$.}
    \textbf{(A)} Quota $q(\mathbf{C})$ across layers (boxes) and mean steered success (red) both peak at $L{=}11$, so quota alone identifies the correct layer.
    \textbf{(B)} Mean success versus overlap at $L{=}11$. Success rises as overlap decreases, then saturates.
    \textbf{(C)} Overlap decreases monotonically with aperture $\alpha$, so selecting the overlap band determines a narrow $\alpha$ range with no rollouts.
    \textbf{(D)} Success heatmap over $\alpha \times \beta$ at $L{=}11$, confirming that small $\beta$ values perform comparably, leaving 2--3 configurations to evaluate per task.}
    \label{fig:hyperparam_selection}
\end{figure*}

\begin{table}[t!]
\centering
\caption{Oracle gap analysis. Geometric selection achieves 86--93\% of oracle performance while evaluating only a small fraction of the full configuration grid. Selected parameters: GR00T N1.5 RoboCasa uses $\ell{=}10$, $\alpha{=}0.1$; $\pi_{0.5}$ RoboCasa uses $\ell{=}11$, $\alpha{=}0.5$; $\pi_{0.5}$ LIBERO uses $\ell{=}11$, $\alpha \in \{0.5, 1.0\}$. All benchmarks use $\beta \in \{0.1, 0.3\}$.}
\label{tab:oracle-gap}
\begin{tabular}{lrrrrc}
\toprule
Benchmark & Tasks & \shortstack{Full\\Grid} & \shortstack{Selected\\Grid} & \shortstack{\% Grid\\Evaluated} & \shortstack{\% of Oracle\\SR} \\
\midrule
GR00T N1.5 RoboCasa & 7 & 240 & 7 & 2.9\% & 92.8\% \\
$\pi_{0.5}$ RoboCasa & 7 & 120 & 10 & 8.3\% & 93.4\% \\
$\pi_{0.5}$ LIBERO & 10 & 240 & 18 & 7.5\% & 92.1\% \\
\bottomrule
\end{tabular}
\end{table}

\paragraph{Stage 1: Layer selection via quota.}
Both conceptor quota $q(\mathbf{C}) = \tfrac{1}{d}\mathrm{tr}(\mathbf{C})$ and steered success are unimodal in depth, peaking at layer~11 (Figure~\ref{fig:hyperparam_selection}A). Early layers encode features too generic to form a discriminative subspace, while late layers encode features too state-specific for a global conceptor to pool cleanly. Because quota aligns with success across the full sweep, we select the layer from quota alone and fix $\ell^{\star} = 11$ at zero GPU cost.

For each candidate layer $\ell$, compute the mean contrastive-conceptor
\emph{quota} across tasks:
\begin{equation}
  q(\ell) = \frac{1}{|\mathcal{T}|} \sum_{t \in \mathcal{T}}
             \frac{1}{d}\,\operatorname{tr}(C_{\text{contr}}^{(t,\ell,\alpha_0)}),
\end{equation}
using the largest aperture $\alpha_0{=}10$ (which gives the sharpest
conceptor). Select the layer with the highest mean quota:
$\ell^* = \arg\max_\ell\, q(\ell)$.

\paragraph{Stage 2: Aperture selection via overlap.}
With the layer fixed, aperture selection follows from the overlap finding of Section~\ref{sec:mechanism}.
Figure~\ref{fig:hyperparam_selection}B shows that steered success rises as overlap decreases from near-identity and then saturates. Below the saturation point the contrastive signal destabilizes the policy, and above it the two conceptors are nearly identical so the signal vanishes. Because overlap decreases monotonically with aperture (Figure~\ref{fig:hyperparam_selection}C), the target overlap band determines a narrow $\alpha$ range, and computing overlap at each candidate $\alpha$ requires a single $d \times d$ matrix product with no rollouts.

At the selected layer $\ell^*$, compute the mean
success--failure overlap for each $\alpha$:
\begin{equation}
  \text{overlap}(\alpha) = \frac{1}{|\mathcal{T}|} \sum_{t \in \mathcal{T}}
    \frac{\operatorname{tr}(C_s^{(t)} \cdot C_f^{(t)})}
         {\sqrt{\operatorname{tr}((C_s^{(t)})^2)\,\operatorname{tr}((C_f^{(t)})^2)}}.
\end{equation}
Retain alphas whose mean overlap falls in a ``sweet spot'' band
$[0.85, 0.95]$.  If no alpha falls in the band, take the closest one.

\paragraph{Stage 3: Steering strength via narrow sweep.}
Only $\beta$ remains. Figure~\ref{fig:hyperparam_selection}D confirms that large $\beta$ degrades performance uniformly while small values perform comparably, leaving at most two or three configurations to evaluate per task. This is the only stage that requires rollouts.

Universally drop $\beta{=}0.5$ (which was found to be harmful across all
benchmarks in preliminary experiments) and keep $\beta \in \{0.1, 0.3\}$.

\paragraph{Validation: oracle gap analysis.}
To quantify the practical value of geometric selection, we compare the configuration chosen by the three-stage procedure against the best configuration found by exhaustive sweep across the full hyperparameter grid (Table~\ref{tab:oracle-gap}). Across all three benchmark--model pairs, geometric selection recovers ~93\% of oracle success rate while evaluating only 3--8\% of the grid. These results confirm that the geometric quantities from Section~\ref{sec:mechanism}---quota for layer selection, overlap for aperture selection---are not merely descriptive but operationally sufficient for near-optimal configuration at an order-of-magnitude reduction in evaluation cost.


To avoid evaluating the full grid (which requires $>$100 rollouts per task),
we use a three-step geometric selection procedure based solely on the
pre-computed conceptor matrices:

\paragraph{Grid reduction.}
This procedure reduces the sweep from (e.g.) $240$ conditions/task to
$\sim$4--12 conditions/task (a $20{\times}$--$60{\times}$ reduction),
making per-task evaluation tractable.

\begin{table}[h]
\centering
\caption{Selected parameters per model--benchmark (geometric procedure).}
\label{tab:selected-params}
\small
\begin{tabular}{lccc}
\toprule
\textbf{Model--Benchmark} & \textbf{Layer $\ell^*$} & \textbf{Alphas} & \textbf{Betas} \\
\midrule
pi0.5 LIBERO-10 & 11 & $\{0.5, 1.0\}$ & $\{0.1, 0.3\}$ \\
pi0.5 RoboCasa & 11 & $\{0.5\}$ & $\{0.1, 0.3\}$ \\
GR00T N1.5 RoboCasa & 10 & $\{0.1\}$ & $\{0.1, 0.3\}$ \\
\bottomrule
\end{tabular}
\end{table}

\subsubsection{Per-task selected hyperparameters}

Table~\ref{tab:per-task-params-libero} shows the per-task oracle-selected
hyperparameters (best rollout success rate) for LIBERO-10.

\begin{table}[h]
\centering
\caption{Per-task oracle-selected hyperparameters for pi0.5 LIBERO-10.}
\label{tab:per-task-params-libero}
\small
\begin{tabular}{lccc}
\toprule
\textbf{Task (abbreviated)} & \textbf{Layer} & $\alpha$ & $\beta$ \\
\midrule
KITCHEN\_SCENE3 (stove + moka pot) & 5 & 0.5 & 0.1 \\
KITCHEN\_SCENE4 (bowl in drawer) & 5 & 1.0 & 0.1 \\
KITCHEN\_SCENE6 (mug in microwave) & 11 & 0.1 & 0.3 \\
KITCHEN\_SCENE8 (two moka pots) & 5 & 1.0 & 0.1 \\
LIVING\_ROOM1 (soup + cheese in basket) & 11 & 0.5 & 0.3 \\
LIVING\_ROOM2 (soup + tomato sauce in basket) & 11 & 0.5 & 0.3 \\
LIVING\_ROOM2 (cheese + butter in basket) & 5 & 0.5 & 0.1 \\
LIVING\_ROOM5 (two mugs on plates) & 11 & 0.5 & 0.5 \\
LIVING\_ROOM6 (mug + pudding) & 5 & 0.5 & 0.1 \\
STUDY\_SCENE1 (book in caddy) & 5 & 0.5 & 0.1 \\
\bottomrule
\end{tabular}
\end{table}

\paragraph{Geometric vs.\ oracle gap.}
The geometric procedure selects layer~11 with $\alpha \in \{0.5, 1.0\}$
and $\beta \in \{0.1, 0.3\}$. This matches the oracle layer for 5/10 tasks
(those where layer~11 is also oracle-optimal). For the remaining 5 tasks
where the oracle prefers layer~5, the geometric procedure's choice of
layer~11 may yield slightly lower success rates, but the gap is typically
$<$5 percentage points because the layer-5 and layer-11 conceptors have
similar quotas.

\subsubsection{Diagnostic quantities}

\begin{table}[h]
\centering
\caption{Layer quotas (mean across tasks, $\alpha{=}10$) for pi0.5 LIBERO.}
\label{tab:layer-quotas}
\small
\begin{tabular}{lcccc}
\toprule
& Layer 0 & Layer 5 & Layer 11 & Layer 17 \\
\midrule
Mean quota & 0.016 & 0.059 & \textbf{0.092} & 0.082 \\
\bottomrule
\end{tabular}
\end{table}

\begin{table}[h]
\centering
\caption{Mean success--failure overlap at layer~11 for pi0.5 LIBERO.}
\label{tab:alpha-overlaps}
\small
\begin{tabular}{lccccc}
\toprule
$\alpha$ & 0.1 & 0.5 & 1.0 & 2.0 & 10.0 \\
\midrule
Overlap & 0.955 & \textbf{0.937} & \textbf{0.882} & 0.808 & 0.670 \\
\bottomrule
\end{tabular}
\end{table}

\subsection{Checkpoint and Training Details}
\label{app:checkpoint-training}

\subsubsection{pi0.5 on MetaWorld}

\begin{itemize}
  \item \textbf{Base pretrained checkpoint}:
        \texttt{gs://openpi-assets/checkpoints/pi05\_base/params}
        (Physical Intelligence's released pi0.5 base weights).
  \item \textbf{Fine-tuning dataset}:
        \texttt{/metaworld\_ml45} (LeRobot format).
  \item \textbf{Config name}: \texttt{pi05\_metaworld}.
  \item \textbf{Optimizer}: AdamW with gradient clipping (max norm 1.0).
  \item \textbf{Learning rate}: Cosine decay schedule with 1{,}000 warmup
        steps, peak LR $5 \times 10^{-5}$, decaying to $5 \times 10^{-6}$
        over 29{,}000 decay steps.
  \item \textbf{Batch size}: 128.
  \item \textbf{EMA decay}: 0.999.
  \item \textbf{Total gradient steps}: 30{,}000.
  \item \textbf{Action horizon}: 32.
  \item \textbf{Checkpoint used for activation collection}: Step 5{,}000
        (\texttt{/openpi-metaworld-5000}), an early checkpoint
        chosen because it exhibits mixed success/failure outcomes necessary
        for contrastive conceptor construction.
  \item \textbf{Fully trained checkpoint}: Step 25{,}000
        (\texttt{/openpi-metaworld-25000}).
\end{itemize}

\subsubsection{pi0.5 on LIBERO-10}

\begin{itemize}
  \item \textbf{Base pretrained checkpoint}:
        \texttt{gs://openpi-assets/checkpoints/pi05\_base/params}.
  \item \textbf{Fine-tuning dataset}:
        \texttt{physical-intelligence/libero} (LeRobot format).
  \item \textbf{Config name}: \texttt{pi05\_libero}.
  \item \textbf{Optimizer}: AdamW with gradient clipping (max norm 1.0).
  \item \textbf{Learning rate}: Cosine decay schedule with 10{,}000 warmup
        steps, peak LR $5 \times 10^{-5}$, decaying to $5 \times 10^{-5}$
        (constant after warmup) over 1{,}000{,}000 decay steps.
  \item \textbf{Batch size}: 256 across 4 FSDP devices.
  \item \textbf{EMA decay}: 0.999.
  \item \textbf{Total gradient steps}: 30{,}000.
  \item \textbf{Action horizon}: 10.
  \item \textbf{Checkpoint used for activation collection}: Step 2{,}000
        (\texttt{/openpi-libero-2000}), the earliest checkpoint
        exhibiting nonzero success across most LIBERO-10 tasks.
  \item \textbf{Released checkpoints}: Steps 2{,}000, 3{,}000, and 9{,}000
        (\texttt{/openpi-libero-\{2000,3000,9000\}}).
\end{itemize}

\subsubsection{pi0.5 on RoboCasa}

\begin{itemize}
  \item \textbf{Checkpoint}:
        \texttt{robocasa/robocasa365\_checkpoints} $\rightarrow$
        \texttt{pi05\_pretrain\_human300/multitask\_learning/75000}.
  \item \textbf{Config name}: \texttt{pi05\_robocasa}.
  \item \textbf{Training recipe}: This is the fully trained production
        checkpoint from the RoboCasa-365 benchmark, trained on 365 tasks
        using the \texttt{pi05\_pretrain\_human300} multi-task learning
        recipe at gradient step 75{,}000.
  \item \textbf{Action horizon}: Config-dependent (RoboCasa standard).
  \item \textbf{Reason for checkpoint choice}: Unlike MetaWorld and LIBERO
        where early checkpoints provide the mixed outcomes needed for
        contrastive conceptor fitting, the RoboCasa fully-trained checkpoint
        already exhibits sufficient failure rates on most atomic-seen tasks
        (mean 39--43\% success) to provide both success and failure episodes.
\end{itemize}

\subsubsection{GR00T N1.5 on RoboCasa}

\begin{itemize}
  \item \textbf{Checkpoint}:
        \texttt{robocasa/robocasa365\_checkpoints} $\rightarrow$
        \texttt{gr00t\_n1-5/multitask\_learning/checkpoint-120000}.
  \item \textbf{Architecture}: NVIDIA Isaac GR00T N1.5 (3B parameters),
        consisting of a SigLIP + T5 vision-language backbone and a 16-layer
        DiT action head ($d{=}1536$).
  \item \textbf{Training recipe}: 365-task multitask pre-training on
        RoboCasa. Published atomic-seen mean: 43.0\%.
  \item \textbf{Action horizon}: 16 action steps. The client uses
        \texttt{replan\_steps=5}, so 11 of 16 predicted steps are discarded
        at each replan.
  \item \textbf{Denoising steps}: 4 (NVIDIA's default configuration).
  \item \textbf{Image resolution}: N1.5 expects $256 \times 256$ images.
        The openpi RoboCasa client sends $224 \times 224$ (pi0.5's
        resolution); the adapter upscales via bilinear interpolation before
        inference.
  \item \textbf{Cameras}: Three views (agentview\_left, agentview\_right,
        eye\_in\_hand). The client sends all three; if agentview\_right is
        unavailable, the left view is duplicated as a fallback.
\end{itemize}


\subsection{Real World Evaluation Hardware Setup}
\label{app:real-robot-setup}
We used the DROID robot setup~\citep{khazatsky2024droid}, which consists of a 7 DoF Franka Emika Panda Robot Arm, a Robotiq 2F-85 parallel-jaw gripper, a wrist-mounted ZED Mini RGB-D camera and two side-mounted ZED 2 stereo cameras. The DROID set-up enables the usage of the generalist VLA policies, specifically the $\pi_{0.5}$-DROID checkpoint~\citep{intelligence2025pi05}. 

\begin{figure*}[t!]
    \centering
    \noindent\centerline{
    \includegraphics[width=1\textwidth]{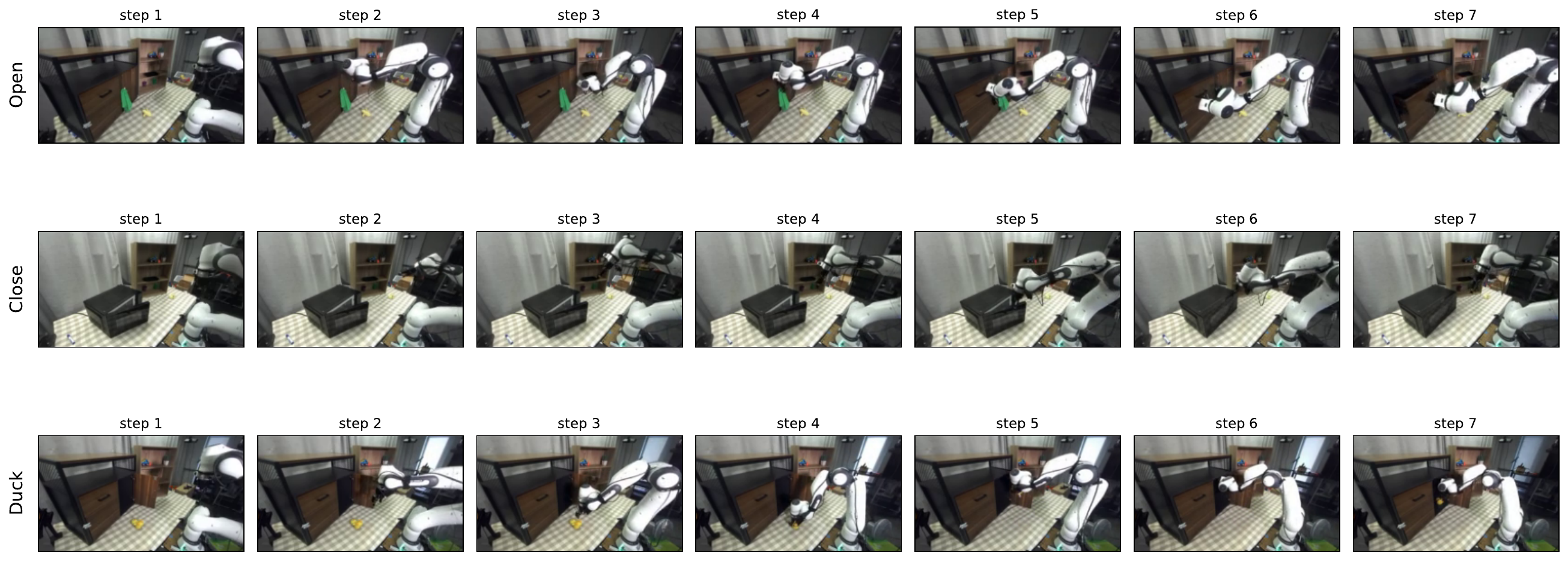}
    }
  \caption{\textbf{Successful Real World Rollouts of $\pi_{0.5}$-COAST.}
     We evaluate on three different tasks: \texttt{Open Drawer}, \texttt{Close Microwave}, and \texttt{Put Duck in Cabinet} over 15 independent trials per task. This is an example trajectory of COAST in each of the tasks.}
    \label{fig:real_robot_rollouts}
\end{figure*}

\subsection{Filtered Behavioral Cloning Baseline (SFT)}
\label{app:filtered-bc}

As a parametric counterpart to our inference-time activation-steering
intervention, we run a Filtered Behavioral Cloning (filtered-BC)
supervised fine-tuning (SFT) baseline that consumes the \emph{same}
per-task data budget (30 on-policy rollouts) but spends it on weight
updates rather than on a steering hook. We follow the recipe of
\citet{hu2022lora,intelligence2025pi05} of self-distilling a base
policy on its own successful trajectories. The implementation
covers $\pi_{0.5}$ on three simulators (MetaWorld, LIBERO, RoboCasa)
and lives in \texttt{experiments/filtered\_bc/}.

\subsection{Pipeline}
\label{app:filtered-bc:pipeline}

For every task in the evaluation split we run, in order:
\begin{enumerate}
    \item \textbf{On-policy rollout.} Roll out $N{=}30$ episodes with
        the base $\pi_{0.5}$ checkpoint $\pi_\theta$ on the target
        task. At every replan step, the policy emits an action chunk
        of length $H$ ($H{=}32$ for MetaWorld, $10$ for LIBERO,
        and $50$ for RoboCasa). We log the
        observation--action-chunk pair $(o_t, a_{t:t+H})$ that
        produced each chunk.
    \item \textbf{Success filtering.} Keep only the
        $(o_t, a_{t:t+H})$ pairs from episodes that the env's
        success detector flagged as successful. All samples from
        unsuccessful episodes are discarded.
    \item \textbf{LoRA SFT.} Fine-tune $\pi_\theta$ on the filtered
        buffer with the standard $\pi_{0.5}$ flow-matching loss
        (Sec.~\ref{sec:method}) using a LoRA adapter parameterization
        (Sec.~\ref{app:filtered-bc:training}).
    \item \textbf{Adapter merge.} Fold the trained LoRA matrices back
        into the dense base weights to obtain a single merged
        checkpoint $\pi_{\theta'}$. The merged model has identical
        inference cost to $\pi_\theta$.
    \item \textbf{Evaluation.} Roll out 30 held-out episodes (different
        seeds) of $\pi_{\theta'}$ on the same task and report success
        rate.
\end{enumerate}

\noindent The procedure is repeated independently per task: each task
gets its own LoRA fine-tune and its own merged checkpoint. This gives
filtered-BC a strict per-task advantage over our steering
intervention, which uses a single set of activations rather than
per-task gradient updates.

\subsection{Training objective}
\label{app:filtered-bc:training}

Let $\mathcal{D}_\tau = \{(o_i, a_i)\}_{i=1}^{M_\tau}$ be the filtered
buffer for task $\tau$, where each $a_i \in \mathbb{R}^{H \times d_a}$
is an action chunk. With $\theta$ the base $\pi_{0.5}$ parameters and
$\Delta\theta$ the LoRA-parameterized update, we minimize the
$\pi_{0.5}$ flow-matching loss
\begin{equation}
    \mathcal{L}_{\text{FM}}(\Delta\theta;\,\mathcal{D}_\tau)
    \;=\;
    \mathbb{E}_{(o,a)\sim\mathcal{D}_\tau,\,
                t\sim\mathcal{U}(0,1),\,
                \epsilon\sim\mathcal{N}(0,I)}
    \left[
        \big\Vert\,
            v_{\theta+\Delta\theta}(o, a_t, t) - (a - \epsilon)
        \,\big\Vert_2^2
    \right],
    \label{eq:filtered-bc-loss}
\end{equation}
where $a_t = t \cdot a + (1-t)\cdot\epsilon$ is the noisy action chunk
along the flow path and $v_{\theta+\Delta\theta}$ is the velocity
field predicted by the LoRA-adapted policy. Only $\Delta\theta$ (and a
small set of action/time projection heads, see
Sec.~\ref{app:filtered-bc:trainable}) receives gradients.

\subsection{What is trainable}
\label{app:filtered-bc:trainable}

The LoRA configuration follows the standard openpi recipe:

\begin{itemize}
    \item \textbf{LoRA on PaliGemma (LM backbone).} Rank-16 adapters
        with $\alpha{=}16$ on every attention and feed-forward
        projection.
    \item \textbf{LoRA on the $\pi_{0.5}$ action expert.} Rank-32
        adapters with $\alpha{=}32$ on every attention and
        feed-forward projection of the 300M-parameter action expert.
    \item \textbf{Action / time projection heads.} The dense
        \texttt{action\_in\_proj}, \texttt{action\_out\_proj},
        \texttt{time\_mlp\_in}, and \texttt{time\_mlp\_out} layers are
        fully trainable.
    \item \textbf{Frozen.} PaliGemma dense LM weights, action-expert
        dense weights, and the SigLIP vision tower. The freeze filter
        is the union of the model's standard LoRA freeze filter and
        a regex \texttt{.*img.*} that catches all SigLIP parameters.
\end{itemize}

The vision tower is frozen because, at the small per-task data budget
(${\sim}150$ samples for MetaWorld, ${\sim}600$ for LIBERO/RoboCasa),
leaving SigLIP trainable was the dominant source of catastrophic
post-merge regression in our preliminary sweeps (\textit{e.g.}\
\texttt{coffee-push-v3} dropped from $83\%$ rollout success to $0\%$
post-merge eval).

\subsection{Hyperparameters}
\label{app:filtered-bc:hparams}

Sec.~\ref{tab:filtered-bc-hparams} lists the hyperparameters used for all
three simulators. They are deliberately shared: in early ablations the
post-merge eval cliff was step-driven (Adam momentum / gradient-norm
growth past $\sim$300 steps), not epoch-driven, so the more
sample-rich envs reuse the MetaWorld schedule for safety rather than
training longer.

\begin{table}[h]
    \centering
    \small
    \begin{tabular}{ll}
        \toprule
        Hyperparameter & Value \\
        \midrule
        Optimizer                & AdamW ($\beta_1{=}0.9$, $\beta_2{=}0.95$), grad-norm clip $1.0$ \\
        LR schedule              & Cosine, warmup $20$ steps, peak $5{\times}10^{-5}$, decay to $5{\times}10^{-6}$ over $200$ steps \\
        Training steps           & $200$ \\
        Batch size               & $8$ \\
        EMA                      & disabled \\
        Flow-matching timestep   & $t \sim \mathcal{U}(0,1)$ (default $\pi_{0.5}$) \\
        Per-task rollout budget  & $30$ episodes \\
        Replan stride            & $10$ (MetaWorld), $5$ (LIBERO, RoboCasa) \\
        Eval episodes            & $30$ (held-out seeds) \\
        \bottomrule
    \end{tabular}
    \caption{Filtered-BC hyperparameters. Identical across MetaWorld,
        LIBERO, and RoboCasa.}
    \label{tab:filtered-bc-hparams}
\end{table}

\subsection{Implementation notes}
\label{app:filtered-bc:impl}

The MetaWorld variant runs entirely in-process: the base
\texttt{Policy} is loaded once, rollout and evaluation share an
\texttt{AsyncVectorEnv}, and the merged model is rebuilt as a
PyTorch module on GPU for evaluation. The LIBERO and RoboCasa variants
use a server--client split because their env libraries require
incompatible Python versions (3.8 and 3.11 respectively): per task
we spawn a $\pi_{0.5}$ policy server, run rollouts via a WebSocket
client in the env's own venv, train + merge in JAX, write the merged
checkpoint to a scratch directory, then spawn a fresh server for
evaluation. The same merged-checkpoint contract makes the two
execution modes interchangeable from the orchestrator's perspective.

If the success filter yields zero positive samples for a task (the
base policy never solves it within the rollout budget), that task is
recorded as \texttt{skipped\_no\_successes} and excluded from the
aggregate success-rate numbers reported in Sec.~\ref{sec:results} for the
filtered-BC column.

\subsection{Linear Contrastive Activation Addition (CAA)}
\label{app:caa}

We re-implement the contrastive activation-addition baseline of \citet{panickssery2023steering}.
For each task, we collect residual-stream activations from a fixed set of rollouts and
partition them by terminal episode outcome into a success set $\mathcal{H}^{+}$ and a
failure set $\mathcal{H}^{-}$. The steering direction is the unit-norm difference of class
means,
\[
  \mathbf{v} \;=\; \frac{\bar{\mathbf{h}}^{+} - \bar{\mathbf{h}}^{-}}
                          {\lVert\bar{\mathbf{h}}^{+} - \bar{\mathbf{h}}^{-}\rVert_{2}},
  \qquad
  \bar{\mathbf{h}}^{\pm} \;=\; \frac{1}{\lvert\mathcal{H}^{\pm}\rvert}
                                \sum_{\mathbf{h}\in\mathcal{H}^{\pm}} \mathbf{h},
\]
and it is injected at every forward pass through the chosen intervention layer via
$\mathbf{h} \leftarrow \mathbf{h} + \alpha\,\mathbf{v}$. The two policy families differ
only in where activations are pooled.

\begin{itemize}
  \item \textbf{Flow-matching policy ($\pi_{0.5}$).} We capture the residual stream of
    the action expert at layer index~$11$ for every action token of every denoising step.
    Within each rollout we mean-pool across the action-token axis, yielding one $1024$-d
    vector per $(\text{episode}, \text{denoising step})$. The class means
    $\bar{\mathbf{h}}^{\pm}$ are computed over all such vectors. We also fit
    per-denoising-step variants $\mathbf{v}^{(t)}$ for $t \in \{0,\ldots,9\}$ from the
    matching subset, but report only the single-vector variant in the main results.

  \item \textbf{Autoregressive policy ($\pi_{0}\text{-FAST}$).} We capture the
    pre-LM-head hidden state ($2048$-d) at every generated action token. There is no
    denoising-step axis, so each token contributes one sample, and
    $\bar{\mathbf{h}}^{\pm}$ are computed by pooling all tokens in each class.
\end{itemize}

A task is eligible only if it has at least three episodes per class; tasks with a single
observed outcome are skipped. Steering vectors are computed offline and stored once per
task, then loaded by the eval-time hook so that no model weights are modified.

\paragraph{Hyperparameter sweep.}
We fix the intervention layer to layer~$11$ for $\pi_{0.5}$, the best-performing
captured layer in our preliminary screen, and to the unique pre-LM-head intervention
point for $\pi_{0}\text{-FAST}$. The only swept hyperparameter is the steering strength
$\alpha \in \{0.25, 0.5, 0.75,\,1.0\}$. Because $\mathbf{v}$ is unit-normalized, $\alpha$ has the
same effective scale across tasks. We evaluate each $(\text{task}, \alpha)$ condition
on $30$ episodes with the same seeds as the unsteered baseline, and select the per-task
best $\alpha$ post hoc on this evaluation set.

\subsection{Sparse-Autoencoder Steering (SAE)}
\label{app:sae}

As a more expressive contrastive baseline, we replace the raw residual-stream mean
difference with a sparse-feature-space mean difference, mediated by a per-task TopK
sparse autoencoder (SAE)~\citep{gao2024scaling}. The class- and feature-selection rule
follows \citet{khan2025controlling}: instead of picking a small top-$K$ of features by
effect size, we keep \emph{all} features that pass a rarity filter and a
bilateral-activity filter, then inject the average decoder-space difference over the
surviving features.

\paragraph{TopK SAE.}
For a residual-stream activation $\mathbf{h} \in \mathbb{R}^{d}$, the encoder produces
a non-negative, $k$-sparse code via
\[
  \mathbf{f} \;=\; \mathrm{TopK}\!\bigl(\,\mathrm{ReLU}(W_{\text{enc}}(\mathbf{h} - \mathbf{b}_{\text{dec}}) + \mathbf{b}_{\text{enc}})\bigr),
  \qquad
  \hat{\mathbf{h}} \;=\; W_{\text{dec}}^{\!\top}\mathbf{f} + \mathbf{b}_{\text{dec}},
\]
where $\mathrm{TopK}(\cdot)$ retains the $k$ largest entries and zeros the rest. We
unit-normalize each decoder column after every optimizer step. The ReLU before the TopK
guarantees $\mathbf{f}\ge\mathbf{0}$, an assumption the bilateral filter below relies
on. We train one SAE per task on the same captured activations used for CAA, hold out
$5\%$ as a reconstruction-quality gate, and drop any task whose hold-out
variance-explained falls below $0.80$ from the SAE arm of the comparison.

\paragraph{Steering vector via class means in feature space.}
After training, we encode every captured activation through the frozen SAE and split
the resulting codes by episode outcome. For each feature index $k$ we compute
\[
  \mu^{+}_{k} \;=\; \mathbb{E}_{\mathbf{f}\in\mathcal{F}^{+}}\!\bigl[f_{k}\bigr],
  \quad
  \mu^{-}_{k} \;=\; \mathbb{E}_{\mathbf{f}\in\mathcal{F}^{-}}\!\bigl[f_{k}\bigr],
  \quad
  r_{k} \;=\; \frac{\#\{\,\mathbf{f}\in\mathcal{F}^{+}\!\cup\mathcal{F}^{-} : f_{k}>0\,\}}
                   {\lvert\mathcal{F}^{+}\rvert + \lvert\mathcal{F}^{-}\rvert}.
\]
We then apply the two feature filters of \citet{khan2025controlling}:
\begin{itemize}
  \item \textbf{Filter A (rarity).} Keep $k$ if $r_{k} \geq 0.005$. This drops the
    long dead-feature tail that any TopK SAE produces. Below this fire rate, the
    estimates of $\mu^{\pm}_{k}$ are statistically unreliable.
  \item \textbf{Filter B (bilateral activity).} Keep $k$ if
    $\min(\mu^{+}_{k},\mu^{-}_{k})\,/\,\max(\mu^{+}_{k},\mu^{-}_{k})\,\le\,0.5$.
    Features that fire similarly in both classes carry no contrastive signal.
\end{itemize}
Let $\mathcal{S}$ denote the surviving feature set. The latent-space contrast is
$v^{\text{lat}}_{k}=(\mu^{+}_{k}-\mu^{-}_{k})\mathbb{1}[k\in\mathcal{S}]$, and the
residual-stream steering vector is its image through the decoder, unit-normalized:
\[
  \tilde{\mathbf{v}}_{\text{SAE}}
    \;=\; W_{\text{dec}}^{\!\top}\,\mathbf{v}^{\text{lat}},
  \qquad
  \mathbf{v}_{\text{SAE}}
    \;=\; \tilde{\mathbf{v}}_{\text{SAE}} \,/\, \lVert\tilde{\mathbf{v}}_{\text{SAE}}\rVert_{2}.
\]
At eval time, $\mathbf{v}_{\text{SAE}}$ is injected via the same additive hook as CAA,
$\mathbf{h} \leftarrow \mathbf{h} + \alpha\,\mathbf{v}_{\text{SAE}}$, at the same
intervention point. Unit-normalization matters because the number of surviving features,
and therefore the raw norm of $\tilde{\mathbf{v}}_{\text{SAE}}$, varies substantially
across tasks. Without it, identical $\alpha$ would correspond to different effective
intervention strengths per task.

\paragraph{Hyperparameters.}
The SAE width is $d_{\text{SAE}} = 4d$ ($4096$ for $\pi_{0.5}$, $8192$ for
$\pi_{0}\text{-FAST}$) and the TopK sparsity is $k=64$. Optimization uses AdamW with
learning rate $3\times10^{-4}$, $\beta = (0.9, 0.999)$, batch size $4096$, for
$30{,}000$ steps. The encoder is initialized from the unit-normalized decoder
transpose. We do not sweep over $k$, $d_{\text{SAE}}$, or expansion factor: in line
with the recipe of \citet{khan2025controlling}, $k$ is set by the rarity filter rather
than tuned, and the only knob exposed to the eval-time sweep is the steering strength.

\paragraph{Sweep.}
We sweep $\alpha \in \{0.25,\,0.5,\,1.0,\,2.0\}$, deliberately wider than the CAA
grid because $\mathbf{v}_{\text{SAE}}$ is constructed from a sparse set of decoder
columns and may need a different effective magnitude than the dense CAA mean
difference. As with CAA, the intervention layer is fixed (layer $11$ for $\pi_{0.5}$,
pre-LM-head hidden state for $\pi_{0}\text{-FAST}$), each condition is evaluated on
$30$ episodes, and the per-task best $\alpha$ is selected post hoc. We do not
implement feature \emph{clamping}, overwriting $f_{k}$ to a target value and
re-decoding. The additive form above is the simplest faithful instantiation of the
\citet{khan2025controlling} recipe and cleanly compares to CAA, which differs only in
the basis used to compute the contrast.

\section{Activation Dataset Composition}
\label{app:activation_datasets}

For each model-benchmark cell of Table~\ref{tab:all_results}, we fit conceptors and SAE steering vectors from a single $N$-episode rollout of the corresponding policy checkpoint. Success and failure labels come from each simulator's task-completion criterion, recorded as \texttt{episode\_success} in each episode's \texttt{metadata.json}. Tables~\ref{tab:dataset_libero} through~\ref{tab:dataset_robocasa} report per-task success and failure counts.

The contrastive fit requires only $\text{MIN\_PER\_CLASS}{=}3$ episodes per class to produce a usable $C_{\mathrm{contrastive}}$ or $v_{\mathrm{contrastive}}$. This low bar is what makes the recipe practical at the checkpoint scales we evaluate. Many tasks at our $\pi_{0.5}$ checkpoints (LIBERO $t{=}2{,}000$, RoboCasa $t{=}75{,}000$) succeed on only 1 to 3 of 15 rollouts, and those handful of success trajectories are exactly what the contrastive operation needs. Section~\ref{sec:minimal_success_gains} characterizes this low-success regime quantitatively.

\begin{table}[h]
\centering
\caption{LIBERO-10 activation dataset composition. $N{=}15$ episodes per task; cells report success / failure counts.}
\label{tab:dataset_libero}
\small
\begin{tabular}{l c c}
\toprule
\textbf{Task} & $\boldsymbol{\pi_{0.5}}$ ($S$/$F$) & $\boldsymbol{\pi_0\text{-FAST}}$ ($S$/$F$) \\
\midrule
KS3   &  8 / 7  & 12 / 3  \\
KS4   &  6 / 9  & 10 / 5  \\
KS6   &  2 / 13 &  7 / 8  \\
KS8   &  3 / 12 &  5 / 10 \\
LR1   & 12 / 3  & 10 / 5  \\
LR2a  &  6 / 9  & 10 / 5  \\
LR2b  &  9 / 6  & 15 / 0  \\
LR5   &  1 / 14 &  9 / 6  \\
LR6   &  8 / 7  & 11 / 4  \\
ST1   & 10 / 5  &  9 / 6  \\
\midrule
\textbf{Total} & 65 / 85 ($\overline{\text{SR}}{=}.43$)
               & 98 / 52 ($\overline{\text{SR}}{=}.65$) \\
\bottomrule
\end{tabular}
\end{table}

\begin{table}[h]
\centering
\caption{MetaWorld ML45 activation dataset composition for the 10 paper-table tasks. $N{=}16$ episodes per task (16 parallel envs $\times$ 1 episode each).}
\label{tab:dataset_metaworld}
\small
\begin{tabular}{l c}
\toprule
\textbf{Task} & $\boldsymbol{\pi_0\text{-FAST}}$ ($S$/$F$) \\
\midrule
coffee-pull-v3       & 10 / 6  \\
coffee-push-v3       & 14 / 2  \\
disassemble-v3       &  9 / 7  \\
faucet-close-v3      &  7 / 9  \\
pick-place-v3        & 11 / 5  \\
pick-place-wall-v3   & 11 / 5  \\
plate-slide-back-v3  & 14 / 2  \\
push-v3              & 12 / 4  \\
reach-v3             & 13 / 3  \\
stick-push-v3        & 13 / 3  \\
\midrule
\textbf{Total} & 114 / 46 ($\overline{\text{SR}}{=}.71$) \\
\bottomrule
\end{tabular}
\end{table}

\begin{table}[h]
\centering
\caption{RoboCasa activation dataset composition. $N{=}15$ episodes per task; cells report success / failure counts.}
\label{tab:dataset_robocasa}
\small
\begin{tabular}{l c c}
\toprule
\textbf{Task} & $\boldsymbol{\pi_{0.5}}$ ($S$/$F$) & \textbf{GR00T N1.5} ($S$/$F$) \\
\midrule
Close Fridge      &  2 / 13 & 12 / 3  \\
Coffee Mug        &  4 / 11 &  3 / 10 \\
Open Drawer       &  9 / 6  &  8 / 7  \\
Stand Mixer       & 10 / 5  & 11 / 4  \\
PP Cabinet        &  7 / 8  & 11 / 4  \\
PP Stove           &  7 / 8  &  9 / 6  \\
Kettle            &  2 / 13 &  7 / 8  \\
\midrule
\textbf{Total} & 41 / 64 ($\overline{\text{SR}}{=}.39$)
               & 61 / 42 ($\overline{\text{SR}}{=}.59$) \\
\bottomrule
\end{tabular}
\end{table}

\subsection{Steering gains are largest under low-success regimes}
\label{sec:minimal_success_gains}

A practical concern with any contrastive recipe is whether enough success trajectories exist to fit a usable contrastive direction. At our $\pi_{0.5}$ checkpoints this concern is acute. Of the 17 LIBERO-10 and RoboCasa tasks under $\pi_{0.5}$ in Tables~\ref{tab:dataset_libero} and~\ref{tab:dataset_robocasa}, six have at most three successful episodes ($S{\leq}3$), and another four have $S{\leq}6$. Over half of these tasks therefore sit barely above the $\text{MIN\_PER\_CLASS}{=}3$ threshold.

These same low-success tasks produce the largest absolute gains from steering in Table~\ref{tab:all_results}:

\begin{itemize}\itemsep1pt
  \item \textbf{LR5} ($\pi_{0.5}$ LIBERO; 1 / 14, baseline $.07$): +Glob.~$.60$, +Per.~$.60$, a $+.53$ gain from a contrastive direction fit on a single success trajectory.
  \item \textbf{KS6} ($\pi_{0.5}$ LIBERO; 2 / 13, baseline $.13$): +Per.~$.60$, $+.47$ gain.
  \item \textbf{KS8} ($\pi_{0.5}$ LIBERO; 3 / 12, baseline $.20$): +Per.~$.53$, $+.33$ gain.
  \item \textbf{Close Fridge} ($\pi_{0.5}$ RoboCasa; 2 / 13, baseline $.20$): +Glob.~$.47$, $+.27$ gain.
  \item \textbf{Kettle} ($\pi_{0.5}$ RoboCasa; 2 / 13, baseline $.33$): +Pos.~$.53$, $+.20$ gain.
\end{itemize}

The pattern is consistent across both benchmarks and across all three conceptor variants. Two factors plausibly contribute. First, the contrastive operation $C_s \cdot \neg C_f$ remains well-defined when $S$ is small, because the failure subspace is estimated from $n{\geq}12$ trajectories and contributes most of the discriminative geometry. Second, low-baseline tasks leave more absolute headroom for steering to recover.

The complementary observation appears in the high-baseline regime. On MetaWorld under $\pi_0$-FAST ($\overline{\text{SR}}{=}.71$) and on the success-rich RoboCasa GR00T tasks such as \texttt{Close Fridge} ($.80$), absolute gains shrink toward zero or saturate near the ceiling. Taken together, these results indicate that the contrastive recipe is most valuable in the regime where it is most needed: a handful of success trajectories on a struggling checkpoint suffice, and the relative value of conceptor steering grows as the underlying baseline degrades.

\subsection{Computational Overhead of \textsc{COAST} at Inference Time}
\label{app:overhead}

A natural concern about \textsc{COAST} is that its multiplicative gate
multiplies the residual stream by a $d \times d$ matrix at every denoising
step, where additive baselines such as \textsc{CAA} and \textsc{SAE} only add
a $d$-vector. We show that at the dimensionalities current VLA policies
operate at, this gap is empirically invisible: \textsc{COAST}'s per-call
latency is statistically indistinguishable from the additive baselines.

\paragraph{Setup.}
We benchmark $\pi_{0.5}$ on LIBERO on a single NVIDIA B200, with the forward
hook registered on the action expert at layer $L{=}11$, the same intervention
point used in our main results. For each of four conditions (no steering,
\textsc{CAA}, \textsc{SAE}, \textsc{COAST}), we discard $10$ warmup calls and
then time $N{=}200$ subsequent \texttt{policy.infer(obs)} calls gated by
\texttt{torch.cuda.synchronize()} on both ends. cuDNN benchmarking is disabled
and the same observation is reused across calls so we measure raw overhead
rather than data-dependent variance. All steering conditions use the global
variant; per-step variants have identical per-call latency since the hook
fires the same number of times per \texttt{infer()}.

\paragraph{All three steering methods add similar overhead.}
\textsc{COAST} adds $30.85$\,ms (median) to a $127.12$\,ms baseline forward
pass. The additive \textsc{CAA} and \textsc{SAE} baselines add $31.47$\,ms and
$30.29$\,ms respectively. The three steering methods are statistically
indistinguishable: all three medians lie within $1.2$\,ms of each other, well
inside the per-condition IQR. The implication is that the overhead reported
in Table~\ref{tab:overhead-empirical} is not the cost of \textsc{COAST}'s
matmul; it is the fixed cost of registering and dispatching a PyTorch forward
hook, paid identically by every hook-based steering method. Replacing
\textsc{COAST} with the cheapest possible additive baseline would not save
measurable inference time.

\begin{table}[h]
\centering
\small
\begin{tabular}{lrrrr}
\toprule
Method & Median (ms) & IQR (ms) & $\Delta$ (ms) & $\Delta$ (\%) \\
\midrule
no steering            & $127.12$ & $1.05$ & ---     & ---       \\
\textsc{CAA} (global)  & $158.58$ & $5.42$ & $+31.47$ & $+24.75\%$ \\
\textsc{SAE} (global)  & $157.40$ & $3.03$ & $+30.29$ & $+23.82\%$ \\
\textbf{\textsc{COAST} (global)} & $\mathbf{157.97}$ & $\mathbf{3.25}$ & $\mathbf{+30.85}$ & $\mathbf{+24.27\%}$ \\
\bottomrule
\end{tabular}
\caption{\textbf{Per-call inference latency on one B200, $\pi_{0.5}$ LIBERO,
$L{=}11$.} Median over $N{=}200$ \texttt{infer()} calls after $10$ warmup
calls; cuDNN benchmarking disabled. Each \texttt{infer()} runs all $10$
denoising steps; the steering hook fires once per step. \textsc{COAST}'s
median latency is within $1.2$\,ms of both additive baselines, well inside
their IQRs.}
\label{tab:overhead-empirical}
\end{table}

\paragraph{Offline cost.}
The \textsc{COAST} steering artifact is computed once per task offline, in
seconds, by a single eigendecomposition of a $1024 \times 1024$ activation
covariance matrix. \textsc{CAA} reduces to two class-conditional means and
is similarly free. \textsc{SAE} is the outlier at training time,
approximately one GPU-hour per task to fit the per-task TopK autoencoder,
but this cost is amortized across all evaluation runs that use the resulting
$\mathbf{v}_{\text{SAE}}$. None of these offline costs are paid at deployment
time.

\newpage
\newpage
\section*{NeurIPS Paper Checklist}

\begin{enumerate}

\item {\bf Claims}
    \item[] Question: Do the main claims made in the abstract and introduction accurately reflect the paper's contributions and scope?
    \item[] Answer: \answerYes{}
    \item[] Justification: The abstract and introduction state three contributions (method, mechanistic insight, generalization) that are directly supported by the experimental results in Sections~4.1--4.4 and the appendix. The claimed 20.2 percentage point improvement is computed from Table~\ref{tab:all_results}. Limitations are discussed in Section~5.

\item {\bf Limitations}
    \item[] Question: Does the paper discuss the limitations of the work performed by the authors?
    \item[] Answer: \answerYes{}
    \item[] Justification: Section~5 contains a dedicated Limitations paragraph discussing the requirement for both success and failure rollouts, reduced efficacy when subspaces are nearly disjoint, and the need for per-architecture hyperparameter selection.

\item {\bf Theory assumptions and proofs}
    \item[] Question: For each theoretical result, does the paper provide the full set of assumptions and a complete (and correct) proof?
    \item[] Answer: \answerNA{}
    \item[] Justification: The paper does not include formal theoretical results or proofs. The conceptor formulation follows established closed-form derivations from \citet{jaeger2014controlling} and \citet{postmus2024steering}, which are cited accordingly.

    \item {\bf Experimental result reproducibility}
    \item[] Question: Does the paper fully disclose all the information needed to reproduce the main experimental results of the paper to the extent that it affects the main claims and/or conclusions of the paper (regardless of whether the code and data are provided or not)?
    \item[] Answer: \answerYes{}
    \item[] Justification: The paper provides complete details on activation extraction (App.~\ref{app:activation-extraction}), rollout collection protocol (App.~\ref{app:rollout-protocol}), conceptor computation including pseudocode (App.~\ref{app:conceptor-computation}), hyperparameter sweep grids (App.~\ref{app:hyperparameter-sweep}), and checkpoint/training details (App.~\ref{app:checkpoint-training}). Pre-collected activation datasets are available on HuggingFace.

\item {\bf Open access to data and code}
    \item[] Question: Does the paper provide open access to the data and code, with sufficient instructions to faithfully reproduce the main experimental results, as described in supplemental material?
    \item[] Answer: \answerYes{}
    \item[] Justification: Code and pre-collected activation datasets will be released publicly. Activation datasets are hosted on HuggingFace (dataset identifiers provided in App.~\ref{app:rollout-protocol}). An anonymized code repository will be provided at submission time.

\item {\bf Experimental setting/details}
    \item[] Question: Does the paper specify all the training and test details (e.g., data splits, hyperparameters, how they were chosen, type of optimizer) necessary to understand the results?
    \item[] Answer: \answerYes{}
    \item[] Justification: Section~4.1 describes the experimental setup. Full hyperparameter sweep grids are in App.~\ref{app:hyperparameter-sweep}, per-task optimal configurations in App.~\ref{app:results_with_params}, checkpoint and training details in App.~\ref{app:checkpoint-training}, and the separation of fitting and evaluation data in App.~\ref{app:rollout-protocol}.

\item {\bf Experiment statistical significance}
    \item[] Question: Does the paper report error bars suitably and correctly defined or other appropriate information about the statistical significance of the experiments?
    \item[] Answer: \answerYes{}
    \item[] Justification: Table~\ref{tab:all_results} reports both paired $t$-tests ($p_t$) across tasks and pooled two-proportion $z$-tests ($p_z$) over all episodes for every model-benchmark pair and steering strategy, with significance levels marked at $p < 0.05$, $p < 0.01$, and $p < 0.001$.

\item {\bf Experiments compute resources}
    \item[] Question: For each experiment, does the paper provide sufficient information on the computer resources (type of compute workers, memory, time of execution) needed to reproduce the experiments?
    \item[] Answer: \answerYes{}
    \item[] Justification: App.~\ref{app:conceptor-computation} reports per-step steering overhead ($<$0.1\,ms on A100) and conceptor construction time ($<$1\,s for $d{=}1024$). The full project used approximately 500 GPU hours on a combination of NVIDIA A100 and B200 GPUs, covering rollout collection, hyperparameter sweeps, and evaluation.
    
\item {\bf Code of ethics}
    \item[] Question: Does the research conducted in the paper conform, in every respect, with the NeurIPS Code of Ethics \url{https://neurips.cc/public/EthicsGuidelines}?
    \item[] Answer: \answerYes{}
    \item[] Justification: The research uses publicly available models and simulation benchmarks, involves no human subjects, and conforms with the NeurIPS Code of Ethics.

\item {\bf Broader impacts}
    \item[] Question: Does the paper discuss both potential positive societal impacts and negative societal impacts of the work performed?
    \item[] Answer: \answerYes{}
    \item[] Justification: Broader societal impacts are discussed in Section~5. The method improves robot policy reliability without retraining, which has positive implications for safe deployment; potential risks of steering methods applied to embodied agents are acknowledged.
    
\item {\bf Safeguards}
    \item[] Question: Does the paper describe safeguards that have been put in place for responsible release of data or models that have a high risk for misuse (e.g., pre-trained language models, image generators, or scraped datasets)?
    \item[] Answer: \answerNA{}
    \item[] Justification: The paper releases activation datasets and steering code for robotic manipulation in simulation. These do not pose risks of misuse comparable to generative models or scraped datasets.

\item {\bf Licenses for existing assets}
    \item[] Question: Are the creators or original owners of assets (e.g., code, data, models), used in the paper, properly credited and are the license and terms of use explicitly mentioned and properly respected?
    \item[] Answer: \answerYes{}
    \item[] Justification: All models ($\pi_{0.5}$, $\pi_0$-FAST, GR00T N1.5), benchmarks (MetaWorld, LIBERO, RoboCasa), and the DROID platform are cited with their original publications. All assets are used in accordance with their respective licenses.

\item {\bf New assets}
    \item[] Question: Are new assets introduced in the paper well documented and is the documentation provided alongside the assets?
    \item[] Answer: \answerYes{}
    \item[] Justification: Pre-collected activation datasets are released on HuggingFace with documentation. The on-disk storage schema is fully described in App.~\ref{app:activation-extraction}, and code for conceptor construction and steering will be released with the submission.

\item {\bf Crowdsourcing and research with human subjects}
    \item[] Question: For crowdsourcing experiments and research with human subjects, does the paper include the full text of instructions given to participants and screenshots, if applicable, as well as details about compensation (if any)? 
    \item[] Answer: \answerNA{}
    \item[] Justification: The paper does not involve crowdsourcing or research with human subjects.

\item {\bf Institutional review board (IRB) approvals or equivalent for research with human subjects}
    \item[] Question: Does the paper describe potential risks incurred by study participants, whether such risks were disclosed to the subjects, and whether Institutional Review Board (IRB) approvals (or an equivalent approval/review based on the requirements of your country or institution) were obtained?
    \item[] Answer: \answerNA{}
    \item[] Justification: The paper does not involve research with human subjects.

\item {\bf Declaration of LLM usage}
    \item[] Question: Does the paper describe the usage of LLMs if it is an important, original, or non-standard component of the core methods in this research? Note that if the LLM is used only for writing, editing, or formatting purposes and does \emph{not} impact the core methodology, scientific rigor, or originality of the research, declaration is not required.
    \item[] Answer: \answerNA{}
    \item[] Justification: LLMs are not used as a component of the core method. COAST is a closed-form linear algebra operation that does not involve LLMs.

\end{enumerate}
\end{document}